\documentclass{article}

\usepackage[final,nonatbib]{nips_2017}

\usepackage[utf8]{inputenc} 
\usepackage[T1]{fontenc}    
\usepackage{hyperref}       
\usepackage{url}            
\usepackage{booktabs}       
\usepackage{amsfonts}       
\usepackage{nicefrac}       
\usepackage{microtype}      
\usepackage{amsmath,amsfonts,amssymb}
\usepackage{cancel}
\usepackage[usenames, dvipsnames]{color}
\allowdisplaybreaks

\usepackage{glossaries}
 
\newacronym{gp}{GP}{Gaussian process}

\usepackage{graphicx}
\usepackage{cleveref}
\usepackage{subcaption}

\newcommand{\kff}{\mathbf{K}_\mathbf{ff}}

\newcommand{\kfu}{\mathbf{K}_\mathbf{fu}}
\newcommand{\kuf}{\mathbf{K}_\mathbf{uf}}
\newcommand{\kuu}{\mathbf{K}_\mathbf{uu}}
\newcommand{\kss}{\mathbf{K}_\mathbf{ss}}

\newcommand{\kaa}{\mathbf{K}_\mathbf{aa}}
\newcommand{\kab}{\mathbf{K}_\mathbf{ab}}
\newcommand{\kba}{\mathbf{K}_\mathbf{ba}}
\newcommand{\ksb}{\mathbf{K}_\mathbf{sb}}
\newcommand{\kbs}{\mathbf{K}_\mathbf{bs}}
\newcommand{\kbb}{\mathbf{K}_\mathbf{bb}}
\newcommand{\lb}{\mathbf{L}_\mathbf{b}}
\newcommand{\kfb}{\mathbf{K}_\mathbf{fb}}
\newcommand{\khatb}{\mathbf{K}_{\hat{\mathbf{f}}\mathbf{b}}}
\newcommand{\kbhat}{\mathbf{K}_{\mathbf{b}\hat{\mathbf{f}}}}
\newcommand{\kbf}{\mathbf{K}_\mathbf{bf}}
\newcommand{\ma}{\mathbf{m}_\mathbf{a}}
\newcommand{\Sa}{\mathbf{S}_\mathbf{a}}
\newcommand{\Wa}{\mathbf{W}_\mathbf{a}}
\newcommand{\Qa}{\mathbf{Q}_\mathbf{a}}
\newcommand{\Da}{\mathbf{D}_\mathbf{a}}
\newcommand{\Ma}{\mathbf{M}_\mathbf{a}}
\newcommand{\Siga}{\Sigma_\mathbf{a}}
\newcommand{\Sigy}{\Sigma_\mathbf{y}}
\newcommand{\Wf}{\mathbf{W}_\mathbf{f}}
\newcommand{\Qf}{\mathbf{Q}_\mathbf{f}}
\newcommand{\la}{\mathbf{L}_\mathbf{q}}
\newcommand{\laT}{\mathbf{L}^\intercal_\mathbf{q}}
\newcommand{\mahat}{\hat{\mathbf{m}}_\mathbf{a}}
\newcommand{\Sahat}{\hat{\mathbf{S}}_\mathbf{a}}
\newcommand{\Syhat}{\hat{\mathbf{S}}_\mathbf{y}}
\newcommand{\mvec}{\mathbf{m}}
\newcommand{\mvecT}{\mathbf{m}^\intercal}
\newcommand{\mahatT}{\hat{\mathbf{m}}_\mathbf{a}^\intercal}
\newcommand{\mcav}{\mathbf{m}_\mathrm{cav}}
\newcommand{\mcavT}{\mathbf{m}_\mathrm{cav}^\intercal}
\newcommand{\Vcav}{\mathbf{V}_\mathrm{cav}}

\newcommand{\norm}{\mathcal{N}}

\newcommand{\xvec}{\mathbf{x}}
\newcommand{\fvec}{\mathbf{f}}

\newcommand{\uvec}{\mathbf{u}}

\newcommand{\yvec}{\mathbf{y}}
\newcommand{\yvecT}{\mathbf{y}^\intercal}
\newcommand{\svec}{\mathbf{s}}
\newcommand{\Svec}{\mathbf{S}}

\newcommand{\Vvec}{\mathbf{V}}

\newcommand{\zvec}{\mathbf{z}}

\newcommand{\zero}{\mathbf{0}}

\newcommand{\dd}{\mathrm{d}}

\newcommand{\avec}{\mathbf{a}}
\newcommand{\bvec}{\mathbf{b}}

\usepackage{xcolor}
\hypersetup{
    colorlinks,
    linkcolor={red!50!black},
    citecolor={blue!50!black},
    urlcolor={blue!80!black}
}

\usepackage[backgroundcolor=white,linecolor=red,bordercolor=white,textsize=tiny,textwidth=10mm]{todonotes}
\usepackage{titlesec}
\titlespacing{\section}{3pt}{0.9ex}{0.7ex}
\titlespacing{\subsection}{3pt}{0.5ex}{0.4ex}
\titlespacing{\subsubsection}{3pt}{0.2ex}{0ex}

\usepackage{appendix}

\title{Streaming Sparse Gaussian Process Approximations}

\author{
Thang D.~Bui\thanks{These authors contributed equally to this work.}\\
\And Cuong V.~Nguyen$^*$\\
\And Richard E.~Turner\\
\and
Department of Engineering, University of Cambridge, UK\\
\texttt{\{tdb40,vcn22,ret26\}@cam.ac.uk} \\
}

\begin{document}

\maketitle

\begin{abstract}
Sparse pseudo-point approximations for Gaussian process (GP) models provide a suite of methods that support deployment of GPs in the large data regime and enable analytic intractabilities to be sidestepped. However, the field lacks a principled method to handle streaming data in which both the posterior distribution over function values and the hyperparameter estimates are updated in an online fashion. The small number of existing approaches either use suboptimal hand-crafted heuristics for hyperparameter learning, or suffer from catastrophic forgetting or slow updating when new data arrive. This paper develops a new principled framework for deploying Gaussian process probabilistic models in the streaming setting, providing  methods for learning hyperparameters and optimising pseudo-input locations. The proposed framework is assessed using synthetic and real-world datasets.
\end{abstract}

\section{Introduction}
\label{sec:intro}

Probabilistic models employing Gaussian processes have become a standard approach to solving many machine learning tasks, thanks largely to the modelling flexibility, robustness to overfitting, and well-calibrated uncertainty estimates afforded by the approach \cite{rasmussen2005gpml}. One of the pillars of the modern Gaussian process probabilistic modelling approach is a set of sparse approximation schemes that allow the prohibitive computational cost of GP methods, typically $\mathcal{O}(N^3)$ for training and $\mathcal{O}(N^2)$ for prediction where $N$ is the number of training points, to be substantially reduced whilst still retaining accuracy. Arguably the most important and influential approximations of this sort are pseudo-point approximation schemes that employ a set of  $M \ll N$ pseudo-points to summarise the observational data thereby reducing computational costs to $\mathcal{O}(NM^2)$ and $\mathcal{O}(M^2)$ for training and prediction, respectively \cite{snelson+ghahramani:2006, titsias2009variational}. Stochastic optimisation methods that employ mini-batches of training data can be used to further reduce computational costs \cite{hensman2013gaussian, HenMatGha15, dezfouli2015scalable,hernandez2016gpc}, allowing GPs to be scaled to datasets comprising millions of data points.

The focus of this paper is to provide a comprehensive framework for deploying the Gaussian process probabilistic modelling approach to streaming data. That is, data that arrive sequentially in an online fashion, possibly in small batches, and whose number are not known a priori (and indeed may be infinite). The vast majority of previous work has focussed exclusively on the batch setting and there is not a satisfactory framework that supports learning and approximation in the streaming setting. A na\"ive approach might simply incorporate each new datum as they arrived into an ever-growing dataset and retrain the GP model from scratch each time. With infinite computational resources, this approach is optimal, but in the majority of practical settings, it is intractable. A feasible alternative would train on just the most recent $K$ training data points, but this completely ignores potentially large amounts of informative training data and it does not provide a method for incorporating the old model into the new one which would save computation (except perhaps through initialisation of the hyperparameters). Existing, sparse approximation schemes could be applied in the same manner, but they merely allow $K$ to be increased, rather than allowing all previous data to be leveraged, and again do not utilise intermediate approximate fits. 

What is needed is a method for performing learning and sparse approximation that incrementally updates the previously fit model using the new data. Such an approach would utilise all the previous training data (as they will have been incorporated into the previously fit model) and leverage as much of the previous computation as possible at each stage (since the algorithm only requires access to the data at the current time point). Existing {\it stochastic} sparse approximation methods could potentially be used by collecting the streamed data into mini-batches. However, the assumptions underpinning these methods are ill-suited to the streaming setting and they perform poorly (see \cref{sec:background,sec:exp}).

This paper provides a new principled framework for deploying Gaussian process probabilistic models in the streaming setting. The framework subsumes  Csat\'{o} and Opper's two seminal approaches to online regression \cite{csato+opper:2002, csato:2002} that were based upon the variational free energy (VFE) and expectation propagation (EP) approaches to approximate inference respectively. In the new framework, these algorithms are recovered as special cases. We also provide principled methods for learning hyperparameters (learning was not treated in the original work and the extension is non-trivial) and optimising pseudo-input locations (previously handled via hand-crafted heuristics). The approach also relates to the streaming variational Bayes framework \cite{broderick2013streaming}. We review background material in the next section and detail the technical contribution in \cref{sec:online_vfe}, followed by several experiments on synthetic and real-world data in \cref{sec:exp}.

\section{Background}
\label{sec:background}

Regression models that employ Gaussian processes are state of the art for many datasets \cite{thang:deepGP}. In this paper we focus on the simplest GP regression model as a test case of the streaming framework for inference and learning. Given $N$ input and real-valued output pairs $\{\xvec_n, y_n\}_{n=1}^{N}$, a standard GP regression model assumes $y_n = f(\xvec_n) + \epsilon_n$, where $f$ is an unknown function that is corrupted by Gaussian observation noise $\epsilon_n \sim \norm (0, \sigma_y^2)$. Typically, $f$ is assumed to be drawn from a zero-mean GP prior $f \sim \mathcal{GP}(\mathbf{0}, k(\cdot, \cdot|\theta))$ whose covariance function depends on hyperparameters $\theta$.
In this simple model, the posterior over $f$, $p(f | \yvec, \theta)$, and the marginal likelihood $p(\yvec | \theta)$ can be computed analytically (here we have collected the observations into a vector $\yvec = \{ y_n \}_{n=1}^N$).\footnote{The dependence on the inputs $\{\xvec_n\}_{n=1}^N$ of the posterior, marginal likelihood, and other quantities is suppressed throughout to lighten the notation.} However, these quantities present a computational challenge resulting in an $O(N^3)$ complexity for maximum likelihood training and $O(N^2)$ per test point for prediction.

This prohibitive complexity of exact learning and inference in GP models has driven the development of many sparse approximation frameworks \cite{quinonero+rasmussen:2005,BuiYanTur17}. In this paper, we focus on the variational free energy approximation scheme \cite{titsias2009variational,matthews+al:2016} which lower bounds the marginal likelihood of the data using a variational distribution $q(f)$ over the latent function:
\begin{align}
\log p(\yvec|\theta) = \log \int \dd f \; p(\yvec, f|\theta) \geq \int \dd f \; q(f) \log \frac{p(\yvec,f|\theta)}{q(f)} = \mathcal{F}_{\mathrm{vfe}}(q, \theta).
\end{align}
Since $\mathcal{F}_{\mathrm{vfe}}(q, \theta) = \log p(\yvec|\theta) - \mathrm{KL}[q(f)||p(f|\yvec, \theta)]$, where $\mathrm{KL}[\cdot||\cdot]$ denotes the Kullback–Leibler divergence, maximising this lower bound with respect to $q(f)$ guarantees the approximate posterior gets {\it closer} to the exact posterior $p(f|\yvec, \theta)$. Moreover, the variational bound $\mathcal{F}_{\mathrm{vfe}}(q, \theta)$  approximates the marginal likelihood and can be used for learning the hyperparameters $\theta$.

In order to arrive at a computationally tractable method, the approximate posterior is parameterized via a set of $M$ pseudo-points $\uvec$ that are a subset of the function values $f = \{f_{\neq \uvec}, \uvec \}$ and which will summarise the data. Specifically, the approximate posterior is assumed to be $q(f) = p(f_{\neq \uvec}|\uvec, \theta) q(\uvec)$, where $q(\uvec)$ is a variational distribution over $\uvec$ and $p(f_{\neq \uvec}|\uvec, \theta)$ is the prior distribution of the remaining latent function values. This assumption allows the following critical cancellation that results in a computationally tractable lower bound:
\begin{align}
\mathcal{F}_{\mathrm{vfe}}(q(\uvec), \theta) 
	&= \int \dd f \; q(f) \log \frac{p(\yvec|f, \theta) p(\uvec | \theta) \bcancel{p (f_{\neq \uvec}|\uvec, \theta)}}{\bcancel{p(f_{\neq \uvec}|\uvec, \theta)} q(\uvec)} \nonumber \\
	&= - \mathrm{KL}[q(\uvec)||p(\uvec|\theta)] + \sum_{n} \int \dd \uvec \; q(\uvec) p(f_n|\uvec,\theta) \log p(y_n|f_n, \theta), \nonumber
\end{align}
where $f_n = f(\xvec_n)$ is the latent function value at $\xvec_n$.
For the simple GP regression model considered here, closed-form expressions for the optimal variational approximation $q_\mathrm{vfe}(f)$ and the optimal variational bound $\mathcal{F}_{\mathrm{vfe}}(\theta) = \mathrm{max}_{q(\uvec)}\mathcal{F}_{\mathrm{vfe}}(q(\uvec), \theta)$ (also called the `collapsed' bound) are available:
\begin{align}
p(f|\yvec, \theta) & \approx q_\mathrm{vfe}(f) \propto p(f_{\neq \uvec}|\uvec, \theta) p(\uvec | \theta) \norm(\yvec; \kfu \kuu^{-1} \uvec, \sigma_y^2\mathrm{I}), \nonumber\\
 \log p(\yvec|\theta) &\approx \mathcal{F}_{\mathrm{vfe}}(\theta) = \log \norm (\yvec; \zero, \kfu\kuu^{-1}\kuf + \sigma_y^2\mathrm{I}) - \frac{1}{2\sigma^2_y} \sum_n (k_{nn} - \mathbf{K}_{n\uvec}\kuu^{-1}\mathbf{K}_{\uvec n}), \nonumber
\end{align}
where $\fvec$ is the latent function values at training points, and $\mathbf{K}_{\mathbf{f}_1 \mathbf{f}_2}$ is the covariance matrix between the latent function values $\mathbf{f}_1$ and $\mathbf{f}_2$.
Critically, the approach leads to $O(NM^2)$ complexity for approximate maximum likelihood learning and $O(M^2)$ per test point for prediction. In order for this method to perform well, it is necessary to adapt the pseudo-point input locations, e.g.~by optimising the variational free energy, so that the pseudo-data distribute themselves over the training data. 

Alternatively, stochastic optimisation may be applied directly to the original, {\it uncollapsed} version of the bound \cite{hensman2013gaussian,cheng2016incremental}.
In particular, an unbiased estimate of the variational lower bound can be obtained using a small number of training points randomly drawn from the training set:
\begin{align}
\mathcal{F}_{\mathrm{vfe}}(q(\uvec), \theta) &\approx - \mathrm{KL}[q(\uvec)||p(\uvec | \theta)] + \frac{N}{|B|} \sum_{y_n \in B} \int \dd \uvec \; q(\uvec) p(f_n|\uvec, \theta) \log p(y_n|f_n, \theta). \nonumber
\end{align}
Since the optimal approximation is Gaussian as shown above, $q(\uvec)$ is often posited as a Gaussian distribution and its parameters are updated by following the (noisy) gradients of the stochastic estimate of the variational lower bound. By passing through the training set a sufficient number of times, the variational distribution converges to the optimal solution above, given appropriately decaying learning rates \cite{hensman2013gaussian}.

In principle, the stochastic uncollapsed approach is applicable to the streaming setting as it refines an approximate posterior based on mini-batches of data that can be considered to arrive sequentially (here $N$ would be the number of data points seen so far).  
However, it is unsuited to this task since stochastic optimisation assumes that the data subsampling process is {\it uniformly random}, that the training set is revisited multiple times, and it typically makes a single gradient update per mini-batch.
These assumptions are incompatible with the streaming setting: continuously arriving data are not typically drawn iid from the input distribution (consider an evolving time-series, for example); the data can only be touched once by the algorithm and not revisited due to computational constraints; each mini-batch needs to be processed intensively as it will not be revisited (multiple gradient steps would normally be required, for example, and this runs the risk of forgetting old data without delicately tuning the learning rates). 
In the following sections, we shall discuss how to tackle these challenges through a novel online inference and learning procedure, and demonstrate the efficacy of this method over the uncollapsed approach and na\"ive online versions of the collapsed approach.

\section{Streaming sparse GP (SSGP) approximation using variational inference}
\label{sec:online_vfe}
The general situation assumed in this paper is that data arrive sequentially so that at each step new data points $\yvec_\mathrm{new}$ are added to the old dataset $\yvec_\mathrm{old}$. The goal is to approximate the marginal likelihood and the posterior of the latent process at each step, which can be used for anytime prediction. The hyperparameters will also be adjusted online. 
Importantly, we assume that we can only access the current data points $\yvec_\mathrm{new}$ directly for computational reasons (it might be too expensive to hold $\yvec_\mathrm{old}$ and $\xvec_{1:N_\mathrm{old}}$ in memory, for example, or approximations made at the previous step must be reused to reduce computational overhead). So the effect of the old data on the current posterior must be propagated through the previous posterior.
We will now develop a new sparse variational free energy approximation for this purpose, that compactly summarises the old data via pseudo-points. The pseudo-inputs will also be adjusted online since this is critical as new parts of the input space will be revealed over time. The framework is easily extensible to more complex non-linear models. 
\subsection{Online variational free energy inference and learning}
\begin{figure}
\centering
\begin{minipage}{.5\textwidth}
  \centering
  \includegraphics[width=\linewidth]{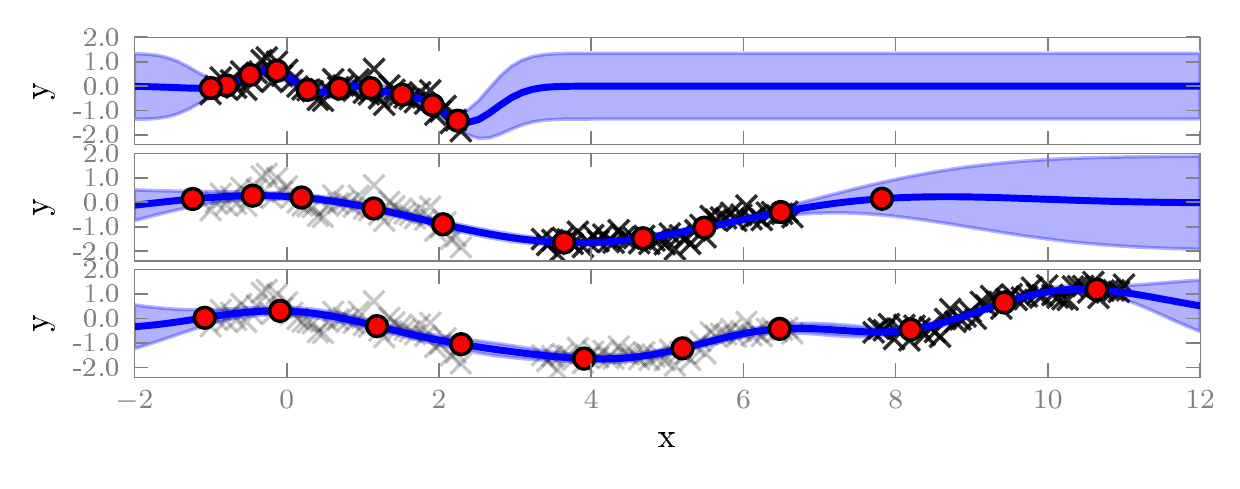}
\end{minipage}%
\begin{minipage}{.5\textwidth}
  \centering
  \includegraphics[width=\linewidth]{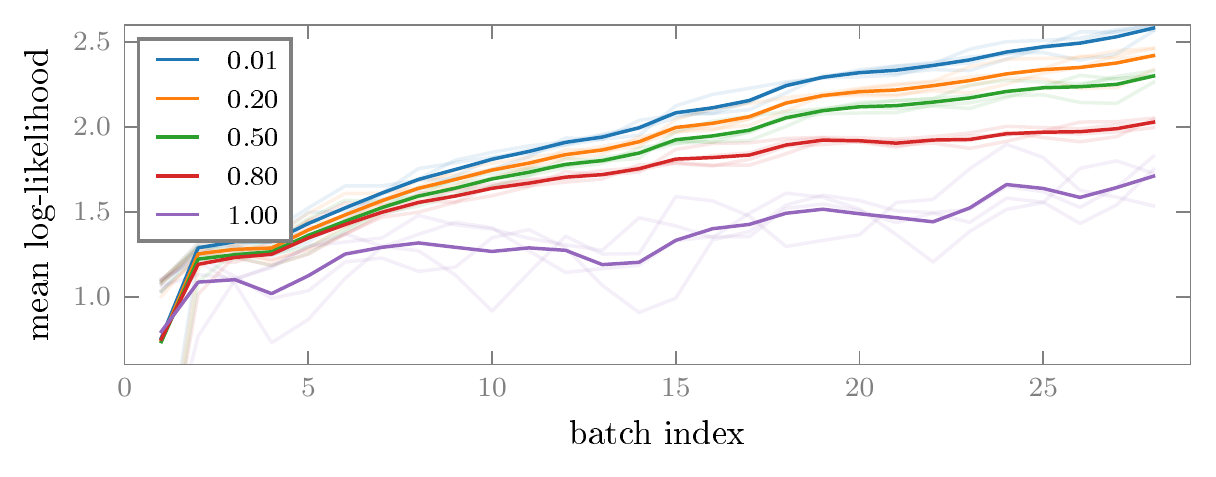}
\end{minipage}
\vspace{-9pt}
\caption{[Left] SSGP inference and learning on a toy time-series using the VFE approach. The black crosses are data points (past points are greyed out), the red circles are pseudo-points, and blue lines and shaded areas are the marginal predictive means and confidence intervals at test points. [Right] Log-likelihood of test data as training data arrives for different $\alpha$ values, for the pseudo periodic dataset (see \cref{sec:reg_real}). We observed that $\alpha=0.01$ is virtually identical to VFE. Dark lines are means over 4 splits and shaded lines are results for each split. Best viewed in colour.\label{fig:vfe_and_alpha}}
\vspace{-8pt}
\end{figure}

Consider an approximation to the true posterior at the previous step, $q_\mathrm{old}(f)$, which must be updated to form the new approximation $q_\mathrm{new}(f)$,
\begin{align}
q_\mathrm{old}(f) &\approx p(f|\yvec_\mathrm{old}) = \frac{1}{\mathcal{Z}_1(\theta_\mathrm{old})} p(f|\theta_\mathrm{old}) p(\yvec_\mathrm{old}|f), \label{eqn:vfe_old}\\
q_\mathrm{new}(f) &\approx p(f|\yvec_\mathrm{old}, \yvec_\mathrm{new}) = \frac{1}{\mathcal{Z}_2(\theta_\mathrm{new})} p(f|\theta_\mathrm{new}) p(\yvec_\mathrm{old}|f) p(\yvec_\mathrm{new}|f)\label{eqn:vfe_new}.
\end{align}
Whilst the updated exact posterior $p(f|\yvec_\mathrm{old}, \yvec_\mathrm{new})$ balances the contribution of old and new data through their likelihoods, the new approximation cannot access $p(\yvec_\mathrm{old}|f)$ directly. 
Instead, we can find an approximation of $p(\yvec_\mathrm{old}|f)$ by inverting \cref{eqn:vfe_old}, that is ${ p(\yvec_\mathrm{old}|f) \approx \mathcal{Z}_1(\theta_\mathrm{old}) q_\mathrm{old}(f)/p(f|\theta_\mathrm{old}) }$. 
Substituting this into \cref{eqn:vfe_new} yields,
\begin{align}
\hat{p}(f|\yvec_\mathrm{old}, \yvec_\mathrm{new}) = \frac{\mathcal{Z}_1(\theta_\mathrm{old})}{\mathcal{Z}_2(\theta_\mathrm{new})} p(f|\theta_\mathrm{new}) p(\yvec_\mathrm{new}|f)  \frac{q_\mathrm{old}(f)}{p(f|\theta_\mathrm{old})}.
\end{align}
Although it is tempting to use this as the new posterior, $q_\mathrm{new}(f) = \hat{p}(f|\yvec_\mathrm{old}, \yvec_\mathrm{new})$, this recovers exact GP regression with fixed hyperparameters (see \cref{sec:fixed}) and it is intractable. So, instead, we consider a variational update that projects the distribution back to a tractable form using pseudo-data.
At this stage we allow the pseudo-data input locations in the new approximation to differ from those in the old one. This is required if new regions of input space are gradually revealed, as for example in typical time-series applications. Let $\avec = f(\zvec_{\mathrm{old}})$ and $\bvec = f(\zvec_{\mathrm{new}})$ be the function values at the pseudo-inputs before and after seeing new data. 
Note that the number of pseudo-points, $M_\avec = |\avec|$ and $M_\bvec = |\bvec|$ are not necessarily restricted to be the same.
The form of the approximate posterior mirrors that in the batch case, that is, the previous approximate posterior, $q_\mathrm{old}(f) = p(f_{\ne \avec}|\avec, \theta_\mathrm{old}) q_\mathrm{old}(\avec)$ where we assume $q_\mathrm{old}(\avec) = \norm(\avec; \ma, \Sa)$.
The new posterior approximation takes the same form, but with the new pseudo-points and new hyperparameters: $q_\mathrm{new}(f) = p(f_{\ne \bvec}|\bvec, \theta_\mathrm{new}) q_\mathrm{new}(\bvec)$. 
Similar to the batch case, this approximate inference problem can be turned into an optimisation problem using variational inference. Specifically, consider
\begin{align}
\mathrm{KL} \lbrack q_\mathrm{new}(f) || \hat{p}(f|\yvec_\mathrm{old}, \yvec_\mathrm{new}) \rbrack 
&= \int \dd f \; q_\mathrm{new}(f) \log \frac{p(f_{\ne \bvec}|\bvec, \theta_\mathrm{new}) q_\mathrm{new}(\bvec)}{\frac{\mathcal{Z}_1(\theta_\mathrm{old})}{\mathcal{Z}_2(\theta_\mathrm{new})} p(f|\theta_\mathrm{new}) p(\yvec_\mathrm{new}|f) \frac{q_\mathrm{old}(f)}{p(f|\theta_\mathrm{old})}} \label{eqn:kl}\\
&= \log\frac{\mathcal{Z}_2(\theta_\mathrm{new})}{\mathcal{Z}_1(\theta_\mathrm{old})} + \int \dd f \; q_\mathrm{new}(f) \left[ \log \frac{p(\avec|\theta_\mathrm{old})q_\mathrm{new}(\bvec)}{p(\bvec|\theta_\mathrm{new}) q_\mathrm{old}(\avec) p(\yvec_\mathrm{new}|f)} \right]. \nonumber
\end{align}
Since the KL divergence is non-negative, the second term in the expression above is the negative approximate lower bound of the online log marginal likelihood (as $\mathcal{Z}_2 / \mathcal{Z}_1 \approx p(\yvec_\mathrm{new}|\yvec_\mathrm{old})$), or the variational free energy $\mathcal{F}(q_\mathrm{new}(f), \theta_\mathrm{new})$.
By setting the derivative of $\mathcal{F}$ w.r.t. $q(\bvec)$ equal to 0, the optimal approximate posterior can be obtained for the regression case,\footnote{Note that we have dropped $\theta_\mathrm{new}$ from $p(\bvec|\theta_\mathrm{new})$, $p(\avec|\bvec, \theta_\mathrm{new})$ and $p(\fvec|\bvec, \theta_\mathrm{new})$ to lighten the notation.}
\begin{align}
q_\mathrm{vfe}(\bvec) 
	&\propto p(\bvec) \exp \Big( \int \dd \avec \; p(\avec|\bvec) \log \frac{q_\mathrm{old}(\avec)}{p(\avec|\theta_\mathrm{old})} + \int \dd \fvec \; p(\fvec|\bvec) \log p(\yvec_\mathrm{new}|\fvec) \Big) \\
	&\propto p(\bvec) \norm(\hat{\yvec}; \khatb \kbb^{-1} \bvec, \Sigma_{\hat{\yvec}, \mathrm{vfe}}), \label{eqn:vfe_opt}
\end{align}
where $\fvec$ is the latent function values at the new training points,
\begin{align}
\hat{\yvec} = \begin{bmatrix} \yvec_\mathrm{new} \\ \Da \Sa^{-1} \ma \end{bmatrix}, \; \khatb = \begin{bmatrix} \kfb \\ \kab \end{bmatrix}, \; \Sigma_{\hat{\yvec}, \mathrm{vfe}} = \begin{bmatrix} \sigma_y^2\mathrm{I} & \zero \\ \zero & \Da \end{bmatrix}, \; \Da = (\Sa^{-1} - \kaa'^{-1})^{-1}. \nonumber
\end{align} 
The negative variational free energy is also analytically available,
\begin{align}
\mathcal{F}(\theta) &= \log \norm (\hat{\yvec}; \zero, \khatb\kbb^{-1}\kbhat + \Sigma_{\hat{\yvec}, \mathrm{vfe}}) - \frac{1}{2\sigma_y^2} \mathrm{tr} (\kff - \kfb\kbb^{-1} \kbf) + \Delta_\avec \text{; where}\label{eqn:vfe_energy}\\
2\Delta_\avec &= - \log \vert \Sa \vert + \log \vert \kaa' \vert + \log \vert \Da \vert + \ma^\intercal (\Sa^{-1}\Da\Sa^{-1} - \Sa^{-1} )\ma - \mathrm{tr} \lbrack \Da^{-1} \Qa \rbrack + \mathrm{const.} \nonumber
\end{align}
\Cref{eqn:vfe_opt,eqn:vfe_energy} provide the complete recipe for online posterior update and hyperparameter learning in the streaming setting.  The computational complexity and memory overhead of the new method is of the same order as the uncollapsed stochastic variational inference approach.
The procedure is demonstrated on a toy regression example as shown in \cref{fig:vfe_and_alpha}[Left].
%
\subsection{Online $\alpha$-divergence inference and learning}
One obvious extension of the online approach discussed above replaces the KL divergence  in \cref{eqn:kl} with a more general $\alpha$-divergence \cite{minka2004pep}. This does not affect tractability: the optimal form of the approximate posterior can be obtained analytically for the regression case, ${ q_\mathrm{pep}(\bvec) \propto p(\bvec) \norm(\hat{\yvec}; \khatb \kbb^{-1} \bvec, \Sigma_{\hat{\yvec}, \mathrm{pep}}) }$ where
\begin{align}
\Sigma_{\hat{\yvec}, \mathrm{pep}} = \begin{bmatrix} \sigma_y^2\mathrm{I} + \alpha \mathrm{diag}(\kff - \kfb\kbb^{-1} \kbf) & \zero \\ \zero & \Da + \alpha (\kaa - \kab\kbb^{-1} \kba) \label{eqn:pep_opt}\end{bmatrix}.
\end{align}
This reduces back to the variational case as $\alpha \rightarrow 0$ (compare to \cref{eqn:vfe_opt}) since then the $\alpha$-divergence is equivalent to the KL divergence.  The approximate online log marginal likelihood is also analytically tractable and recovers the variational case when $\alpha \to 0$. Full details are provided in the appendix.

\subsection{Connections to previous work and special cases}
\label{sec:fixed}

This section briefly highlights connections between the new framework and existing approaches including Power Expectation Propagation (Power-EP), Expectation Propagation (EP), Assumed Density Filtering (ADF), and streaming variational Bayes.

Recent work has unified a range of batch sparse GP approximations as special cases of the Power-EP algorithm \cite{BuiYanTur17}. The online $\alpha$-divergence approach to inference and learning described in the last section is equivalent to running a forward filtering pass of Power-EP. In other words, the current work generalizes the unifying framework to the streaming setting.

When the hyperparameters and the pseudo-inputs are fixed, $\alpha$-divergence inference for sparse GP regression recovers the batch solutions provided by Power-EP. In other words, only a single pass through the data is necessary for Power-EP to converge in sparse GP regression. 
For the case $\alpha = 1 $, which is called Expectation Propagation, we recover the seminal work by Csat\'{o} and Opper \cite{csato+opper:2002}. For the variational free energy case (equivalently where $\alpha \rightarrow 0 $) we recover the seminal work by Csat\'{o} \cite{csato:2002}. The new framework can be seen to extend these methods to allow principled learning and pseudo-input optimisation. Interestingly, in the setting where hyperparameters and the pseudo-inputs are fixed, if pseudo-points are added at each stage at the new data input locations, the method returns the true posterior and marginal likelihood (see appendix).

For fixed hyperparameters and pseudo-points, the new VFE framework is equivalent to the application of streaming variational Bayes (VB) or online variational inference \cite{broderick2013streaming,ghahramani2000online,sato2001online} to the GP setting in which the previous posterior plays a role of an effective prior for the new data.  Similarly, the equivalent algorithm when $\alpha = 1$ is called Assumed Density Filtering \cite{opper1999bayesian}. When the hyperparameters are updated, the new method proposed here is different from streaming VB and standard application of ADF, as the new method propagates approximations to just the old likelihood terms and not the prior. Importantly, we found vanilla application of the streaming VB framework performed catastrophically for hyperparameter learning, so the modification is critical. 


\section{Experiments}
\label{sec:exp}

In this section, the SSGP method is evaluated in terms of speed, memory usage, and accuracy (log-likelihood and error). The method was implemented on GPflow \cite{gpflow} and compared against GPflow's version of the following baselines: exact GP (GP), sparse GP using the collapsed bound (SGP), and stochastic variational inference using the uncollapsed bound (SVI).
In all the experiments, the RBF kernel with ARD lengthscales is used, but this is not a limitation required by the new methods. An implementation of the proposed method can be found at \href{http://github.com/thangbui/streaming_sparse_gp}{http://github.com/thangbui/streaming\_sparse\_gp}. Full experimental results and additional discussion points are included in the appendix.

\subsection{Synthetic data}
\textbf{Comparing $\alpha$-divergences}. We first consider the general online $\alpha$-divergence inference and learning framework and compare the performance of different $\alpha$ values on a toy online regression dataset in \cref{fig:vfe_and_alpha}[Right].
Whilst the variational approach performs well, adapting pseudo-inputs to cover new regions of input space as they are revealed, algorithms using higher $\alpha$ values perform more poorly.  Interestingly this appears to be related to the tendency for EP, in batch settings, to clump pseudo-inputs on top of one another \cite{bauer2016nips}. Here the effect is much more extreme as the clumps accumulate over time, leading to a shortage of pseudo-points if the input range of the data increases. Although heuristics could be introduced to break up the clumps, this result suggests that using small $\alpha$ values for online inference and learning might be more appropriate (this recommendation differs from the batch setting where intermediate settings of $\alpha$ around 0.5 are best \cite{BuiYanTur17}). Due to these findings, for the rest of the paper, we focus on the variational case.  

\textbf{Hyperparameter learning}. We generated multiple time-series from GPs with known hyperparameters and observation noises, and tracked the hyperparameters learnt by the proposed online variational free energy method and exact GP regression. Overall, SSGP can track and learn good hyperparameters, and if there are sufficient pseudo-points, it performs comparatively to full GP on the entire dataset. Interestingly, all models including full GP regression tend to learn bigger noise variances as any discrepancy in the true and learned function values is absorbed into this parameter.

\subsection{Speed versus accuracy}
\label{sec:reg_real}
In this experiment, we compare SSGP to the baselines (GP, SGP, and SVI) in terms of a speed-accuracy trade-off where the mean marginal log-likelihood (MLL) and the root mean squared error (RMSE) are plotted against the accumulated running time of each method after each iteration. The comparison is performed on two time-series datasets and a spatial dataset.

\begin{figure}[!ht]
\begin{center}
    \begin{subfigure}[t]{\textwidth}
    \centering
    \includegraphics[width=6cm]{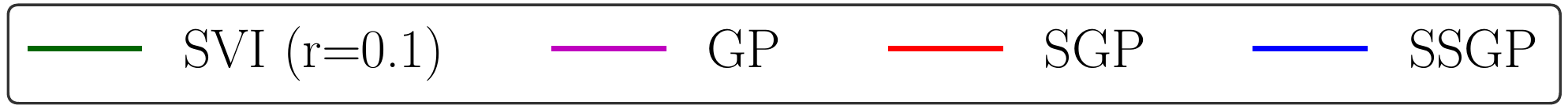}
    \end{subfigure}
    
    \begin{subfigure}[t]{0.48\textwidth}
    \centering
    \includegraphics[width=6.7cm]{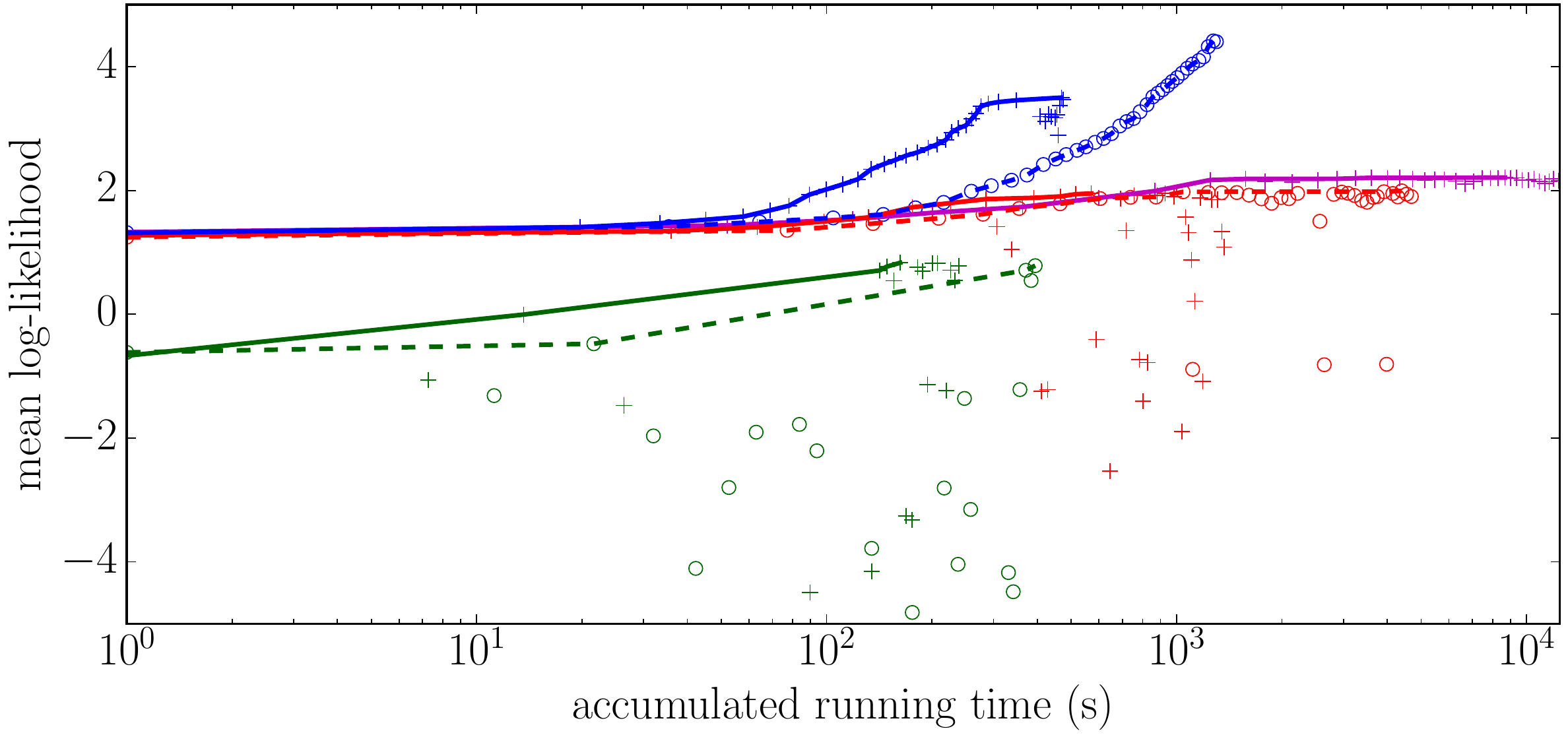}
    \end{subfigure}
    \hspace{2mm}
    \begin{subfigure}[t]{0.48\textwidth}
    \centering
    \includegraphics[width=6.7cm]{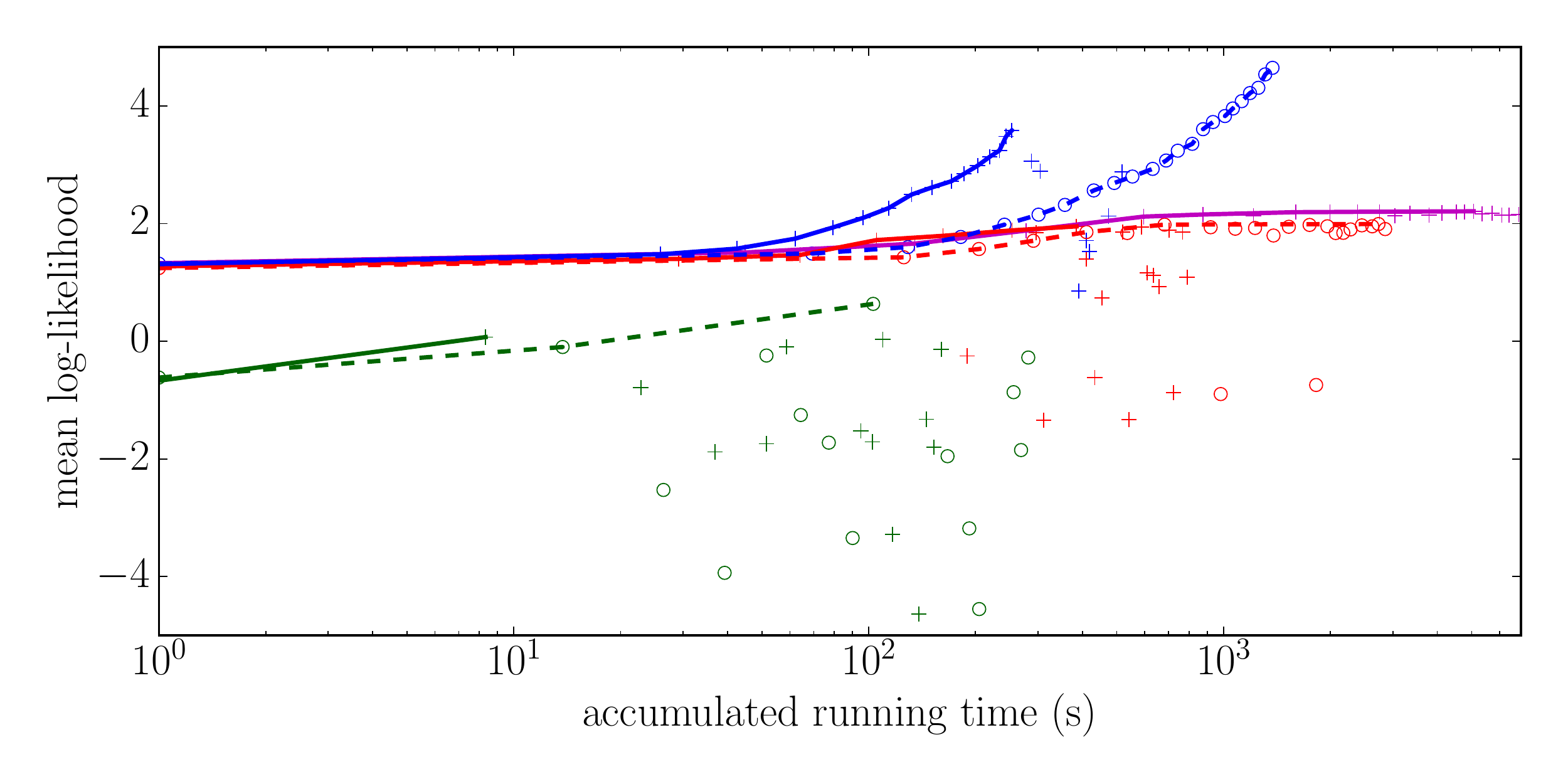}
    \end{subfigure}
    
    \begin{subfigure}[t]{.48\textwidth}
    \centering
    \includegraphics[width=6.7cm]{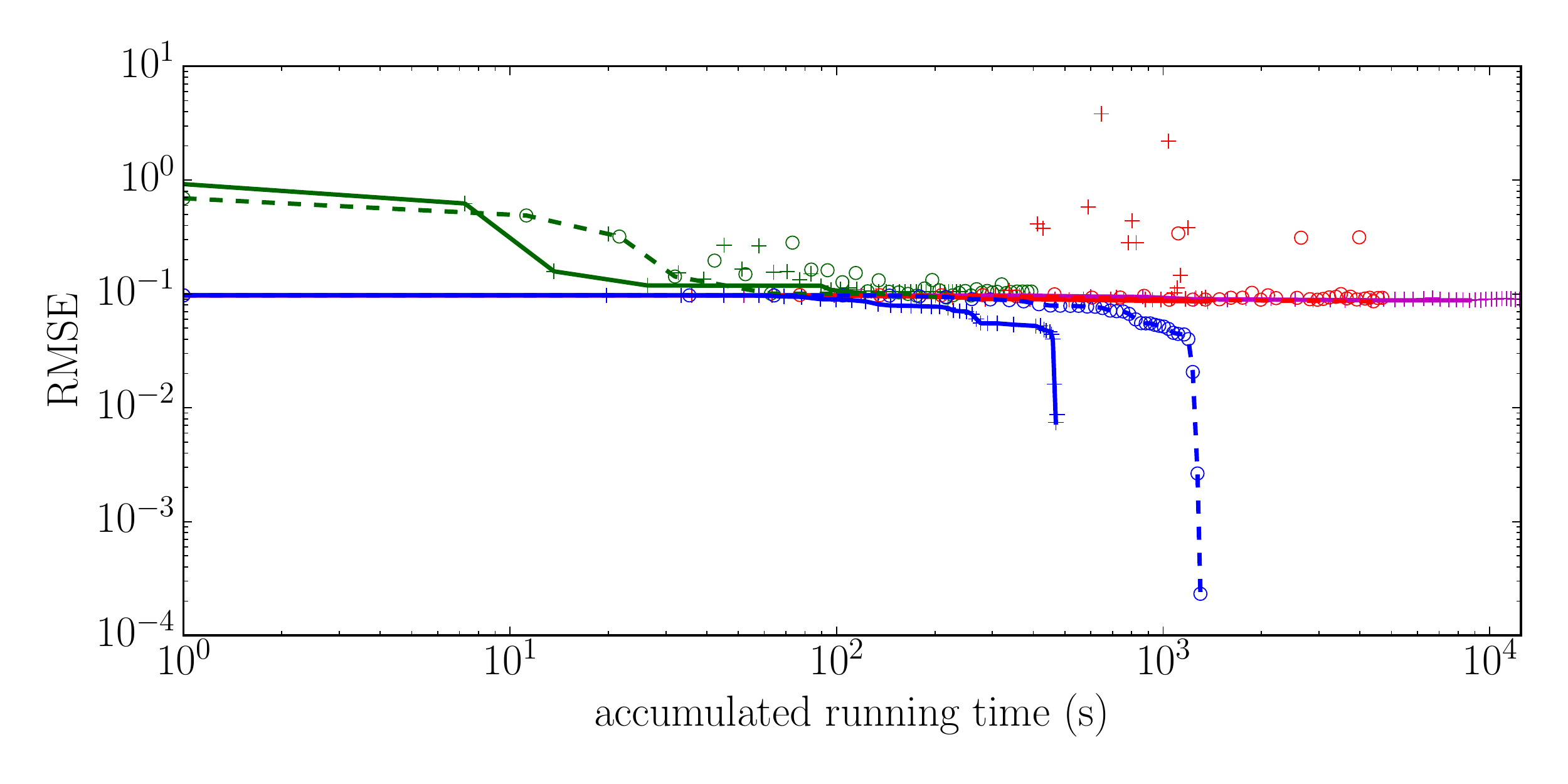}
    \caption*{\qquad pseudo periodic data, batch size = $300$}
    \end{subfigure}
    \hspace{2mm}
    \begin{subfigure}[t]{.48\textwidth}
    \centering
    \includegraphics[width=6.7cm]{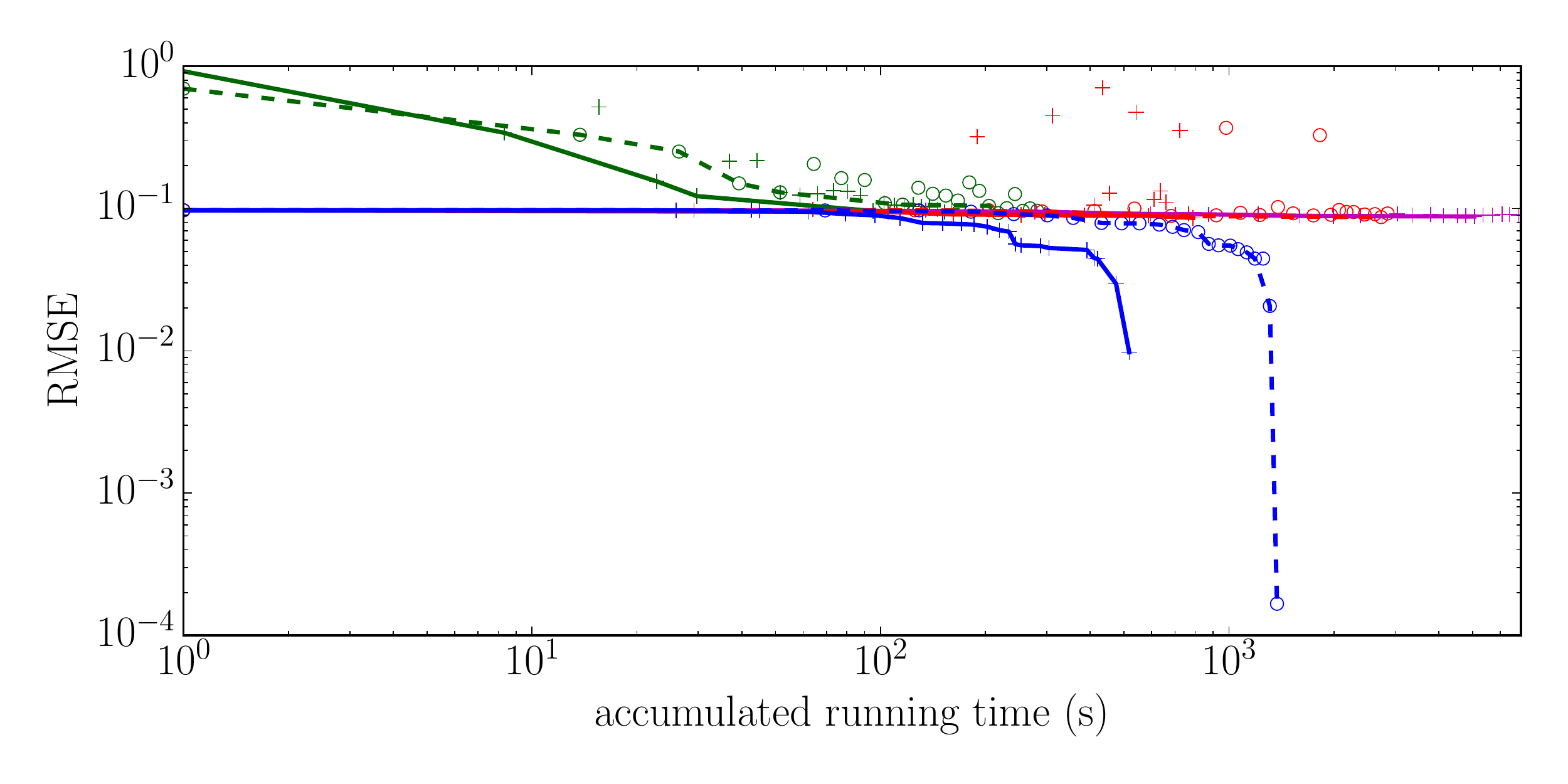}
    \caption*{\qquad pseudo periodic data, batch size = $500$}
    \end{subfigure}
    
    {\vskip 0.1cm}
    \begin{subfigure}[t]{.48\textwidth}
    \centering
    \includegraphics[width=6.7cm]{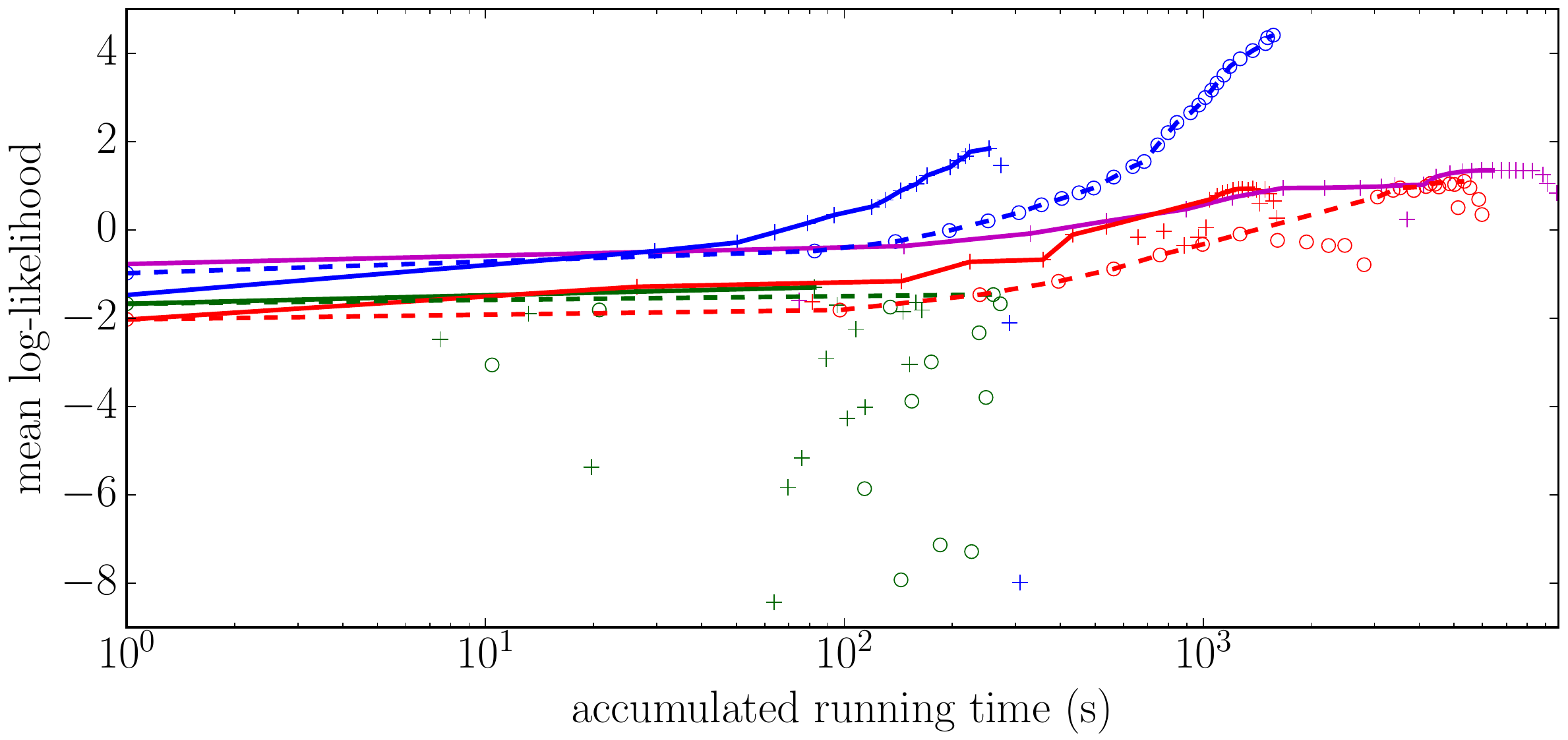}
    \end{subfigure}
    \hspace{2mm}
    \begin{subfigure}[t]{.48\textwidth}
    \centering
    \includegraphics[width=6.7cm]{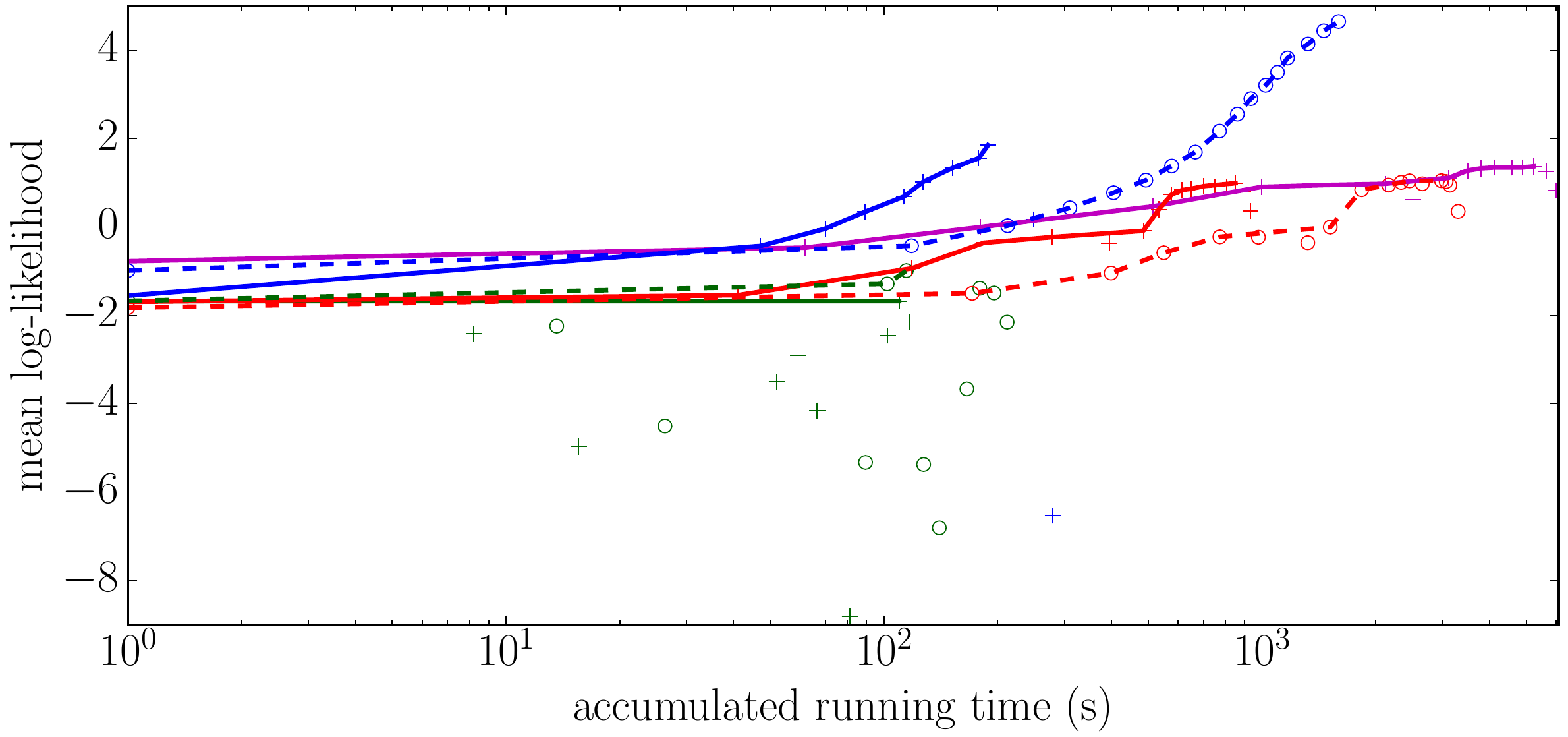}
    \end{subfigure}

    \begin{subfigure}[t]{.48\textwidth}
    \centering
    \includegraphics[width=6.7cm]{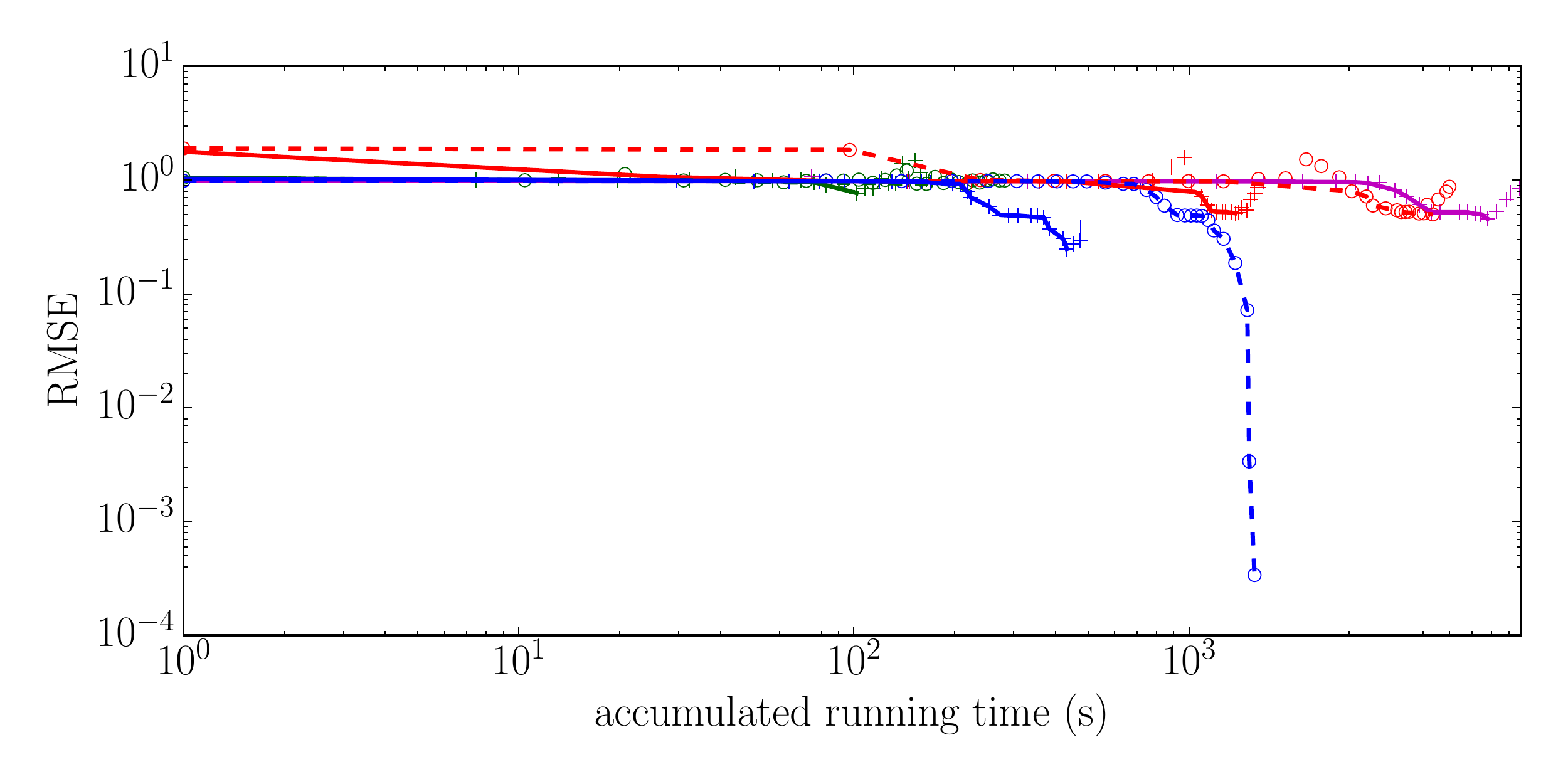}
    \caption*{\qquad audio data, batch size = $300$}
    \end{subfigure}
    \hspace{2mm}
    \begin{subfigure}[t]{.48\textwidth}
    \centering
    \includegraphics[width=6.7cm]{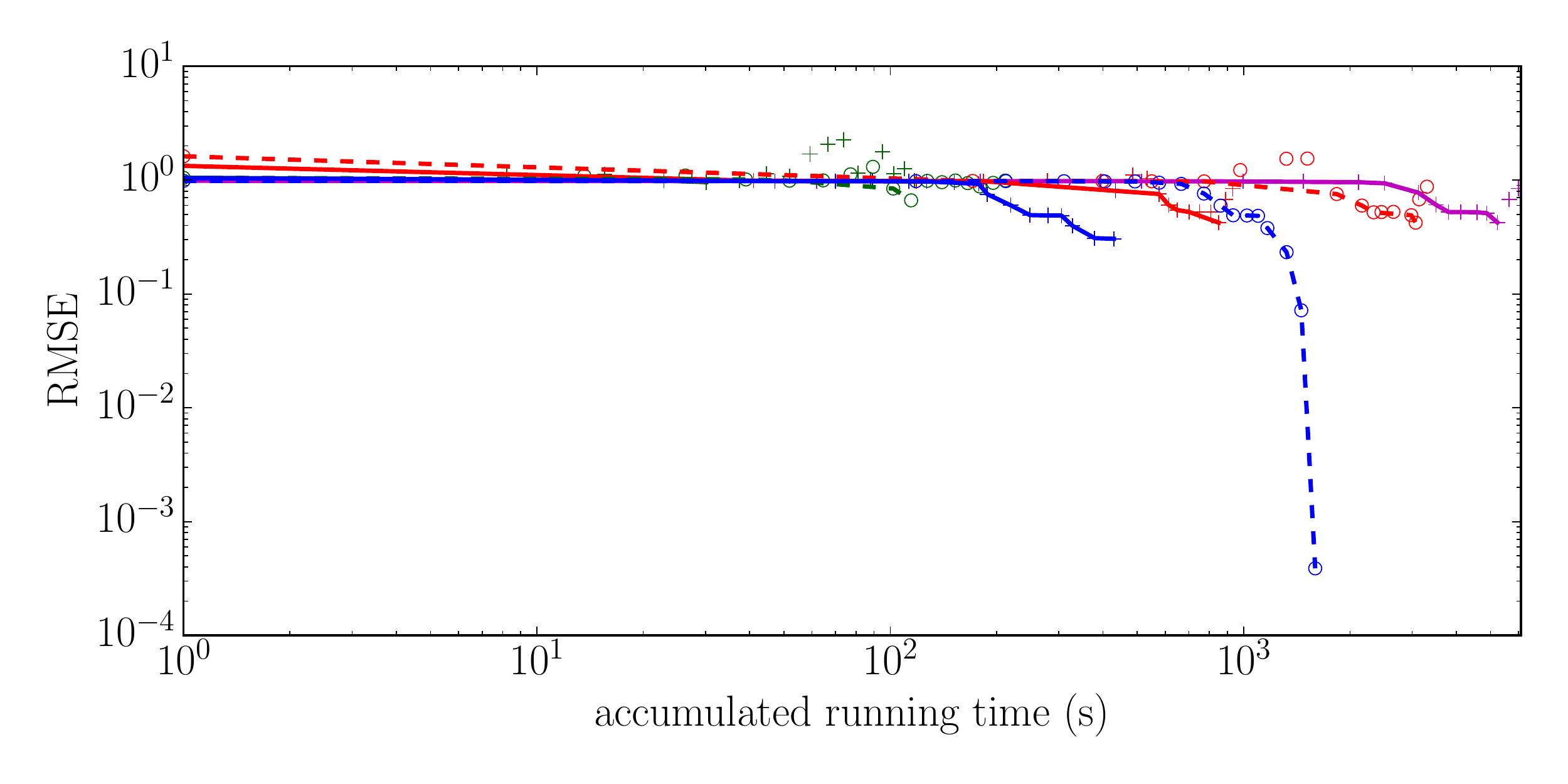}
    \caption*{\qquad audio data, batch size = $500$}
    \end{subfigure}
\caption{Results for time-series datasets with batch sizes $300$ and $500$. Pluses and circles indicate the results for $M = 100, 200$ pseudo-points respectively. For each algorithm (except for GP), the solid and dashed lines are the efficient frontier curves for $M=100, 200$ respectively.}
\label{fig:time-series}
\end{center}
\vspace{-15pt}
\end{figure}

\textbf{Time-series data}. We first consider modelling a segment of the pseudo periodic synthetic dataset \cite{keogh1999indexing},  previously used for testing indexing schemes in time-series databases. The segment contains 24,000 time-steps. Training and testing sets are chosen interleaved so that their sizes are both 12,000.
The second dataset is an audio signal prediction dataset, produced from the TIMIT database \cite{garofolo1993darpa} and previously used to evaluate GP approximations \cite{bui2014nips}.
The signal was shifted down to the baseband and a segment of length 18,000 was used to produce interleaved training and testing sets containing 9,000 time steps. For both datasets, we linearly scale the input time steps to the range $[0,10]$.

All algorithms are assessed in the mini-batch streaming setting with data $\mathbf{y}_{\mathrm{new}}$ arriving in batches of size 300 and 500 taken in order from the time-series.
The first 1,000 examples are used as an initial training set to obtain a reasonable starting model for each algorithm.
In this experiment, we use memory-limited versions of GP and SGP that store the last 3,000 examples.
This number was chosen so that the running times of these algorithms match those of SSGP or are slightly higher.
For all sparse methods (SSGP, SGP, and SVI), we run the experiments with 100 and 200 pseudo-points.

For SVI, we allow the algorithm to make 100 stochastic gradient updates during each iteration and run preliminary experiments to compare 3 learning rates $r = 0.001, 0.01$, and $0.1$. The preliminary results showed that the performance of SVI was not significantly altered and so we only present the results for $r = 0.1$.

Figure \ref{fig:time-series} shows the plots of the accumulated running time (total training and testing time up until the current iteration) against the MLL and RMSE for the considered algorithms. It is clear that SSGP significantly outperforms the other methods both in terms of the MLL and RMSE, once sufficient training data have arrived. The performance of SSGP improves when the number of pseudo-points increases, but the algorithm runs more slowly.
In contrast, the performance of GP and SGP, even after seeing more data or using more pseudo-points, does not increase significantly since they can only model a limited amount of data (the last 3,000 examples).

\begin{figure}[t]
\begin{center}
	\begin{subfigure}[t]{0.48\textwidth}
    \centering
	\includegraphics[width=6.3cm]{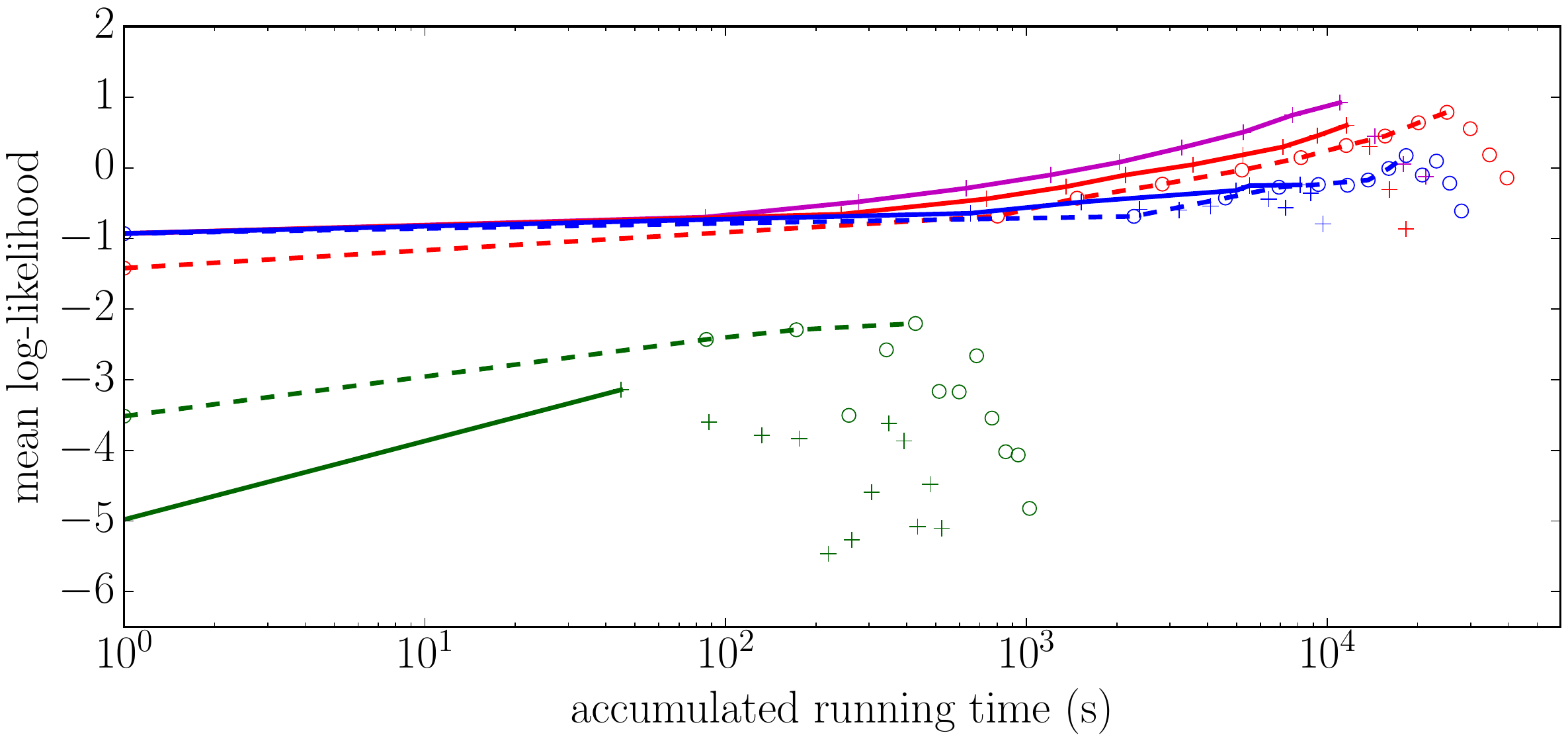}
	\end{subfigure}
	\begin{subfigure}[t]{0.48\textwidth}
    \centering
	\includegraphics[width=6.3cm]{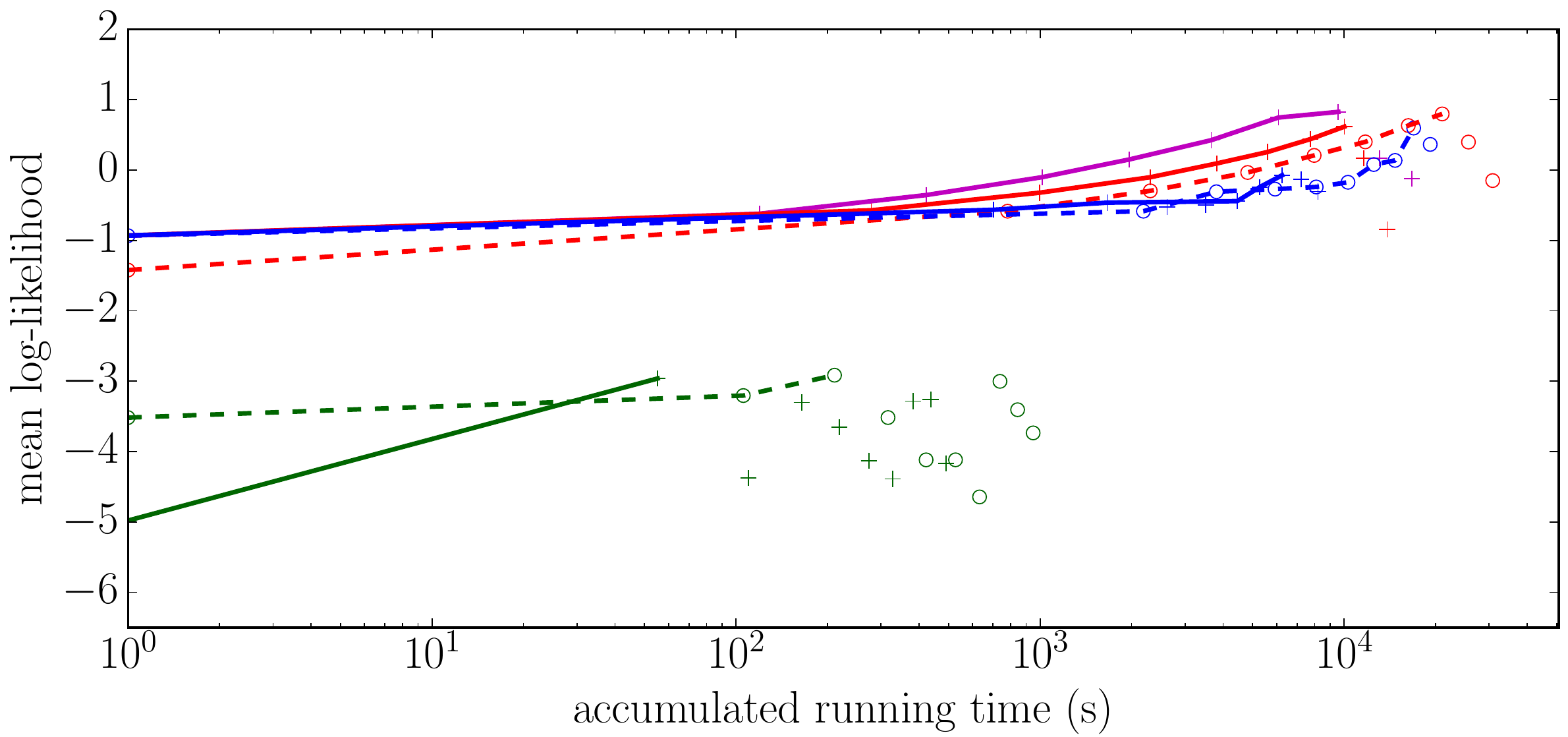}
	\end{subfigure}
    
    \begin{subfigure}[t]{0.48\textwidth}
    \centering
	\includegraphics[width=6.3cm]{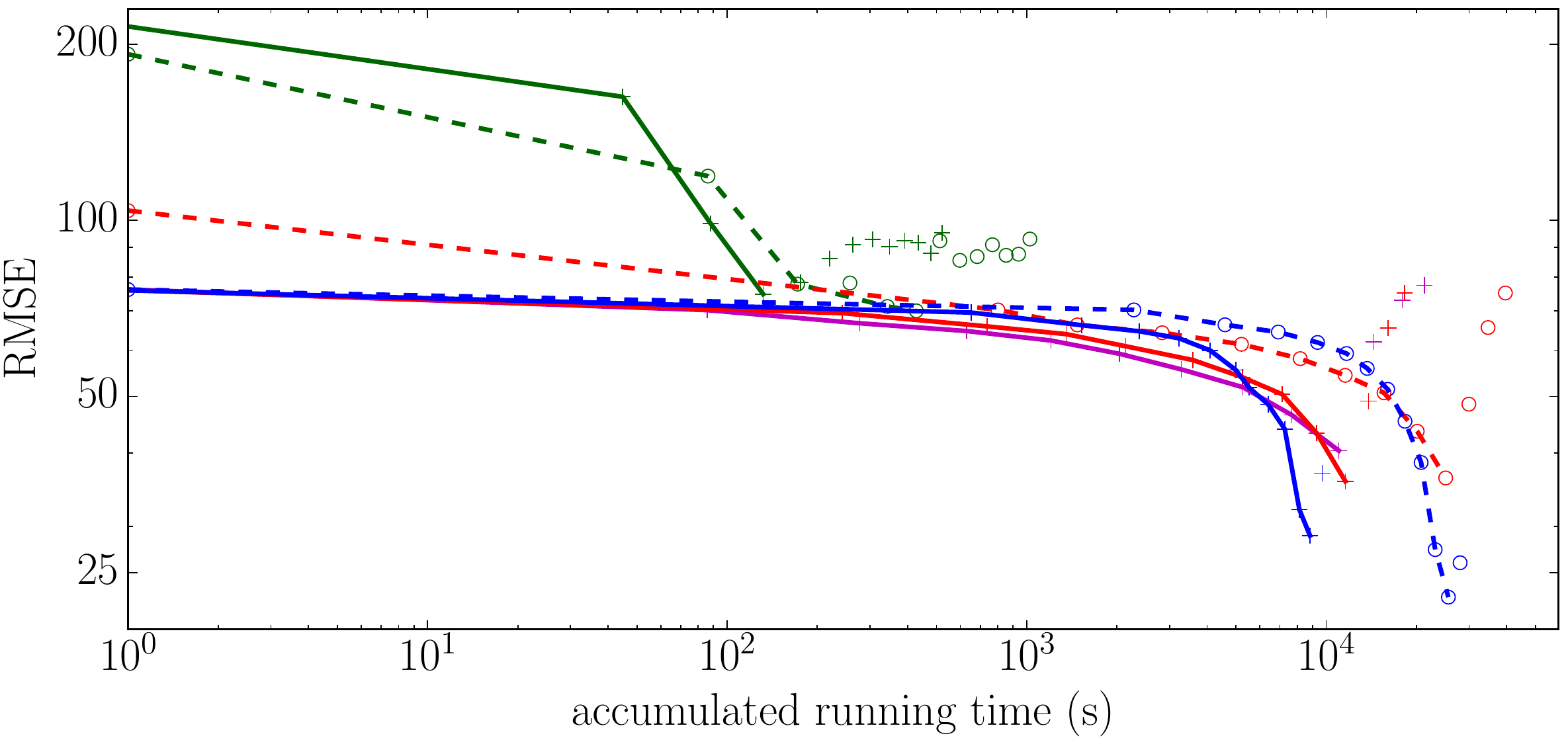}
    \caption*{\qquad terrain data, batch size = $750$}
    \end{subfigure}
    \begin{subfigure}[t]{0.48\textwidth}
    \centering
	\includegraphics[width=6.3cm]{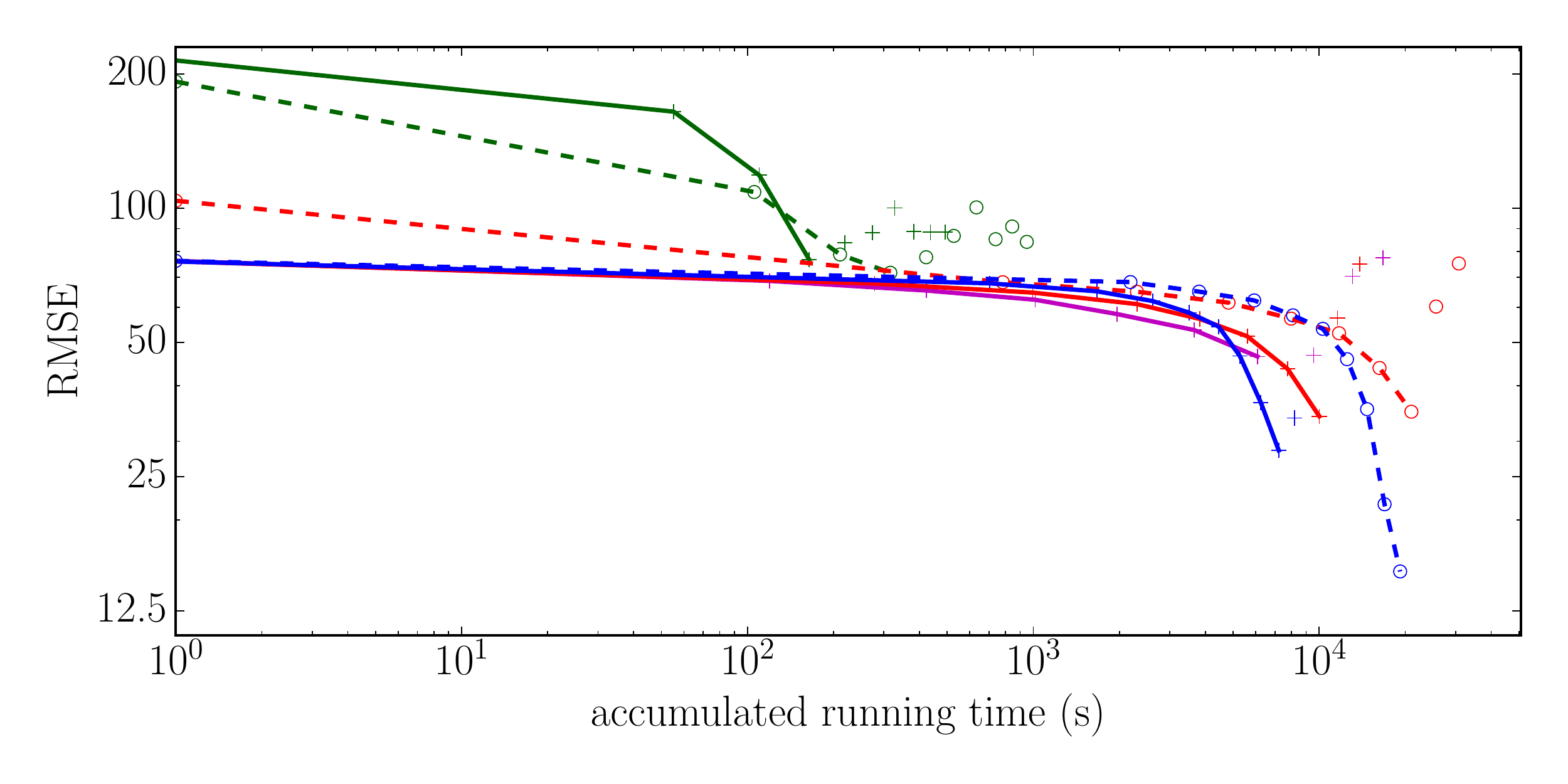}
    \caption*{\qquad terrain data, batch size = $1000$}
    \end{subfigure}
\caption{Results for spatial data (see \cref{fig:time-series} for the legend). Pluses/solid lines and circles/dashed lines indicate the results for $M = 400, 600$ pseudo-points respectively.}
\label{fig:spatial}
\end{center}
\end{figure}

\textbf{Spatial data}. The second set of experiments consider the OS Terrain 50 dataset that contains spot heights of landscapes in Great Britain computed on a grid.\footnote{The dataset is available at: \url{https://data.gov.uk/dataset/os-terrain-50-dtm}.}
A  block of $200 \times 200$ points was split into 10,000 training examples and 30,000 interleaved testing examples.
Mini-batches of data of size 750 and 1,000 arrive in spatial order. The first 1,000 examples were used as an initial training set. For this dataset, we allow GP and SGP to remember the last 7,500 examples and use 400 and 600 pseudo-points for the sparse models.
Figure \ref{fig:spatial} shows the results for this dataset. SSGP performs better than the other baselines in terms of the RMSE although it is worse than GP and SGP in terms of the MLL.

\subsection{Memory usage versus accuracy}

\begin{figure}[t]
\begin{center}
\includegraphics[width=6.3cm]{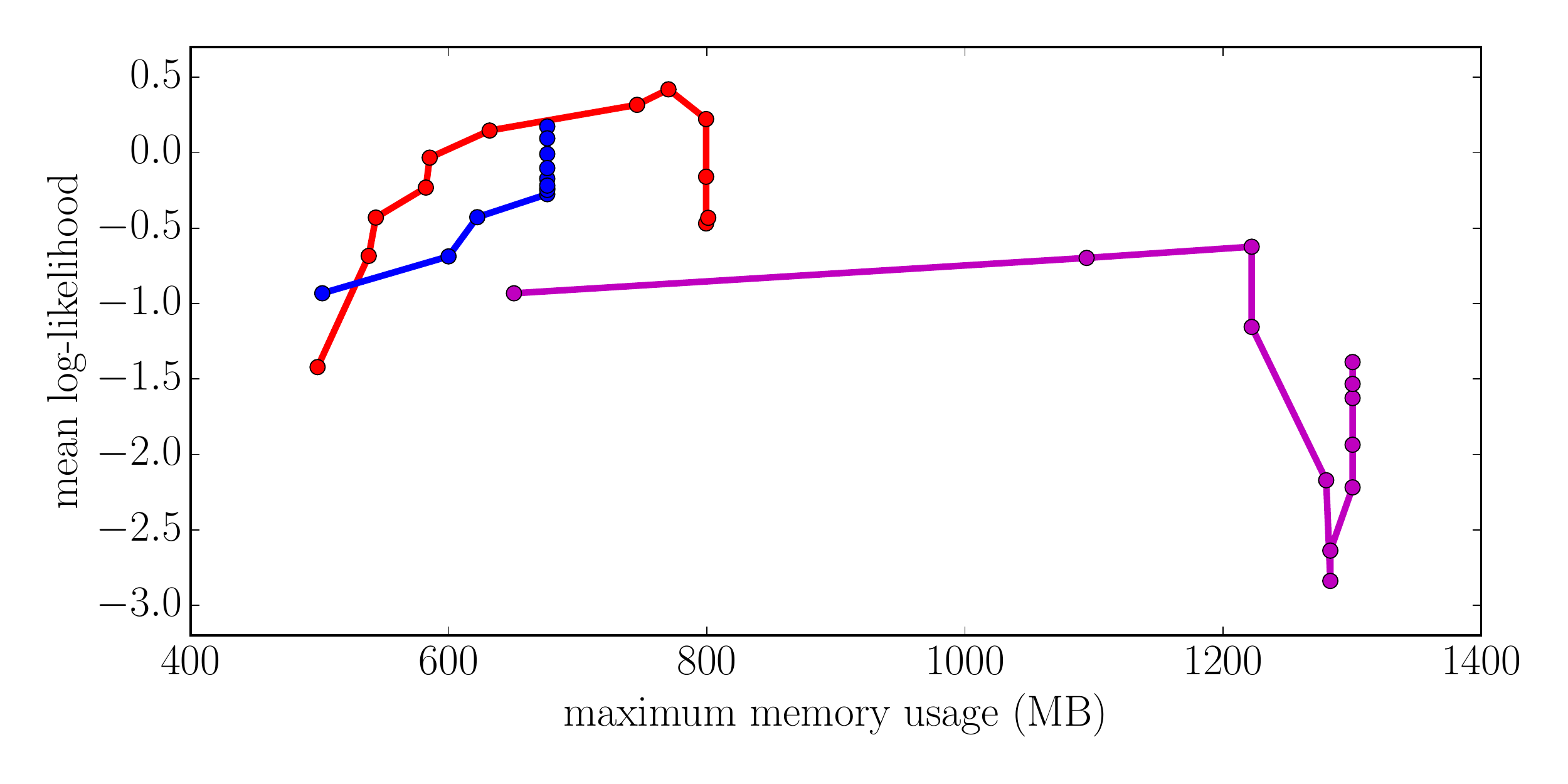}
\hspace{0.25cm}
\includegraphics[width=6.3cm]{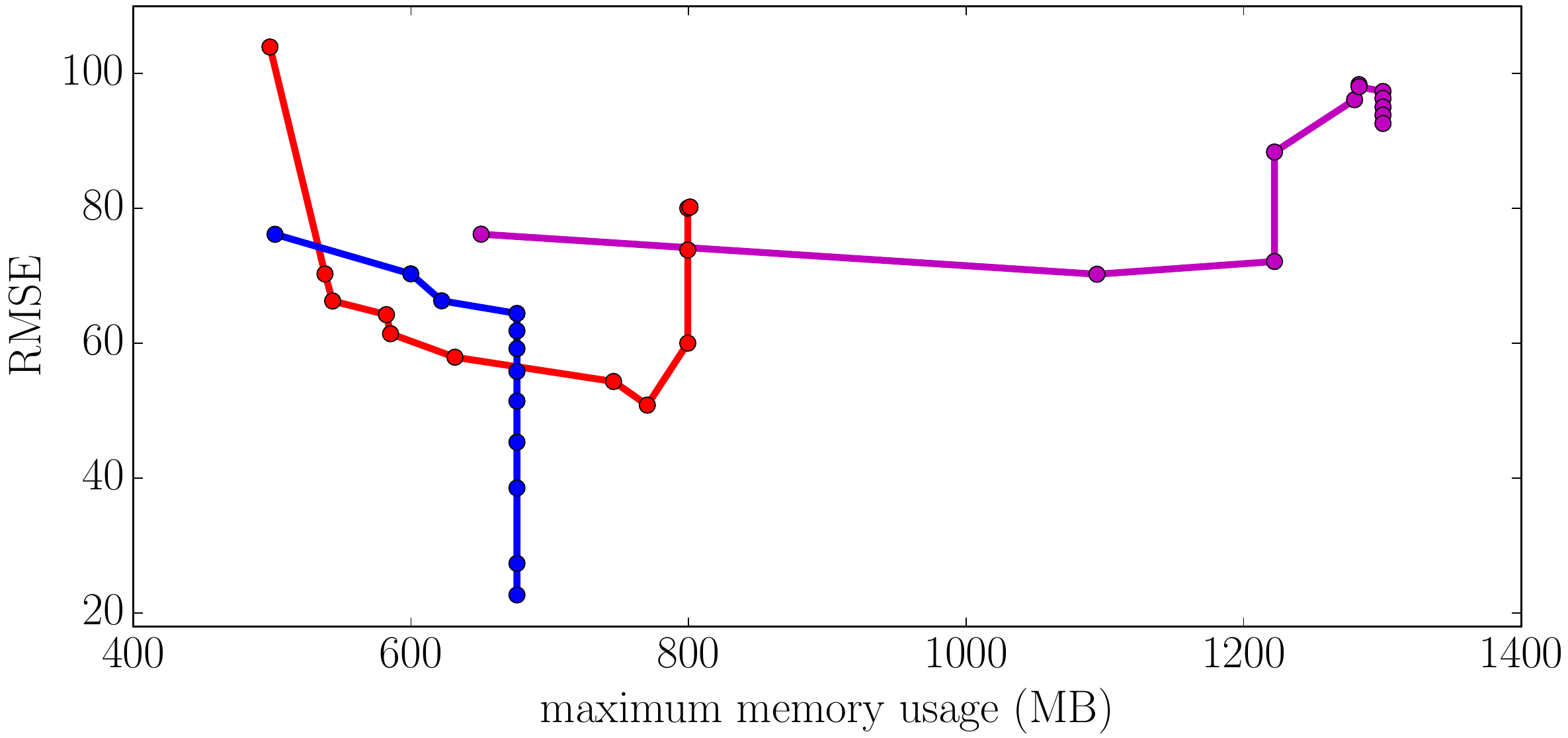}
\caption{Memory usage of SSGP (blue), GP (magenta) and SGP (red) against MLL and RMSE.}
\label{fig:memory}
\end{center}
\end{figure}

\begin{figure}[!ht]
\centering
\includegraphics[width=\textwidth]{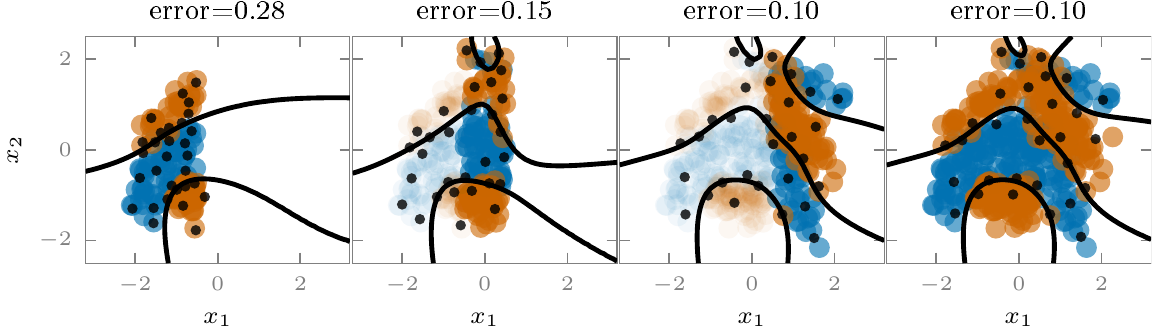}
\caption{SSGP inference and learning on a binary classification task in a non-iid streaming setting. The right-most plot shows the prediction made by using sparse variational inference on full training data \cite{HenMatGha15} for comparison. Past observations are greyed out. The pseudo-points are shown as black dots and the black curves show the decision boundary.\label{fig:binary_1}}
\end{figure}

Besides running time, memory usage is another important factor that should be considered.
In this experiment, we compare the memory usage of SSGP against GP and SGP on the Terrain dataset above with batch size 750 and $M=600$ pseudo-points.
We allow GP and SGP to use the last 2,000 and 6,000 examples for training, respectively.
These numbers were chosen so that the memory usage of the two baselines roughly matches that of SSGP.
Figure \ref{fig:memory} plots the maximum memory usage of the three methods against the MLL and RMSE.
From the figure, SSGP requires small memory usage while it can achieve comparable or better MLL and RMSE than GP and SGP.

\subsection{Binary classification}
We show a preliminary result for GP models with non-Gaussian likelihoods, in particular, a binary classification model on the benchmark {\it banana} dataset. As the optimal form for the approximate posterior is not analytically tractable, the uncollapsed variational free energy is optimised numerically. The predictions made by SSGP in a non-iid streaming setting are shown in \cref{fig:binary_1}. SSGP performs well and achieves the performance of the batch sparse variational method \cite{HenMatGha15}.

\section{Summary}
\label{sec:summary}
We have introduced a novel online inference and learning framework for Gaussian process models. The framework unifies disparate methods in the literature and greatly extends them, allowing sequential updates of the approximate posterior and online hyperparameter optimisation in a principled manner. The proposed approach outperforms existing approaches on a wide range of regression datasets and shows promising results on a binary classification dataset. A more thorough investigation on models with non-Gaussian likelihoods is left as future work. We believe that this framework will be particularly useful for efficient deployment of GPs in sequential decision making problems such as active learning, Bayesian optimisation, and reinforcement learning.

\subsubsection*{Acknowledgements}
The authors would like to thank Mark Rowland, John Bradshaw, and Yingzhen Li for insightful comments and discussion. Thang D.~Bui is supported by the Google European Doctoral Fellowship. Cuong V.~Nguyen is supported by EPSRC grant EP/M0269571. Richard E.~Turner is supported by Google as well as EPSRC grants EP/M0269571 and EP/L000776/1.


\renewcommand\refname{\vskip -1cm}
\section*{References}
{\small
\bibliography{refs}{}

\begin{thebibliography}{1}

\bibitem{bui2016unifying}
T.~D. Bui, J.~Yan, and R.~E. Turner, ``A unifying framework for sparse
  {G}aussian process approximation using power expectation propagation,'' {\em
  arXiv preprint arXiv:1605.07066}, 2016.

\bibitem{HenMatGha15}
J.~Hensman, A.~G. D.~G. Matthews, and Z.~Ghahramani, ``{S}calable variational
  {G}aussian process classification,'' in {\em 18th International Conference on
  Artificial Intelligence and Statistics}, May 2015.

\end{thebibliography}


\begin{thebibliography}{10}

\bibitem{rasmussen2005gpml}
C.~E. Rasmussen and C.~K.~I. Williams, {\em {G}aussian Processes for Machine
  Learning}.
\newblock The MIT Press, 2006.

\bibitem{snelson+ghahramani:2006}
E.~Snelson and Z.~Ghahramani, ``Sparse {G}aussian processes using
  pseudo-inputs,'' in {\em Advances in Neural Information Processing Systems
  (NIPS)}, 2006.

\bibitem{titsias2009variational}
M.~K. Titsias, ``Variational learning of inducing variables in sparse
  {G}aussian processes,'' in {\em International Conference on Artificial
  Intelligence and Statistics (AISTATS)}, 2009.

\bibitem{hensman2013gaussian}
J.~Hensman, N.~Fusi, and N.~D. Lawrence, ``{G}aussian processes for big data,''
  in {\em Conference on Uncertainty in Artificial Intelligence (UAI)}, 2013.

\bibitem{HenMatGha15}
J.~Hensman, A.~G. D.~G. Matthews, and Z.~Ghahramani, ``{S}calable variational
  {G}aussian process classification,'' in {\em International Conference on
  Artificial Intelligence and Statistics (AISTATS)}, 2015.

\bibitem{dezfouli2015scalable}
A.~Dezfouli and E.~V. Bonilla, ``Scalable inference for {G}aussian process
  models with black-box likelihoods,'' in {\em Advances in Neural Information
  Processing Systems (NIPS)}, 2015.

\bibitem{hernandez2016gpc}
D.~Hern{\'a}ndez-Lobato and J.~M. Hern{\'a}ndez-Lobato, ``Scalable {G}aussian
  process classification via expectation propagation,'' in {\em International
  Conference on Artificial Intelligence and Statistics (AISTATS)}, 2016.

\bibitem{csato+opper:2002}
L.~Csat\'{o} and M.~Opper, ``Sparse online {G}aussian processes,'' {\em Neural
  Computation}, 2002.

\bibitem{csato:2002}
L.~Csat\'{o}, {\em Gaussian {P}rocesses -- Iterative Sparse Approximations}.
\newblock PhD thesis, Aston University, 2002.

\bibitem{broderick2013streaming}
T.~Broderick, N.~Boyd, A.~Wibisono, A.~C. Wilson, and M.~I. Jordan, ``Streaming
  variational {B}ayes,'' in {\em Advances in Neural Information Processing
  Systems (NIPS)}, 2013.

\bibitem{thang:deepGP}
T.~D. Bui, D.~Hern{\'a}ndez-Lobato, J.~M. Hern{\'a}ndez-Lobato, Y.~Li, and
  R.~E. Turner, ``Deep {G}aussian processes for regression using approximate
  expectation propagation,'' in {\em International Conference on Machine
  Learning (ICML)}, 2016.

\bibitem{quinonero+rasmussen:2005}
J.~Qui{\~n}onero-Candela and C.~E. Rasmussen, ``A unifying view of sparse
  approximate {G}aussian process regression,'' {\em The Journal of Machine
  Learning Research}, 2005.

\bibitem{BuiYanTur17}
T.~D. Bui, J.~Yan, and R.~E. Turner, ``A unifying framework for {G}aussian
  process pseudo-point approximations using power expectation propagation,''
  {\em Journal of Machine Learning Research}, 2017.

\bibitem{matthews+al:2016}
A.~G.~D.~G. Matthews, J.~Hensman, R.~E. Turner, and Z.~Ghahramani, ``On sparse
  variational methods and the {K}ullback-{L}eibler divergence between
  stochastic processes,'' in {\em International Conference on Artificial
  Intelligence and Statistics (AISTATS)}, 2016.

\bibitem{cheng2016incremental}
C.-A. Cheng and B.~Boots, ``Incremental variational sparse {G}aussian process
  regression,'' in {\em Advances in Neural Information Processing Systems
  (NIPS)}, 2016.

\bibitem{minka2004pep}
T.~Minka, ``Power {EP},'' tech. rep., Microsoft Research, Cambridge, 2004.

\bibitem{ghahramani2000online}
Z.~Ghahramani and H.~Attias, ``{Online variational Bayesian learning},'' in
  {\em NIPS Workshop on Online Learning}, 2000.

\bibitem{sato2001online}
M.-A. Sato, ``Online model selection based on the variational {B}ayes,'' {\em
  Neural Computation}, 2001.

\bibitem{opper1999bayesian}
M.~Opper, ``A {B}ayesian approach to online learning,'' in {\em On-Line
  Learning in Neural Networks}, 1999.

\bibitem{gpflow}
A.~G. D.~G. Matthews, M.~van~der Wilk, T.~Nickson, K.~Fujii, A.~Boukouvalas,
  P.~Le{{\'o}}n-Villagr{{\'a}}, Z.~Ghahramani, and J.~Hensman, ``{GPflow}: A
  {G}aussian process library using {TensorFlow},'' {\em Journal of Machine
  Learning Research}, 2017.

\bibitem{bauer2016nips}
M.~Bauer, M.~van~der Wilk, and C.~E. Rasmussen, ``Understanding probabilistic
  sparse {G}aussian process approximations,'' in {\em Advances in Neural
  Information Processing Systems (NIPS)}, 2016.

\bibitem{keogh1999indexing}
E.~J. Keogh and M.~J. Pazzani, ``An indexing scheme for fast similarity search
  in large time series databases,'' in {\em International Conference on
  Scientific and Statistical Database Management}, 1999.

\bibitem{garofolo1993darpa}
J.~Garofolo, L.~Lamel, W.~Fisher, J.~Fiscus, D.~Pallett, N.~Dahlgren, and
  V.~Zue, ``{TIMIT acoustic-phonetic continuous speech corpus LDC93S1},'' {\em
  Philadelphia: Linguistic Data Consortium}, 1993.

\bibitem{bui2014nips}
T.~D. Bui and R.~E. Turner, ``Tree-structured {G}aussian process
  approximations,'' in {\em Advances in Neural Information Processing Systems
  (NIPS)}, 2014.

\end{thebibliography}
\bibliographystyle{ieeetr}
}

\clearpage
\section*{Appendices}

\begin{appendices}

\section{More discussions on the paper}

\subsection{Can the variational lower bound be derived using Jensen's inequality?}
Yes. There are two equivalent ways of deriving VI:
\begin{enumerate}
\item Applying Jensen's inequality directly to the log marginal likelihood.
\item Explicitly writing down the $\mathrm{KL}(q\|p)$, noting that it is non-negative and rearranging to get the same bound as in (1).
\end{enumerate}
(1) is often used in traditional VI literature. Many recent papers (e.g. \cite{hensman2013gaussian} and our paper) use (2).

\subsection{Comparison to \cite{csato:2002}}
It is not clear how to compare to \cite{csato:2002} fairly since it does not provide methods for learning hyperparameters and their framework does not support such an extension. Accurate hyperparameter learning is required for real datasets like those in the paper. So \cite{csato:2002} performs extremely poorly unless suitable settings for the hyperparameters can be guessed from the first batch of data. Furthermore, our paper goes beyond \cite{csato:2002} by providing a method for optimising pseudo-inputs which has been shown to substantially improve upon the heuristics used in \cite{csato:2002} in the batch setting \cite{snelson+ghahramani:2006}.

\subsection{Are SVI or the stream-based method performing differently due to different approximations?}
No. Conventional SVI is fundamentally unsuited to the streaming setting and it performs very poorly practically compared to both the collapsed and uncollapsed versions of our method. The SVI learning rates require a lot of dataset and iteration specific tuning so the new data can be revisited multiple times without forgetting old data. The uncollapsed versions of our method do not require tuning of this sort and perform just as well as the collapsed version given sufficient updates.

\subsection{Are pseudo-points appropriate for streaming settings?}
In any setting (batch/streaming), pseudo-point approximations require the pseudo-points to cover the input space occupied by the data. This means they can be inappropriate for very long time-series or very high-dimensional inputs. This is a general issue with the approximation class. The development of new pseudo-point approximations to handle very large numbers of pseudo-points is a key and active research area \cite{bui2014nips}, but orthogonal to our focus in this paper. A moving window could be introduced so just recent data are modelled (as we use for SGP/GP) but the utility of this depends on the task. Here we assume all input regions must be modelled which is problematic for windowing.

\subsection{A possible explanation on why all models including full GP regression tend to learn bigger noise variances}
This is a bias that arises because the learned functions are more discrepant from the training data than the true function and so the learned observation noise inflates to accommodate the mismatch.

\subsection{Are the hyperparameters learned in the time-series and spatial data experiments?}
Yes, hyperparameters and pseudo-inputs are optimised using the online variational free energy. This is absolutely central to our approach and the key difference to \cite{csato:2002,csato+opper:2002}.

\subsection{Why is there a non-monotonic behaviour in fig.~4 in the main text?}
This occurs because at some point the GP/SGP memory window cannot cover all observed data. Some parts of the input space are then missed, leading to decreasing performance.

\section{Variational free energy approach for streaming sparse GP regression}
\subsection{The variational lower bound}
Let $\avec = f(\zvec_{\mathrm{old}})$ and $\bvec = f(\zvec_{\mathrm{new}})$ be the function values at the pseudo-inputs before and after seeing new data.
The previous posterior, $q_\mathrm{old}(f) = p(f_{\ne \avec}|\avec, \theta_\mathrm{old}) q(\avec)$, can be used to find the approximate likelihood given by old observations as follows,
\begin{align}
	p(\yvec_\mathrm{old}|f) \approx \frac{q_\mathrm{old}(f) p(\yvec_\mathrm{old}|\theta_\mathrm{old})}{p(f|\theta_\mathrm{old})} \quad\mathrm{as}\quad q_\mathrm{old}(f) \approx \frac{p(f|\theta_\mathrm{old}) p(\yvec_\mathrm{old}|f)} {p(\yvec_\mathrm{old}|\theta_\mathrm{old})}.
\end{align}
Substituting this into the posterior that we want to target gives us:
\begin{align}
p(f|\yvec_\mathrm{old},\yvec_\mathrm{new}) = \frac{p(f|\theta_\mathrm{new}) p(\yvec_\mathrm{old}|f) p(\yvec_\mathrm{new}|f)} {p(\yvec_\mathrm{new}, \yvec_\mathrm{old}|\theta_\mathrm{new})} \approx \frac{p(f|\theta_\mathrm{new}) q_\mathrm{old}(f) p(\yvec_\mathrm{old}|\theta_\mathrm{old}) p(\yvec_\mathrm{new}|f)} {p(f|\theta_\mathrm{old}) p(\yvec_\mathrm{new}, \yvec_\mathrm{old}|\theta_\mathrm{new})}. \nonumber 
\end{align}

The new posterior approximation takes the same form, but with the new pseudo-points and new hyperparameters: $q_\mathrm{new}(f) = p(f_{\ne \bvec}|\bvec, \theta_\mathrm{new}) q(\bvec)$. This approximate posterior can be obtained by minimising the KL divergence,
\begin{align}
\mathrm{KL} \lbrack q_\mathrm{new}(f) || \hat{p}(f|\yvec_\mathrm{old}, \yvec_\mathrm{new}) \rbrack 
&= \int \dd f q_\mathrm{new}(f) \log \frac{p(f_{\ne \bvec}|\bvec, \theta_\mathrm{new}) q_\mathrm{new}(\bvec)}{\frac{\mathcal{Z}_1(\theta_\mathrm{old})}{\mathcal{Z}_2(\theta_\mathrm{new})} p(f|\theta_\mathrm{new}) p(\yvec_\mathrm{new}|f) \frac{q_\mathrm{old}(f)}{p(f|\theta_\mathrm{old})}} \label{eqn:kl}\\
&= \log\frac{\mathcal{Z}_2(\theta_\mathrm{new})}{\mathcal{Z}_1(\theta_\mathrm{old})} + \int \dd f q_\mathrm{new}(f) \left[ \log \frac{p(\avec|\theta_\mathrm{old})q_\mathrm{new}(\bvec)}{p(\bvec|\theta_\mathrm{new}) q_\mathrm{old}(\avec) p(\yvec_\mathrm{new}|f)} \right]. \label{eq:kl-simplified}
\end{align}
The last equation above is obtained by noting that $p(f|\theta_\mathrm{new})/p(f_{\ne \bvec}|\bvec, \theta_\mathrm{new}) = p(\bvec|\theta_\mathrm{new})$ and
\begin{align}
\frac{q_\mathrm{old}(f)}{p(f|\theta_\mathrm{old})} = \frac{\bcancel{p(f_{\ne \avec}|\avec, \theta_\mathrm{old})} q_\mathrm{old}(\avec)}{\bcancel{p(f_{\ne \avec}|\avec, \theta_\mathrm{old})} p(\avec|\theta_\mathrm{old})} = \frac{q_\mathrm{old}(\avec)}{p(\avec|\theta_\mathrm{old})}.\nonumber
\end{align}

Since the KL divergence is non-negative, the second term in \eqref{eq:kl-simplified} is the negative lower bound of the approximate online log marginal likelihood, or the variational free energy, $\mathcal{F}(q_\mathrm{new}(f))$. We can decompose the bound as follows,
\begin{align}
\mathcal{F}(q_\mathrm{new}(f)) 
	&= \int \dd f q_\mathrm{new}(f) \left[ \log \frac{p(\avec|\theta_\mathrm{old})q_\mathrm{new}(\bvec)}{p(\bvec|\theta_\mathrm{new}) q_\mathrm{old}(\avec) p(\yvec_\mathrm{new}|f)} \right]\\
	&= \mathrm{KL}(q(\bvec) || p(\bvec|\theta_\mathrm{new})) - \int q_\mathrm{new}(f) \log p(\yvec_\mathrm{new}|f) \nonumber \\ & \quad \quad + \mathrm{KL}(q_\mathrm{new}(\avec) || q_\mathrm{old}(a)) - \mathrm{KL}(q_\mathrm{new}(\avec) || p(\avec | \theta_\mathrm{old})).
\end{align}
The first two terms form the batch variational bound if the current batch is the whole training data, and the last two terms constrain the posterior to take into account the old likelihood (through the approximate posterior and the prior). 

\subsection{Derivation of the optimal posterior update and the collapsed bound}
The aim is to find the new approximate posterior $q_\mathrm{new}(f)$ such that the free energy is minimised. 
This is achieved by setting the derivative of $\mathcal{F}$ and a Lagrange term \footnote{to ensure $q(\bvec)$ is normalised} w.r.t. $q(\bvec)$ equal 0,
\begin{align}
\frac{\dd \mathcal{F}} {\dd q(\bvec)} + \lambda = \int \dd f_{\ne \bvec} p(f_{\ne \bvec}|\bvec) \left[ \log \frac{p(\avec|\theta_\mathrm{old})q(\bvec)}{p(\bvec|\theta_\mathrm{new}) q(\avec)} - \log p(\yvec|\fvec) \right] + 1 + \lambda = 0,
\end{align}
resulting in,
\begin{align}
q_\mathrm{opt}(\bvec) = \frac{1}{\mathcal{C}} p(\bvec) \exp \Big( \int \dd \avec p(\avec|\bvec) \log \frac{q(\avec)}{p(\avec|\theta_\mathrm{old})} + \int \dd \fvec p(\fvec|\bvec) \log p(\yvec|\fvec) \Big).
\end{align}
Note that we have dropped $\theta_\mathrm{new}$ from $p(\bvec|\theta_\mathrm{new})$, $p(\avec|\bvec, \theta_\mathrm{new})$ and $p(\fvec|\bvec, \theta_\mathrm{new})$ to lighten the notation. 
Substituting the above result into the variational free energy leads to $\mathcal{F}(q_\mathrm{opt}(f)) = -\log \mathcal{C}$. We now consider the exponents in the optimal $q_\mathrm{opt}(\bvec)$, noting that $q(\avec) = \norm(\avec; \ma, \Sa)$ and $p(\avec|\theta_\mathrm{old}) = \norm(\avec; \zero, \kaa')$, and denoting $\Da = (\Sa^{-1} - \kaa'^{-1})^{-1}$, $\mathbf{Q}_\fvec = \kff - \kfb\kbb^{-1} \kbf$, and $\Qa = \kaa - \kab\kbb^{-1} \kba$,
\begin{align}
\small
\mathrm{E}_1
&= \int \dd \avec p(\avec|\bvec) \log \frac{q(\avec)}{p(\avec|\theta_\mathrm{old})} \\
&= \frac{1}{2} \int \dd \avec \norm(\avec; \kab\kbb^{-1} \bvec, \Qa) \Big(- \log \frac{\vert \Sa \vert}{\vert \kaa' \vert} - (\avec - \ma)^\intercal \Sa^{-1} (\avec - \ma)  + \avec^\intercal \kaa'^{-1} \avec \Big)\nonumber\\
&= \log \norm(\Da\Sa^{-1}\ma; \kab\kbb^{-1}\bvec, \Da) + \Delta_1,\\
\mathrm{E}_2 
&= \int \dd \fvec p(\fvec|\bvec) \log p(\yvec|\fvec) \\
&= \int \dd \fvec \mathcal{N}(\fvec; \kfb\kbb^{-1}\bvec, \mathbf{Q}_\fvec) \log\norm(\yvec; \fvec, \sigma^2\mathrm{I})\\
&=\log\norm(\yvec; \kfb\kbb^{-1} \bvec, \sigma^2\mathrm{I}) + \Delta_2,\\
2\Delta_1 &= - \log \frac{\vert \Sa \vert}{\vert \kaa' \vert \vert \Da \vert} + \ma^\intercal\Sa^{-1}\Da\Sa^{-1}\ma - \mathrm{tr} \lbrack \Da^{-1} \Qa \rbrack - \ma^\intercal\Sa^{-1}\ma + M_\mathbf{a}\log(2\pi),\\
\Delta_2 &= -\frac{1}{2\sigma^2} \mathrm{tr} (\mathbf{Q}_\fvec).
\end{align}
Putting these results back into the optimal $q(\bvec)$, we obtain:
\begin{align}
q_\mathrm{opt}(\bvec) 
&\propto p(\bvec) \norm(\hat{\yvec}, \khatb \kbb^{-1} \bvec, \Sigma_{\hat{\yvec}}) \\
&= \norm(\bvec; \kbhat(\khatb\kbb^{-1}\kbhat + \Sigma_{\hat{\yvec}})^{-1} \hat{\yvec}, \kbb - \kbhat(\khatb\kbb^{-1}\kbhat + \Sigma_{\hat{\yvec}})^{-1} \khatb)
\end{align}
where
\begin{align}
\hat{\yvec} = \begin{bmatrix} \yvec \\ \Da\Sa^{-1} \ma \end{bmatrix}, \; \khatb = \begin{bmatrix} \kfb \\ \kab \end{bmatrix}, \; \Sigma_{\hat{\yvec}} = \begin{bmatrix} \sigma_y^2\mathrm{I} & \zero \\ \zero & \Da \end{bmatrix}.
\end{align}
The negative variational free energy, which is the lower bound of the log marginal likelihood, can also be derived,
\begin{align}
\mathcal{F} =  \log \mathcal{C} = \log \norm (\hat{\yvec}; \zero, \khatb\kbb^{-1}\kbhat + \Sigma_{\hat{\yvec}}) + \Delta_1 + \Delta_2. \label{eqn:vfe}
\end{align}

\subsection{Implementation}
In this section, we provide efficient and numerical stable forms for a practical implementation of the above results. 
\subsubsection{The variational free energy}
The first term in \cref{eqn:vfe} can be written as follows,
\begin{align}
	\mathcal{F}_1 
	& = \log \norm (\hat{\yvec}; \zero, \khatb\kbb^{-1}\kbhat + \Sigma_{\hat{\yvec}}) \\
	& = -\frac{N+M_a}{2}\log(2\pi) - \frac{1}{2} \log \vert \khatb\kbb^{-1}\kbhat + \Sigma_{\hat{\yvec}} \vert - \frac{1}{2} \hat{\yvec}^\intercal (\khatb\kbb^{-1}\kbhat + \Sigma_{\hat{\yvec}})^{-1} \hat{\yvec}.
\end{align}
Let $\Svec_y = \khatb\kbb^{-1}\kbhat + \Sigma_{\hat{\yvec}}$ and $\kbb = \lb\lb^{\intercal}$, using the matrix determinant lemma, we obtain,
\begin{align}
\log \vert \Svec_\yvec \vert 
&= \log \vert \khatb \kbb^{-1} \kbhat + \Sigma_{\hat{\yvec}} \vert \\
&= \log \vert \Sigma_{\hat{\yvec}} \vert + \log \vert \mathrm{I} +  \lb^{-1}\kbhat \Sigma_{\hat{\yvec}}^{-1} \khatb \lb^{-\intercal} \vert \\
&=  N \log \sigma_y^2 + \log \vert \mathrm{D}_\avec \vert + \log \vert \mathrm{I} +  \lb^{-1}\kbhat \Sigma_{\hat{\yvec}}^{-1} \khatb \lb^{-\intercal} \vert.
\end{align}
Let $\mathbf{D} = \mathrm{I} +  \lb^{-1}\kbhat \Sigma_{\hat{\yvec}}^{-1} \khatb \lb^{-\intercal}$. Note that,
\begin{align}
\kbhat \Sigma_{\hat{\yvec}}^{-1} \khatb = \frac{1}{\sigma_y^2} \kbf \kfb + \kba \Sa^{-1} \kab - \kba \kaa'^{-1} \kab.
\end{align}
Using the matrix inversion lemma gives us,
\begin{align}
\Svec_\yvec^{-1} 
&= (\khatb \kbb^{-1} \kbhat + \Sigma_{\hat{\yvec}})^{-1} \\
&= \Sigma_{\hat{\yvec}}^{-1} - \Sigma_{\hat{\yvec}}^{-1} \khatb \lb^{-\intercal} \mathbf{D}^{-1} \lb^{-1} \kbhat \Sigma_{\hat{\yvec}}^{-1},
\end{align}
leading to,
\begin{align}
\hat{\yvec}^\intercal \Svec_\yvec^{-1} \hat{\yvec} = \hat{\yvec}^\intercal \Sigma_{\hat{\yvec}}^{-1} \hat{\yvec} - \hat{\yvec}^\intercal \Sigma_{\hat{\yvec}}^{-1} \khatb \lb^{-\intercal} \mathbf{D}^{-1} \lb^{-1} \kbhat \Sigma_{\hat{\yvec}}^{-1} \hat{\yvec}.
\end{align}
Note that,
\begin{align}
\hat{\yvec}^\intercal \Sigma_{\hat{\yvec}}^{-1} \hat{\yvec} &= \frac{1}{\sigma_y^2}\yvec^\intercal\yvec + \ma^\intercal\Sa^{-1}\Da\Sa^{-1}\ma, \\
\text{and}\quad \mathbf{c} &= \kbhat \Sigma_{\hat{\yvec}}^{-1} \hat{\yvec} = \frac{1}{\sigma^2}\kbf \yvec + \kba\Sa^{-1}\ma.
\end{align}
Substituting these results back into equation \cref{eqn:vfe},
\begin{align}
\mathcal{F} &=  -\frac{N}{2}\log(2\pi\sigma^2) - \frac{1}{2}\log\vert\mathbf{D}\vert - \frac{1}{2\sigma^2}\yvec^\intercal\yvec + \frac{1}{2} \mathbf{c}^\intercal \lb^{-\intercal}\mathbf{D}^{-1}\lb^{-1}\mathbf{c} \nonumber \\
& - \frac{1}{2} \log \vert \Sa \vert + \frac{1}{2} \log \vert \kaa' \vert - \frac{1}{2} \mathrm{tr} \lbrack \Da^{-1} \Qa \rbrack - \frac{1}{2} \ma^\intercal\Sa^{-1}\ma - \frac{1}{2\sigma^2} \mathrm{tr} (\mathbf{Q}_\fvec).
\end{align}

\subsubsection{Prediction}
We revisit and rewrite the optimal variational distribution, $q_{\mathrm{opt}}(\bvec)$, using its natural parameters:
\begin{align}
q_\mathrm{opt}(\bvec) 
&\propto p(\bvec) \norm(\hat{\yvec}, \khatb \kbb^{-1} \bvec, \Sigma_{\hat{\yvec}}) \\
&= \norm^{-1}(\bvec;  \kbb^{-1} \kbhat \Sigma_{\hat{\yvec}}^{-1} \hat{\yvec}, \kbb^{-1} + \kbb^{-1}\kbhat \Sigma_{\hat{\yvec}}^{-1} \khatb\kbb^{-1}).
\end{align}
The predictive covariance at some test points $\svec$ is:
\begin{align}
\mathbf{V}_{\svec\svec} 
&= \kss - \ksb \kbb^{-1} \kbs + \ksb \kbb^{-1} (\kbb^{-1} + \kbb^{-1}\kbhat \Sigma_{\hat{\yvec}}^{-1} \khatb\kbb^{-1})^{-1} \kbb^{-1} \kbs \\
&= \kss - \ksb \kbb^{-1} \kbs + \ksb \lb^{-\intercal} (\mathrm{I} + \lb^{-1}\kbhat \Sigma_{\hat{\yvec}}^{-1} \khatb\lb^{-\intercal})^{-1} \lb^{-\intercal} \kbs\\
&= \kss - \ksb \kbb^{-1} \kbs + \ksb \lb^{-\intercal} \mathbf{D}^{-1} \lb^{-\intercal} \kbs.
\end{align}
And the predictive mean is:
\begin{align}
\mathbf{m}_{\svec} 
&= \ksb\kbb^{-1} (\kbb^{-1} + \kbb^{-1}\kbhat \Sigma_{\hat{\yvec}}^{-1} \khatb\kbb^{-1})^{-1} \kbb^{-1} \kbhat \Sigma_{\hat{\yvec}}^{-1} \hat{\yvec} \\
&= \ksb\lb^{-\intercal} (\mathrm{I} + \lb^{-1}\kbhat \Sigma_{\hat{\yvec}}^{-1} \khatb\lb^{-\intercal})^{-1} \lb^{-1} \kbhat \Sigma_{\hat{\yvec}}^{-1} \hat{\yvec} \\
&= \ksb\lb^{-\intercal} \mathbf{D}^{-1} \lb^{-1} \kbhat \Sigma_{\hat{\yvec}}^{-1} \hat{\yvec}.
\end{align}


\section{Power-EP for streaming sparse Gaussian process regression}
Similar to the variational approach above, we also use $\avec = f(\zvec_{\mathrm{old}})$ and $\bvec = f(\zvec_{\mathrm{new}})$ as pseudo-outputs before and after seeing new data.
The exact posterior upon observing new data is
\begin{align}
p(f|\yvec, \yvec_{\mathrm{old}}) 
	&= \frac{1}{\mathcal{Z}}p(f_{\neq \avec}|\avec, \theta_{\mathrm{old}}) q(\avec) p(\yvec | f) \\
	&= \frac{1}{\mathcal{Z}}p(f|\theta_{\mathrm{old}}) \frac{ q(\avec) }{ p(\avec|\theta_{\mathrm{old}}) } p(\yvec | f).
\end{align}
In addition, we assume that the hyperparameters do not change significantly after each online update and as a result, the exact posterior can be approximated by:
\begin{align}
p(f|\yvec, \yvec_{\mathrm{old}}) \approx \frac{1}{\mathcal{Z}} p(f|\theta_{\mathrm{new}}) \frac{ q(\avec) }{ p(\avec|\theta_{\mathrm{old}}) } p(\yvec | f).
\end{align}
We posit the following approximate posterior, which mirrors the form of the exact posterior,
\begin{align}
q(f) \propto p(f|\theta_{\mathrm{new}}) q_1(\bvec) q_2(\bvec),
\end{align}
where $q_1(\bvec)$ and $q_2(\bvec)$ are the approximate effect that $\frac{ q(\avec) }{ p(\avec|\theta_{\mathrm{old}}) }$ and $p(\yvec | f)$ have on the posterior, respectively. Next we describe steps to obtain the closed-form expressions for the approximate factors and the approximate marginal likelihood.

\subsection{$q_1(\bvec)$}
The cavity and tilted distributions are:
\begin{align}
q_{\mathrm{cav}, 1}(f) 
	&= p(f) q_1^{1-\alpha}(\bvec) q_2(\bvec) \\
	&= p(f_{\neq \avec, \bvec}|\bvec)  p(\bvec) q_2(\bvec) p(\avec|\bvec)q_1^{1-\alpha}(\bvec) \\
\text{and}\;\; \tilde{q}_{1}(f) 
	&= p(f_{\neq \avec, \bvec}|\bvec)  p(\bvec) q_2(\bvec) p(\avec|\bvec)q_1^{1-\alpha}(\bvec) \left(\frac{ q(\avec) }{ p(\avec|\theta_{\mathrm{old}}) }\right)^\alpha.
\end{align}
We note that, $q(\avec) = \norm(\avec; \ma, \Sa)$ and $p(\avec|\theta_\mathrm{old}) = \norm(a; 0, \kaa'$), leading to:
\begin{align}
\left(\frac{ q(\avec) }{ p(\avec|\theta_{\mathrm{old}}) }\right)^\alpha &= C_1 \norm(\avec; \mahat, \Sahat) \\
\mathrm{where}\;\; \mahat & = \Da \Sa^{-1} \ma, \\
\Sahat & = \frac{1}{\alpha} \Da, \\
\Da &= (\Sa^{-1} - \kaa'^{-1})^{-1}, \\
C_1 &= (2\pi)^{M/2} |\kaa'|^{\alpha/2} |\Sa|^{-\alpha/2} |\Sahat|^{1/2} \exp(\frac{\alpha}{2} \ma^{\intercal} [\Sa^{-1} \Da \Sa^{-1} - \Sa^{-1}] \ma).
\end{align}
Let $\Siga = \Da + \alpha\Qa$. Note that:
\begin{align}
p(\avec|\bvec) = \norm(\avec; \kab\kbb^{-1}\bvec; \kaa-\kab\kbb^{-1}\kba) = \norm(\avec; \Wa \bvec, \Qa).
\end{align}
As a result,
\begin{align}
\int \dd \avec p(\avec|\bvec) \left(\frac{ q(\avec) }{ p(\avec|\theta_{\mathrm{old}}) }\right)^\alpha 
	&= \int \dd \avec C_1 \norm(\avec; \mahat, \Sahat) \norm(\avec; \Wa \bvec, \Qa) \\
	&= C_1 \norm(\mahat; \Wa \bvec, \Siga / \alpha).
\end{align}
Since this is the contribution towards the posterior from $\avec$, it needs to match $q_1^{\alpha}(\bvec)$ at convergence, that is,
\begin{align}
q_1(\bvec) 
	&\propto \left[ C_1 \norm(\mahat; \Wa \bvec, \Siga / \alpha) \right]^{1/\alpha}\\
	&= \norm(\mahat; \Wa \bvec, \alpha(\Siga / \alpha)) \\
	&= \norm(\mahat; \Wa \bvec, \Siga).
\end{align}
In addition, we can compute:
\begin{align}
\log \tilde{Z}_1 
	&= \log \int \dd f \tilde{q}_1(f) \\
	&= \log C_1 \norm(\mahat; \Wa \mcav, \Siga / \alpha + \Wa  \Vcav \Wa ^\intercal) \\
	&= \log C_1 - \frac{M}{2} \log (2\pi) - \frac{1}{2} \log |\Siga / \alpha + \Wa  \Vcav \Wa^\intercal| - \frac{1}{2} \mahatT (\Siga / \alpha + \Wa  \Vcav \Wa^\intercal)^{-1} \mahat \nonumber\\ &\;\;\; + \mcavT \Wa^\intercal (\Siga / \alpha + \Wa  \Vcav \Wa^\intercal)^{-1} \mahat - \frac{1}{2} \mcavT \Wa ^\intercal (\Siga / \alpha + \Wa \Vcav \Wa^\intercal)^{-1} \Wa \mcav.
\end{align}
Note that: 
\begin{align}
\Vvec^{-1} &= \Vcav^{-1} + \Wa ^\intercal (\Siga / \alpha)^{-1} \Wa, \\
\Vvec^{-1}\mvec &= \Vcav^{-1} \mcav + \Wa ^\intercal (\Siga / \alpha)^{-1} \mahat.
\end{align}
Using matrix inversion lemma gives
\begin{align}
\Vvec = \Vcav - \Vcav \Wa ^\intercal (\Siga / \alpha + \Wa \Vcav \Wa^\intercal)^{-1} \Wa \Vcav.
\end{align}
Using matrix determinant lemma gives
\begin{align}
|\Vvec^{-1}| = |\Vcav^{-1}| |(\Siga / \alpha)^{-1}| |\Siga / \alpha + \Wa \Vcav \Wa ^\intercal|.
\end{align}
We can expand terms in $\log \tilde{Z}_1$ above as follows:
\begin{align}
\log \tilde{Z}_{1A} 
	&= - \frac{1}{2} \log |\Siga / \alpha + \Wa \Vcav \Wa^\intercal| \\
	&= - \frac{1}{2} (\log |\Vvec^{-1}| - \log |\Vcav^{-1}| - \log |(\Siga / \alpha)^{-1}|)\\
	&= \frac{1}{2} \log |\Vvec| - \frac{1}{2} \log |\Vcav| - \frac{1}{2}\log |(\Siga / \alpha)|. \\
\log \tilde{Z}_{1B} 
	&= - \frac{1}{2} \mahatT (\Siga / \alpha + \Wa \Vcav \Wa^\intercal)^{-1} \mahat \\
	&= - \frac{1}{2} \mahatT (\Siga / \alpha)^{-1} \mahat + \frac{1}{2} \mahatT (\Siga / \alpha)^{-1} \Wa \Vvec \Wa^\intercal(\Siga / \alpha)^{-1} \mahat.\\
\log \tilde{Z}_{1C}
	&= \mcavT \Wa^\intercal (\Siga / \alpha + \Wa \Vcav \Wa^\intercal)^{-1} \mahat \\
	&= \mcavT \Wa^\intercal  (\Siga / \alpha)^{-1} \mahat - \mcavT \Wa^\intercal  (\Siga / \alpha)^{-1} \Wa \Vvec \Wa^\intercal(\Siga / \alpha)^{-1} \mahat.\\
\log \tilde{Z}_{1D}
	&= - \frac{1}{2} \mcavT \Wa^\intercal (\Siga / \alpha + \Wa \Vcav \Wa^\intercal)^{-1} \Wa \mcav\\
	&= - \frac{1}{2} \mcavT \Vcav^{-1} \mcav + \frac{1}{2} \mcavT \Vcav^{-1} \Vvec \Vcav^{-1} \mcav \\
	&= - \frac{1}{2} \mcavT \Vcav^{-1} \mcav + \frac{1}{2} \mvecT \Vvec^{-1} \mvec \nonumber \\ & \;\; + \frac{1}{2} \mahatT (\Siga / \alpha)^{-1} \Wa \Vvec \Wa^\intercal(\Siga / \alpha)^{-1} \mahat - \mahat (\Siga / \alpha)^{-1} \Wa \mvec.\\
\log \tilde{Z}_{1DA}
	&= - \mahat (\Siga / \alpha)^{-1} \Wa \mvec \\
	&= - \mahatT (\Siga / \alpha)^{-1} \Wa \Vvec \Vcav^{-1} \mcav - \mahatT (\Siga / \alpha)^{-1} \Wa \Vvec \Wa^\intercal(\Siga / \alpha)^{-1} \mahat.\\
\log \tilde{Z}_{1DA1}
	&= - \mahatT((\Siga / \alpha)^{-1}) \Wa \Vvec \Vcav^{-1} \mcav \\
	&= - \mahatT (\Siga / \alpha)^{-1} \Wa (\mathrm{I} - \Vvec \Wa^\intercal \Siga / \alpha)^{-1} \Wa) \mcav \\
	&= - \mcavT \Wa^\intercal  (\Siga / \alpha)^{-1} \mahat + \mcavT \Wa^\intercal  (\Siga / \alpha)^{-1} \Wa \Vvec \Wa^\intercal(\Siga / \alpha)^{-1} \mahat.
\end{align}
which results in:
\begin{align}
\log \tilde{Z}_{1} + \phi_{\mathrm{cav, 1}} - \phi_{\mathrm{post}} = \log C_1 - \frac{M}{2} \log (2\pi) - \frac{1}{2}\log |(\Siga / \alpha)| - \frac{1}{2} \mahatT (\Siga / \alpha)^{-1} \mahat.
\end{align}


\subsection{$q_2(\bvec)$}
We repeat the above procedure to find $q_2(\bvec)$.
The cavity and tilted distributions are,
\begin{align}
q_{\mathrm{cav}, 2}(f) 
	&= p(f) q_1(\bvec) q_2^{1-\alpha}(\bvec) \\
	&= p(f_{\neq \fvec, \bvec|\bvec) p(\bvec) q_1(\bvec) p(\fvec|\bvec}) q_2^{1-\alpha}(\bvec) \\
\text{and}\;\; \tilde{q}_{2}(f) 
	&= p(f_{\neq \fvec, \bvec}|\bvec)  p(\bvec) q_1(\bvec) p(\avec|\bvec)q_2^{1-\alpha}(\bvec) p^\alpha(\yvec|\fvec)
\end{align}
We note that, $p(\yvec|\fvec) = \norm(\yvec; \fvec, \sigma_y^2\mathrm{I})$ leading to,
\begin{align}
p^\alpha(\yvec|\fvec) &= C_2 \norm(\yvec; \fvec, \Syhat) \\
\mathrm{where}\;\; \Syhat & = \frac{\sigma_y^2}{\alpha} \mathrm{I} \\
C_2 &= (2\pi\sigma_y^2)^{N(1-\alpha)/2} \alpha^{-N/2}
\end{align}
Let $\Sigy = \sigma_y^2\mathrm{I} + \alpha\Qf$. Note that,
\begin{align}
p(\fvec|\bvec) = \norm(\fvec; \kfb\kbb^{-1}\bvec; \kff-\kfb\kbb^{-1}\kbf) = \norm(\avec; \Wf\bvec, \Qf)
\end{align}
As a result,
\begin{align}
\int \dd \avec p(\fvec|\bvec) p^\alpha(\yvec|\fvec)
	&= \int \dd \fvec C_2 \norm(\yvec; \fvec, \Syhat) \norm(\fvec; \Wf\bvec, B) \\
	&= C_2 \norm(\yvec; \Wf\bvec, \Syhat + \Qf)
\end{align}
Since this is the contribution towards the posterior from $\yvec$, it needs to match $q^{\alpha}(\bvec)$ at convergence, that is,
\begin{align}
q_2(\bvec) 
	&\propto \left[ C_2 \norm(\yvec; \Wf\bvec, \Syhat + \Qf) \right]^{1/\alpha}\\
	&= \norm(\yvec; \Wf\bvec, \alpha(\Sigy / \alpha)) \\
	&= \norm(\yvec; \Wf\bvec, \Sigy)
\end{align}
In addition, we can compute,
\begin{align}
\log \tilde{Z}_2 
	&= \log \int \dd f \tilde{q}_2(f) \\
	&= \log C_2 \norm(\yvec; \Wf \mcav, \Sigy / \alpha + \Wf \Vcav \Wf^\intercal) \\
	&= \log C_2 - \frac{N}{2} \log (2\pi) - \frac{1}{2} \log |\Sigy / \alpha + \Wf \Vcav \Wf^\intercal| - \frac{1}{2} \yvecT (\Sigy / \alpha + \Wf \Vcav \Wf^\intercal)^{-1} \yvec \nonumber\\ &\;\;\; + \mcavT \Wf^\intercal (\Sigy / \alpha + \Wf \Vcav \Wf^\intercal)^{-1} \yvec - \frac{1}{2} \mcavT \Wf^\intercal (\Sigy / \alpha + \Wf \Vcav \Wf^\intercal)^{-1} \Wf \mcav
\end{align}
By following the exact procedure as shown above for $q_1(\bvec)$, we can obtain,
\begin{align}
\log \tilde{Z}_{2} + \phi_{\mathrm{cav, 2}} - \phi_{\mathrm{post}} = \log C_2 - \frac{N}{2} \log (2\pi) - \frac{1}{2}\log |(\Sigy / \alpha)| - \frac{1}{2} \yvecT (\Sigy / \alpha)^{-1} \yvec
\end{align}


\subsection{Approximate posterior}
Putting the above results together gives the approximate posterior over $\bvec$ as follows,
\begin{align}
q_\mathrm{opt}(\bvec) 
&\propto p(\bvec) q_1(\bvec) q_2(\bvec) \\
&\propto p(\bvec) \norm(\hat{\yvec}, \khatb \kbb^{-1} \bvec, \Sigma_{\hat{\yvec}}) \\
&= \norm(\avec; \kbhat(\khatb\kbb^{-1}\kbhat + \Sigma_{\hat{\yvec}})^{-1} \hat{\yvec}, \kbb - \kbhat(\khatb\kbb^{-1}\kbhat + \Sigma_{\hat{\yvec}})^{-1} \khatb)
\end{align}
where
\begin{align}
\hat{\yvec} = \begin{bmatrix} \yvec \\ \yvec_a \end{bmatrix} = \begin{bmatrix} \yvec \\ \Da \Sa^{-1}\ma \end{bmatrix}, \; \khatb = \begin{bmatrix} \kfb \\ \kab \end{bmatrix}, \; \Sigma_{\hat{\yvec}} = \begin{bmatrix} \Sigma_\yvec & \zero \\ \zero & \Sigma_\avec \end{bmatrix},
\end{align}
and $\Sigma_\yvec = \sigma^2\mathrm{I} + \alpha \mathrm{diag}\Qf$, and $\Sigma_\avec = \Da + \alpha \Qa$.

\subsection{Approximate marginal likelihood}
The Power-EP procedure above also provides us an approximation to the marginal likelihood, which can be used to optimise the hyperparameters and the pseudo-inputs,
\begin{align}
\mathcal{F} = \phi_{\mathrm{post}} - \phi_{\mathrm{prior}} + \frac{1}{\alpha}(\log \tilde{Z}_{1} + \phi_{\mathrm{cav, 1}} - \phi_{\mathrm{post}}) + \frac{1}{\alpha}(\log \tilde{Z}_{2} + \phi_{\mathrm{cav, 2}} - \phi_{\mathrm{post}})
\end{align}
Note that,
\begin{align}
\Delta_0 
	&= \phi_{\mathrm{post}} - \phi_{\mathrm{prior}}\\
	&= \frac{1}{2} \log |\Vvec| + \frac{1}{2} \mvecT \Vvec^{-1} \mvec - \frac{1}{2} \log |\kbb| \\
	&= - \frac{1}{2} \log |\Sigma_{\hat{\yvec}}| + \frac{1}{2} \log |\Sigma_a| + \frac{1}{2} \log |\Sigma_y| - \frac{1}{2} \hat{\yvec}^{\intercal} \Sigma_{\hat{\yvec}}^{-1} \hat{\yvec} + \frac{1}{2} \yvec^{\intercal} \Sigma_{\yvec}^{-1} \yvec + \frac{1}{2} \yvec_{\avec}^{\intercal} \Sigma_{\avec}^{-1} \yvec_{\avec}\\
\Delta_1 
	&= \frac{1}{\alpha}(\log \tilde{Z}_{1} + \phi_{\mathrm{cav, 1}} - \phi_{\mathrm{post}}) \\
	&= \frac{1}{2} \log \frac{|\kaa'|}{|\Sa|} - \frac{1}{2\alpha}\log |\mathrm{I} + \alpha \Da^{-1}\Qa| - \frac{1}{2} \yvec_{\avec}^{\intercal} \Sigma_{\avec}^{-1} \yvec_{\avec} + \frac{1}{2} \ma^{\intercal} [\Sa^{-1} \Da \Sa^{-1} - \Sa^{-1}] \ma \\
\Delta_2
	&= \frac{1}{\alpha}(\log \tilde{Z}_{2} + \phi_{\mathrm{cav, 2}} - \phi_{\mathrm{post}}) \\
	&= -\frac{N}{2} \log (2\pi) + \frac{N(1-\alpha)}{2\alpha} \log (\sigma_y^2) - \frac{1}{2\alpha} \log |\Sigma_\yvec| - \frac{1}{2} \yvec^{\intercal} \Sigma_{\yvec}^{-1} \yvec
\end{align}
Therefore,
\begin{align}
\mathcal{F} &= \log \norm (\hat{\yvec}; 0, \Sigma_{\hat{\yvec}}) + \frac{N(1-\alpha)}{2\alpha} \log (\sigma_y^2) - \frac{1-\alpha}{2\alpha} \log |\Sigma_\yvec| \nonumber \\ &\;\;\; + \frac{1}{2} \log |\kaa'| - \frac{1}{2} \log |\Sa| + \frac{1}{2} \log |\Sigma_a| - \frac{1}{2\alpha}\log |\mathrm{I} + \alpha \Da^{-1}\Qa| \nonumber \\ &\;\;\; + \frac{M_a}{2} \log (2\pi) + \frac{1}{2} \ma^{\intercal} [\Sa^{-1} \Da \Sa^{-1} - \Sa^{-1}] \ma \label{eqn:pep_energy}
\end{align}
The limit as $\alpha$ tends to 0 is the variational free energy in \cref{eqn:vfe}. This is achieved similar to the batch case as detailed in \cite{BuiYanTur17} and by further observing that as $\alpha \to 0$,
\begin{align}
\frac{1}{2\alpha}\log |\mathrm{I} + \alpha \Da^{-1}\Qa| 
	& \approx \frac{1}{2\alpha} \log (1 + \alpha\mathrm{tr}(\Da^{-1}\Qa) + \mathcal{O}(\alpha^2)) \\
	& \approx \frac{1}{2} \mathrm{tr}(\Da^{-1}\Qa)
\end{align}


\subsection{Implementation}

In this section, we provide efficient and numerical stable forms for a practical implementation of the above results. 
\subsubsection{The Power-EP approximate marginal likelihood}
The first term in \cref{eqn:pep_energy} can be written as follows,
\begin{align}
	\mathcal{F}_1 
	& = \log \norm (\hat{\yvec}; \zero, \khatb\kbb^{-1}\kbhat + \Sigma_{\hat{\yvec}}) \\
	& = -\frac{N+M_a}{2}\log(2\pi) - \frac{1}{2} \log \vert \khatb\kbb^{-1}\kbhat + \Sigma_{\hat{\yvec}} \vert - \frac{1}{2} \hat{\yvec}^\intercal (\khatb\kbb^{-1}\kbhat + \Sigma_{\hat{\yvec}})^{-1} \hat{\yvec}
\end{align}
Let denote $\Svec_\yvec = \khatb\kbb^{-1}\kbhat + \Sigma_{\hat{\yvec}}$, $\kbb = \lb\lb^\intercal$, $\Qa = \la\laT$, $\Ma = \mathrm{I} + \alpha \laT\Da^{-1}\la$ and $\mathbf{D} = \mathrm{I} +  \lb^{-1}\kbhat \Sigma_{\hat{\yvec}}^{-1} \khatb \lb^{-\intercal}$. By using the matrix determinant lemma, we obtain,
\begin{align}
\log \vert \Svec_\yvec \vert 
&= \log \vert \khatb \kbb^{-1} \kbhat + \Sigma_{\hat{\yvec}} \vert \\
&= \log \vert \Sigma_{\hat{\yvec}} \vert + \log \vert \mathrm{I} +  \lb^{-1}\kbhat \Sigma_{\hat{\yvec}}^{-1} \khatb \lb^{-\intercal} \vert\\
& = \log |\Sigma_\yvec| + \log |\Sigma_\avec| + \log \vert D \vert
\end{align}
Note that,
\begin{align}
\kbhat \Sigma_{\hat{\yvec}}^{-1} \khatb &= \kbf \Sigma_{\yvec}^{-1} \kfb + \kba \Sigma_{\avec}^{-1} \kab \\
\kbf \Sigma_{\yvec}^{-1} \kfb &= \kbf (\sigma_y^2\mathrm{I} + \alpha\Qf)^{-1} \kfb \\
\kba \Sigma_{\avec}^{-1} \kab 
	&= \kba (\Da + \alpha \Qa)^{-1} \kab\\
	&= \kba (\Da^{-1} - \alpha\Da^{-1} \la [\mathrm{I} + \alpha \laT\Da^{-1}\la]^{-1} \laT\Da^{-1}) \kab\\
	&= \kba \Da^{-1} \kab - \alpha \kba \Da^{-1} \la \Ma ^{-1} \laT\Da^{-1} \kab
\end{align}

Using the matrix inversion lemma gives us,
\begin{align}
\Svec_\yvec^{-1} 
&= (\khatb \kbb^{-1} \kbhat + \Sigma_{\hat{\yvec}})^{-1} \\
&= \Sigma_{\hat{\yvec}}^{-1} - \Sigma_{\hat{\yvec}}^{-1} \khatb \lb^{-\intercal} \mathbf{D}^{-1} \lb^{-1} \kbhat \Sigma_{\hat{\yvec}}^{-1}
\end{align}
leading to,
\begin{align}
\hat{\yvec}^\intercal \Svec_\yvec^{-1} \hat{\yvec} = \hat{\yvec}^\intercal \Sigma_{\hat{\yvec}}^{-1} \hat{\yvec} - \hat{\yvec}^\intercal \Sigma_{\hat{\yvec}}^{-1} \khatb \lb^{-\intercal} \mathbf{D}^{-1} \lb^{-1} \kbhat \Sigma_{\hat{\yvec}}^{-1} \hat{\yvec}
\end{align}
Note that,
\begin{align}
\hat{\yvec}^\intercal \Sigma_{\hat{\yvec}}^{-1} \hat{\yvec} &= \yvec^\intercal \Sigma_{\yvec}^{-1} \yvec + \yvec_\avec^\intercal \Sigma_{\yvec_\avec}^{-1} \yvec_\avec \\
\yvec^\intercal \Sigma_{\yvec}^{-1} \yvec &= \yvec^\intercal (\sigma_y^2\mathrm{I} + \alpha\Qf)^{-1} \yvec \\
\yvec_\avec^\intercal \Sigma_{\yvec_\avec}^{-1} \yvec_\avec 
	&= \yvec_\avec^\intercal (\Da + \alpha \Qa)^{-1} \yvec_\avec\\
	&= \ma^\intercal\Sa^{-1} \Da^{-1} \Sa^{-1}\ma - \alpha \ma^\intercal\Sa^{-1}\la\Ma^{-1}\laT\Sa^{-1}\ma\\
\text{and}\quad \mathbf{c} 
	&= \kbhat \Sigma_{\hat{\yvec}}^{-1} \hat{\yvec} \\
	&= \kbf \Sigma_{\yvec}^{-1} \yvec + \kba \Sigma_{\avec}^{-1} \yvec_\avec \\
	&= \kbf \Sigma_{\yvec}^{-1} \yvec + \kba \Sa^{-1} \ma - \alpha \kba \Da^{-1} \la\Ma^{-1}\laT\Sa^{-1}\ma
\end{align}
Substituting these results back into equation \cref{eqn:pep_energy},
\begin{align}
\mathcal{F} &=  -\frac{1}{2} \yvec^\intercal \Sigma_{\yvec}^{-1} \yvec + \frac{1}{2} \alpha \ma^\intercal\Sa^{-1}\la\Ma^{-1}\laT\Sa^{-1}\ma + \frac{1}{2} \mathbf{c}^\intercal \lb^{-\intercal}\mathbf{D}^{-1}\lb^{-1}\mathbf{c} \nonumber \\
&\quad - \frac{1}{2}\log\vert\mathbf{\Sigma_y}\vert - \frac{1}{2}\log\vert\mathbf{D}\vert - \frac{1}{2} \log \vert \Sa \vert + \frac{1}{2} \log \vert \kaa' \vert - \frac{1}{2\alpha}\log |\Ma| - \frac{1}{2} \ma^\intercal\Sa^{-1}\ma \nonumber \\
&\quad + \frac{N(1-\alpha)}{2\alpha} \log (\sigma_y^2) - \frac{1-\alpha}{2\alpha} \log |\Sigma_\yvec| -\frac{N}{2}\log(2\pi)
\end{align}
\subsubsection{Prediction}
We revisit and rewrite the optimal approximate distribution, $q_{\mathrm{opt}}(\bvec)$, using its natural parameters:
\begin{align}
q_\mathrm{opt}(\bvec) 
&\propto p(\bvec) \norm(\hat{\yvec}, \khatb \kbb^{-1} \bvec, \Sigma_{\hat{\yvec}}) \\
&= \norm^{-1}(\bvec;  \kbb^{-1} \kbhat \Sigma_{\hat{\yvec}}^{-1} \hat{\yvec}, \kbb^{-1} + \kbb^{-1}\kbhat \Sigma_{\hat{\yvec}}^{-1} \khatb\kbb^{-1})
\end{align}
The predictive covariance at some test points $\svec$ is,
\begin{align}
\mathbf{V}_{\svec\svec} 
&= \kss - \ksb \kbb^{-1} \kbs + \ksb \kbb^{-1} (\kbb^{-1} + \kbb^{-1}\kbhat \Sigma_{\hat{\yvec}}^{-1} \khatb\kbb^{-1})^{-1} \kbb^{-1} \kbs \\
&= \kss - \ksb \kbb^{-1} \kbs + \ksb \lb^{-\intercal} (\mathrm{I} + \lb^{-1}\kbhat \Sigma_{\hat{\yvec}}^{-1} \khatb\lb^{-\intercal})^{-1} \lb^{-\intercal} \kbs\\
&= \kss - \ksb \kbb^{-1} \kbs + \ksb \lb^{-\intercal} \mathbf{D}^{-1} \lb^{-\intercal} \kbs\
\end{align}
And, the predictive mean,
\begin{align}
\mathbf{m}_{\svec} 
&= \ksb\kbb^{-1} (\kbb^{-1} + \kbb^{-1}\kbhat \Sigma_{\hat{\yvec}}^{-1} \khatb\kbb^{-1})^{-1} \kbb^{-1} \kbhat \Sigma_{\hat{\yvec}}^{-1} \hat{\yvec} \\
&= \ksb\lb^{-\intercal} (\mathrm{I} + \lb^{-1}\kbhat \Sigma_{\hat{\yvec}}^{-1} \khatb\lb^{-\intercal})^{-1} \lb^{-1} \kbhat \Sigma_{\hat{\yvec}}^{-1} \hat{\yvec} \\
&= \ksb\lb^{-\intercal} \mathbf{D}^{-1} \lb^{-1} \kbhat \Sigma_{\hat{\yvec}}^{-1} \hat{\yvec} 
\end{align}

\section{Equivalence results}
\begin{figure}[!ht]
\centering
\includegraphics[width=\textwidth]{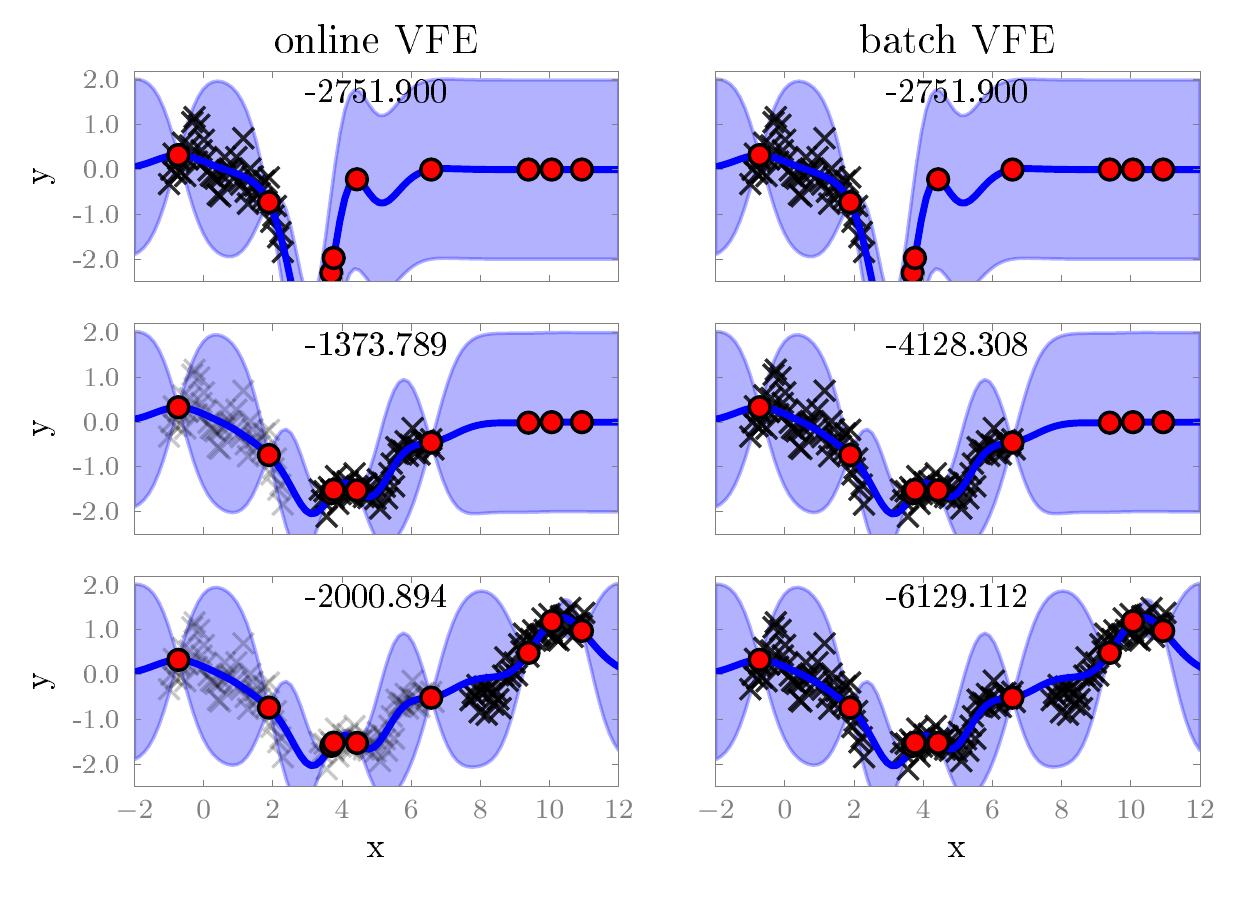}
\caption{Equivalence between the streaming variational approximation and the batch variational approximation when hyperparameters and pseudo-inputs are fixed. The inset numbers are the approximate marginal likelihood (the variational free energy) for each model. Note that the numbers in the batch case are the cumulative sum of the numbers on the left for the streaming case. Small differences, if any, are merely due to numerical computation. \label{fig:equi_sparse}}
\end{figure}

When the hyperparameters and the pseudo-inputs are fixed, $\alpha$-divergence inference for streaming sparse GP regression recovers the batch solutions provided by Power-EP with the same $\alpha$ value. In other words, only a single pass through the data is necessary for Power-EP to converge in sparse GP regression. This result is in a similar vein to the equivalence between sequential inference and batch inference in full GP regression, when the hyperparameters are kept fixed. As an illustrative example, assume that $\zvec_a = \zvec_b$ and $\theta$ is kept fixed, and $\{\xvec_1, \yvec_1\}$ and $\{\xvec_2, \yvec_2\}$ are the first and second data batches respectively. The optimal variational update gives,
\begin{align}
q_1(\avec) 
	&\propto p(\avec) \exp \int \dd \fvec_1 p(\fvec_1|\avec) \log p(\yvec_1|\fvec_1) \\
q_2(\avec)
	&\propto q_1(\avec) \exp \int \dd \fvec_2 p(\fvec_2|\avec) \log p(\yvec_2|\fvec_2) \propto p(\avec) \exp \int \dd \fvec p(\fvec|\avec) \log p(\yvec|\fvec) \label{eqn:opt_fixed}
\end{align}
where $\yvec = \{\yvec_1, \yvec_2\}$ and $\fvec = \{\fvec_1, \fvec_2\}$. \Cref{eqn:opt_fixed} is exactly identical to the optimal variational approximation for the batch case of \cite{titsias2009variational}, when we group all data batches into one. A similar procedure can be shown for Power-EP.
We demonstrate this equivalence in \cref{fig:equi_sparse}.

\begin{figure}[!ht]
\centering
\includegraphics[width=\textwidth]{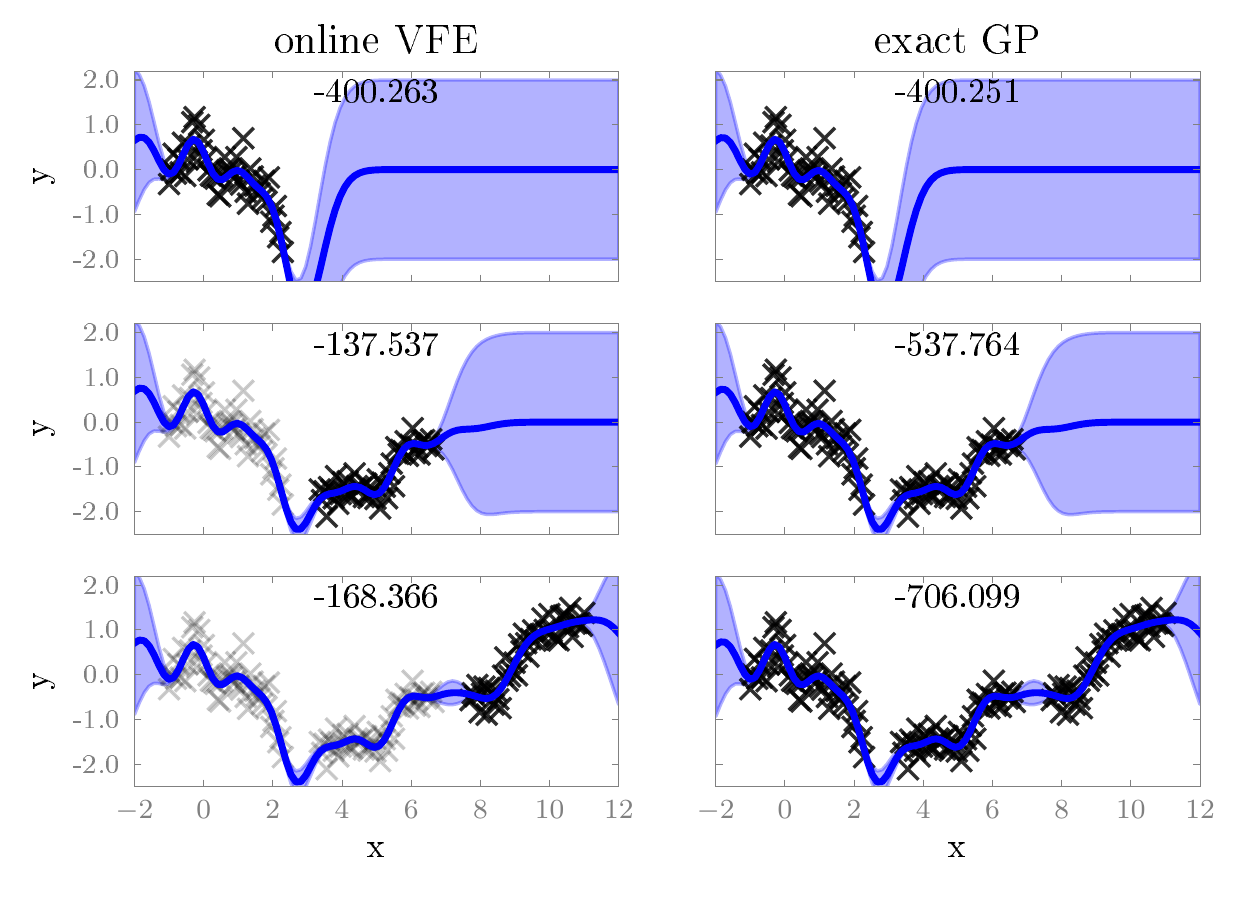}
\caption{Equivalence between the streaming variational approximation and the exact GP regression when hyperparameters and pseudo-inputs are fixed, and the pseudo-points are at the training points. The inset numbers are the (approximate) marginal likelihood for each model. Note that the numbers in the batch case are the cumulative sum of the numbers on the left for the streaming case. Small differences, if any, are merely due to numerical computation.\label{fig:equi_full}}
\end{figure}

In addition, in the setting where hyperparameters and the pseudo-inputs are fixed, if pseudo-points are added at each stage at the new data input locations, the method returns the true posterior and marginal likelihood. This equivalence is demonstrated in \cref{fig:equi_full}.

\section{Extra experimental results}

\subsection{Hyperparameter learning on synthetic data}
\begin{figure}[!ht]
\centering
\includegraphics[width=\textwidth]{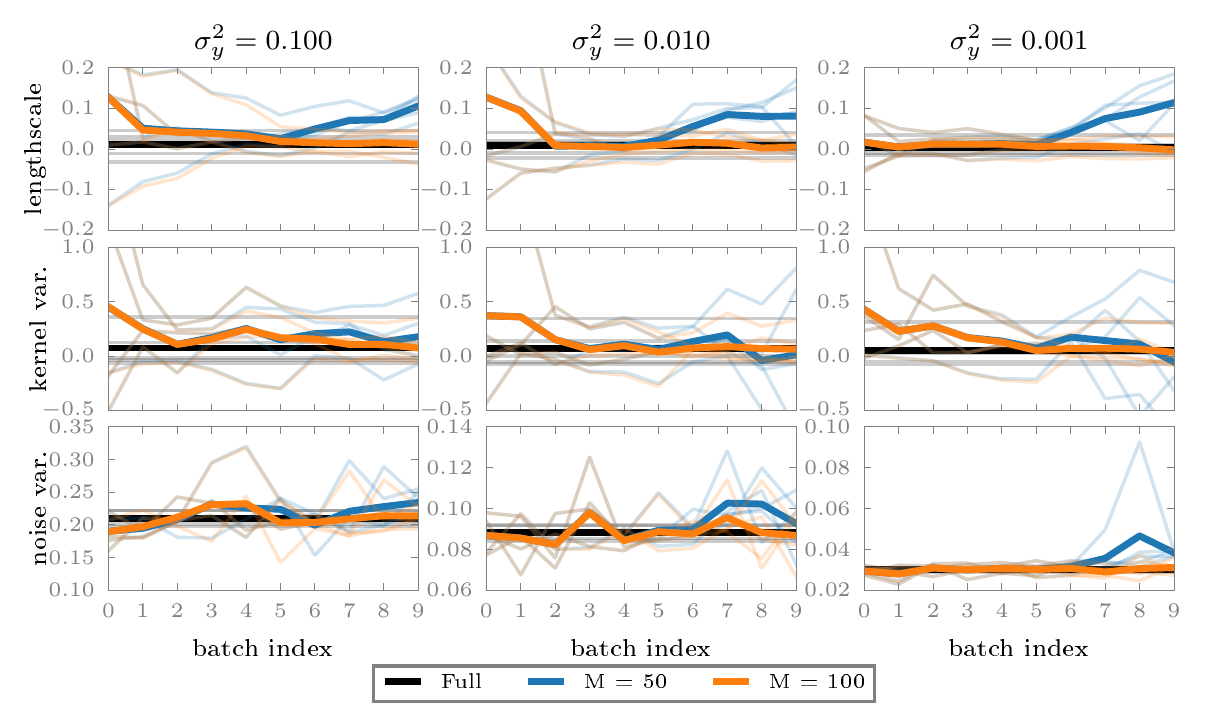}
\caption{Learnt hyperparameters on a time series dataset, that was generated from a GP with an exponentiated quadratic kernel and with a lengthscale of 0.5\label{fig:hyper_1}. Note the $y-$axis show the difference between the learnt values and the groundtruth.}
\end{figure}

In this experiment, we generated several time series from GPs with known kernel hyperparameters and observation noise. We tracked the hyperparameters as the streaming algorithm learns and plot their traces in \cref{fig:hyper_1,fig:hyper_2}. It could be seen that for the smaller lengthscale, we need more pseudo-points to cover the input space and to learn correct hyperparameters. Interestingly, all models including full GP regression on the entire dataset tend to learn bigger noise variances. Overall, the proposed streaming method can track and learn good hyperparameters; and if there is enough pseudo-points, this method performs comparatively to full GP on the entire dataset.

\begin{figure}
\centering
\includegraphics[width=\textwidth]{{{figs/hyper_compare_plot_0_80}}}
\caption{Learnt hyperparameters on a time series dataset, that was generated from a GP with an exponentiated quadratic kernel and with a lengthscale of 0.8\label{fig:hyper_2}. Note the $y-$axis show the difference between the learnt values and the groundtruth.}
\end{figure}

\subsection{Learning and inference on a toy time series}
\begin{figure}
\centering
  \includegraphics[width=1.0\textwidth]{./figs/VFE_opt_M_10_False}
  \includegraphics[width=1.0\textwidth]{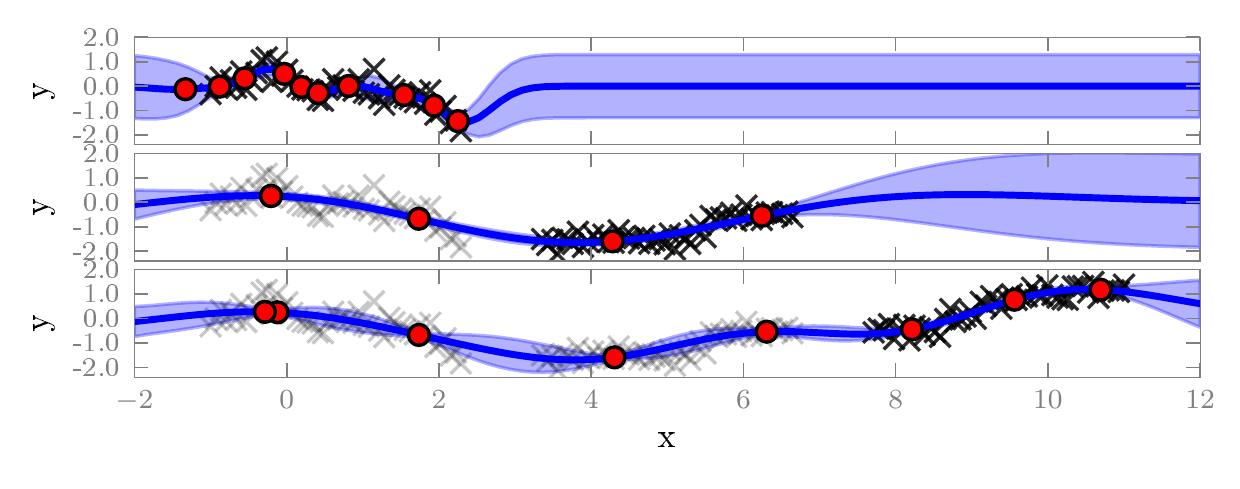}
\caption{Online regression on a toy time series using variational inference (top) and Power-EP with $\alpha=0.5$ (bottom), in a non-iid setting. The black crosses are data points (past points are greyed out), the red circles are pseudo-points, and blue lines and shaded areas are the marginal predictive means and confidence intervals at test points.\label{fig:toy_1}}
\end{figure}

\begin{figure}
\centering
\includegraphics[width=1.0\textwidth]{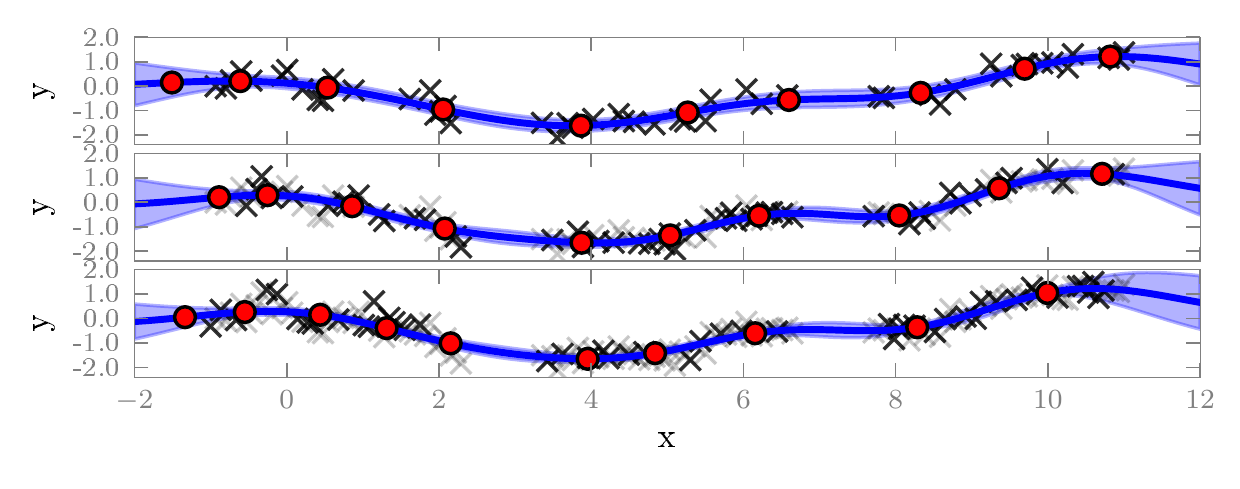}
\includegraphics[width=1.0\textwidth]{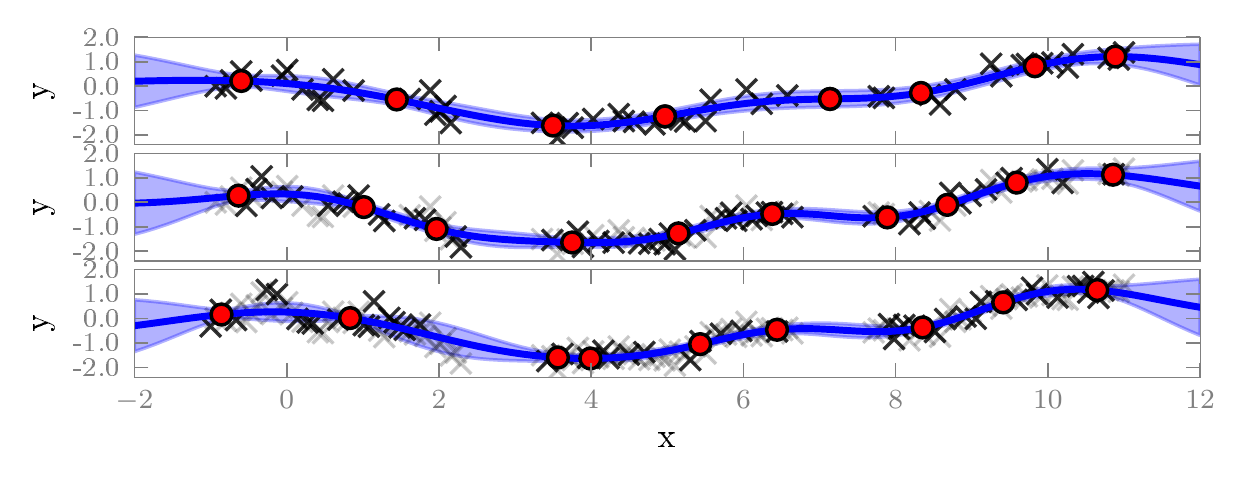}
\caption{Online regression on a toy time series using variational inference (top) and Power-EP with $\alpha=0.5$ (bottom), in an iid setting. The black crosses are data points (past points are greyed out), the red circles are pseudo-points, and blue lines and shaded areas are the marginal predictive means and confidence intervals at test points.\label{fig:toy_2}}
\end{figure}

As shown in the main text, we construct a synthetic time series to demonstrate the learning procedure as data arrives sequentially. \Cref{fig:toy_1,fig:toy_2} show the results for non-iid and iid streaming settings respectively. 

\subsection{Binary classification}
We consider a binary classification task on the benchmark {\it banana} dataset. In particular, we test two streaming settings, non-iid and iid, as shown in \cref{fig:binary_1,fig:binary_2} respectively. In all cases, the streaming algorithm performs well and reaches the performance of the batch case using a sparse variational method \cite{HenMatGha15} (as shown in the right-most plots).
\begin{figure}[!ht]
\centering
\includegraphics[width=\textwidth]{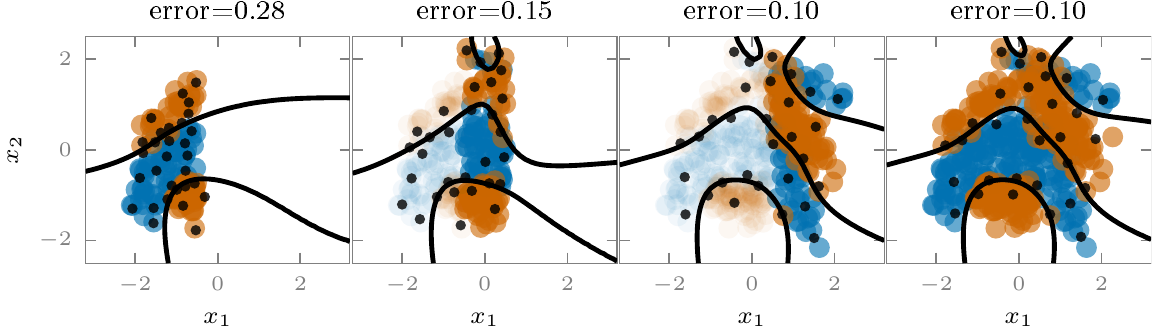}
\caption{Classifying binary data in a non-iid streaming setting. The right-most plot shows the prediction made by using sparse variational inference on full training data.\label{fig:binary_1}}
\end{figure}

\begin{figure}[!ht]
\centering
\includegraphics[width=\textwidth]{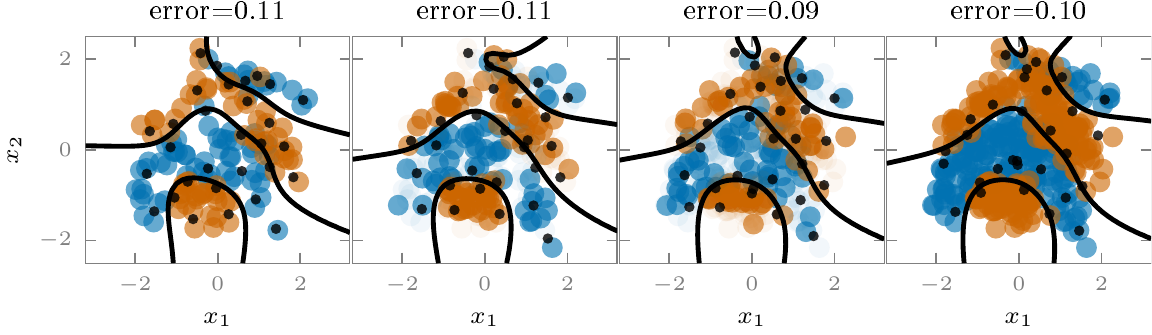}
\caption{Classifying binary data in an iid streaming setting. The right-most plot shows the prediction made by using sparse variational inference on full training data.\label{fig:binary_2}}
\end{figure}

\subsection{Sensitivity to the order of the data}
We consider the classification task above but now with more (smaller) mini-batches and the order of the batches are varied. The aim is to evaluate the sensitivity of the algorithm to the order of the data. The classification errors as data arrive are included in \cref{tab:order} and are consistent with what we included in the main text.

\begin{table}[h!]
\small
\centering
\caption{Classification errors as data arrive in different orders}
\label{tab:order}
 \begin{tabular}{|| c || c | c | c | c | c | c | c | c | c | c ||} 
 \hline
 Order/Index & 1 & 2 & 3 & 4 & 5 & 6 & 7 & 8 & 9 & 10 \\ [0.5ex]
 \hline\hline
Left to Right & 0.255 & 0.145 & 0.1325 & 0.1225 & 0.1075 & 0.11 & 0.105 & 0.1 & 0.0925 & 0.0875 \\
Right to Left & 0.255 & 0.1475 & 0.1325 & 0.12 & 0.105 & 0.1025 & 0.0975 & 0.0925 & 0.09 & 0.095\\
Random & 0.5025 & 0.2775 & 0.26 & 0.2725 & 0.2875 & 0.1975 & 0.1125 & 0.125 & 0.105 & 0.095\\
Batch & & & & & & & & & & 0.095\\ [1ex] 
\hline
\end{tabular}
\end{table}

\subsection{Additional plots for the time-series and spatial datasets}

\begin{figure}[!ht]
\begin{center}
    \begin{subfigure}[t]{\textwidth}
    \centering
    \includegraphics[width=6cm]{figs/legend.pdf}
    \end{subfigure}
    
    \begin{subfigure}[t]{0.48\textwidth}
    \centering
    \includegraphics[width=6cm]{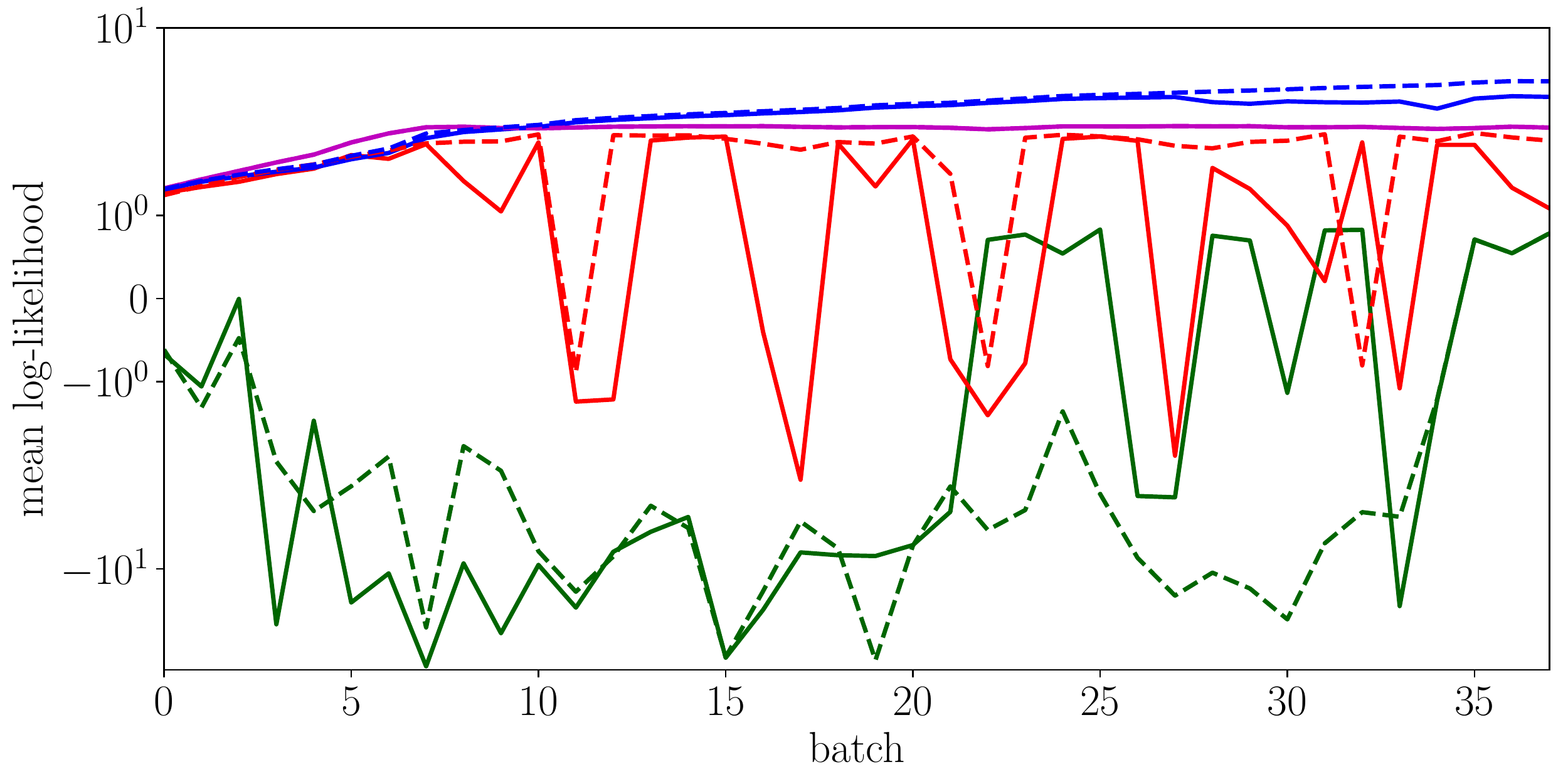}
    \end{subfigure}
    \begin{subfigure}[t]{0.48\textwidth}
    \centering
    \includegraphics[width=6cm]{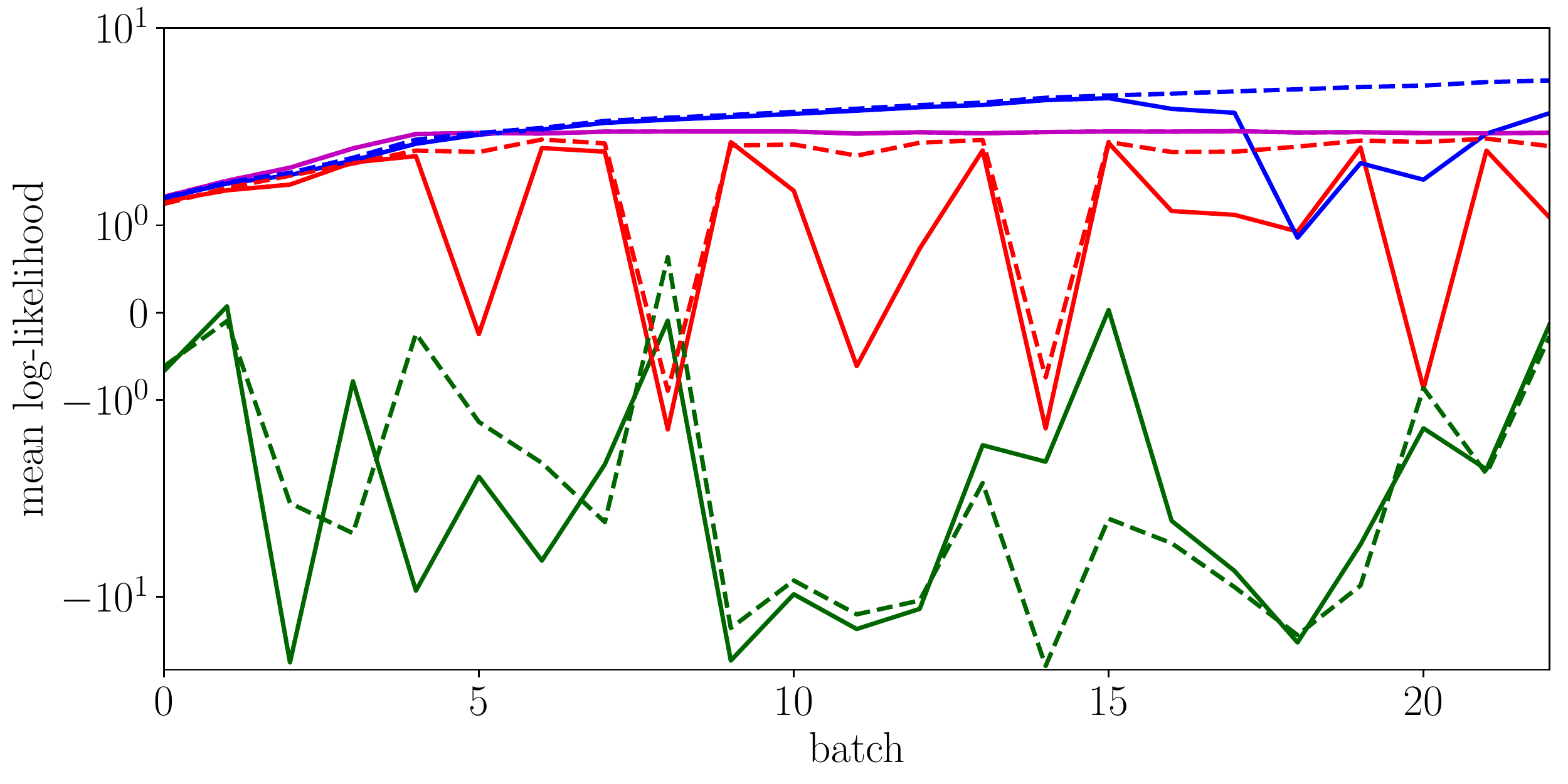}
    \end{subfigure}
    
    \begin{subfigure}[t]{.48\textwidth}
    \centering
    \includegraphics[width=6.1cm]{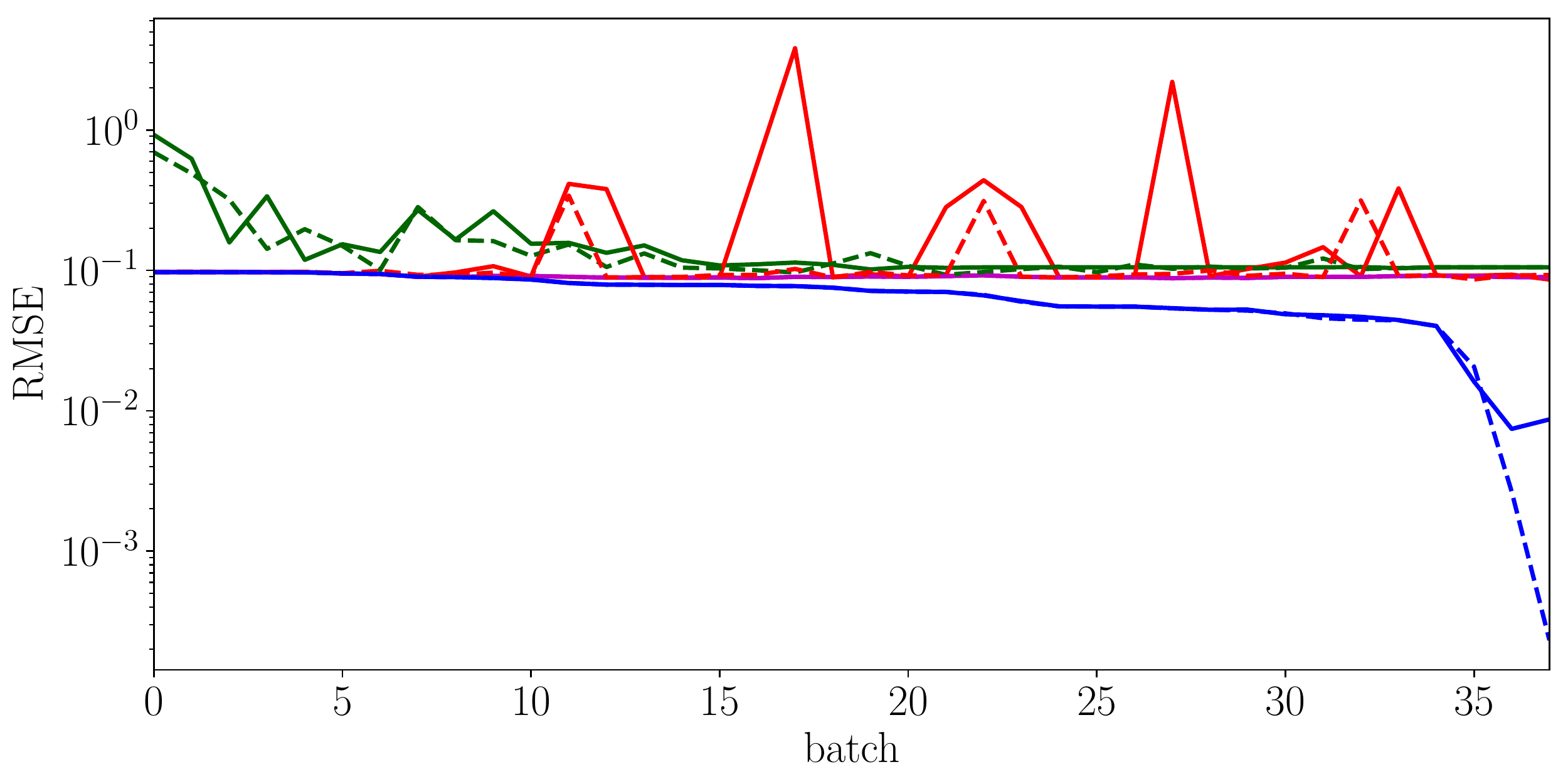}
    \caption*{\qquad pseudo periodic data, batch size = $300$}
    \end{subfigure}
    \begin{subfigure}[t]{.48\textwidth}
    \centering
    \includegraphics[width=6.1cm]{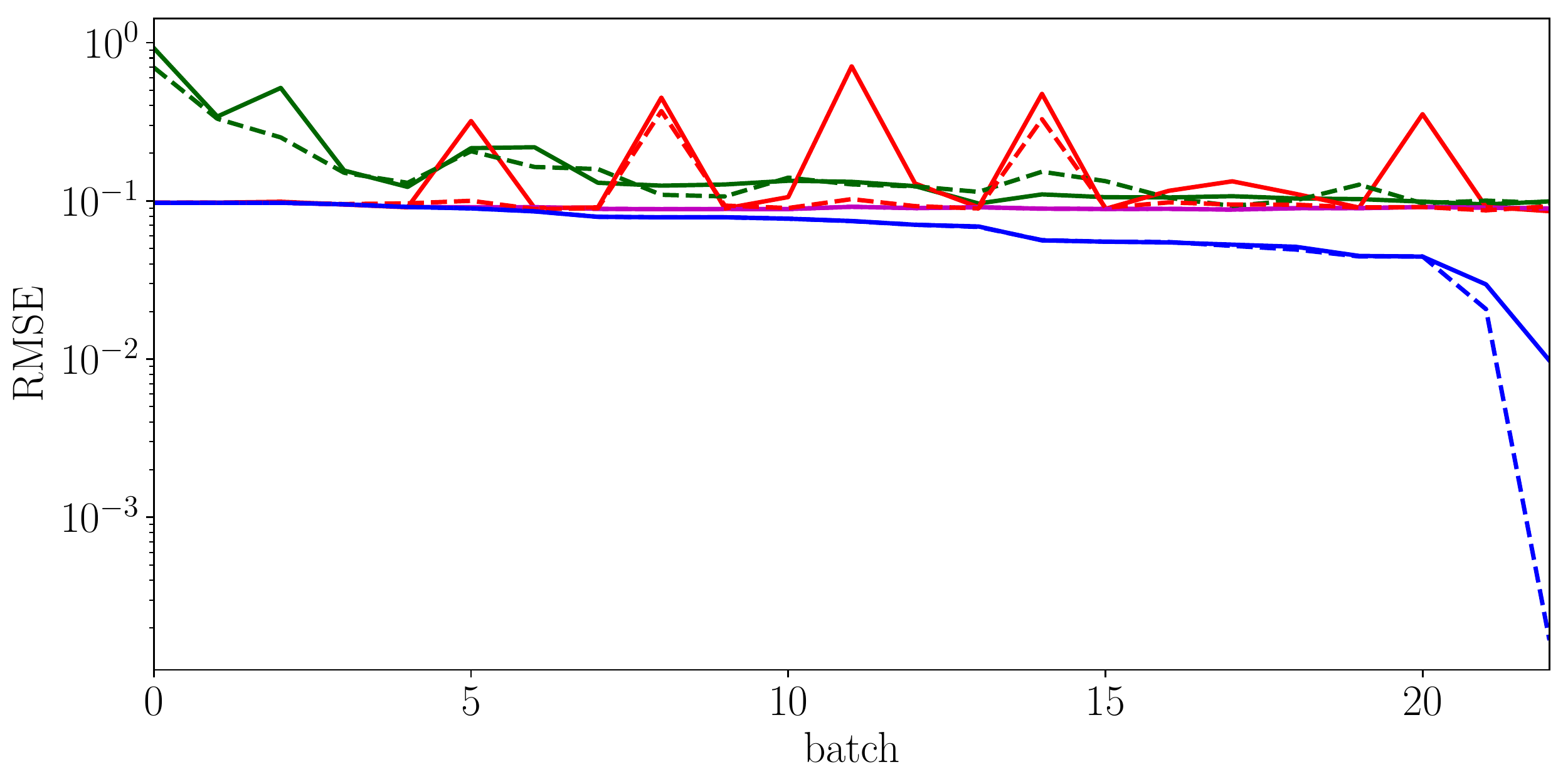}
    \caption*{\qquad pseudo periodic data, batch size = $500$}
    \end{subfigure}
    
    {\vskip 0.2cm}
    \begin{subfigure}[t]{.48\textwidth}
    \centering
    \includegraphics[width=6cm]{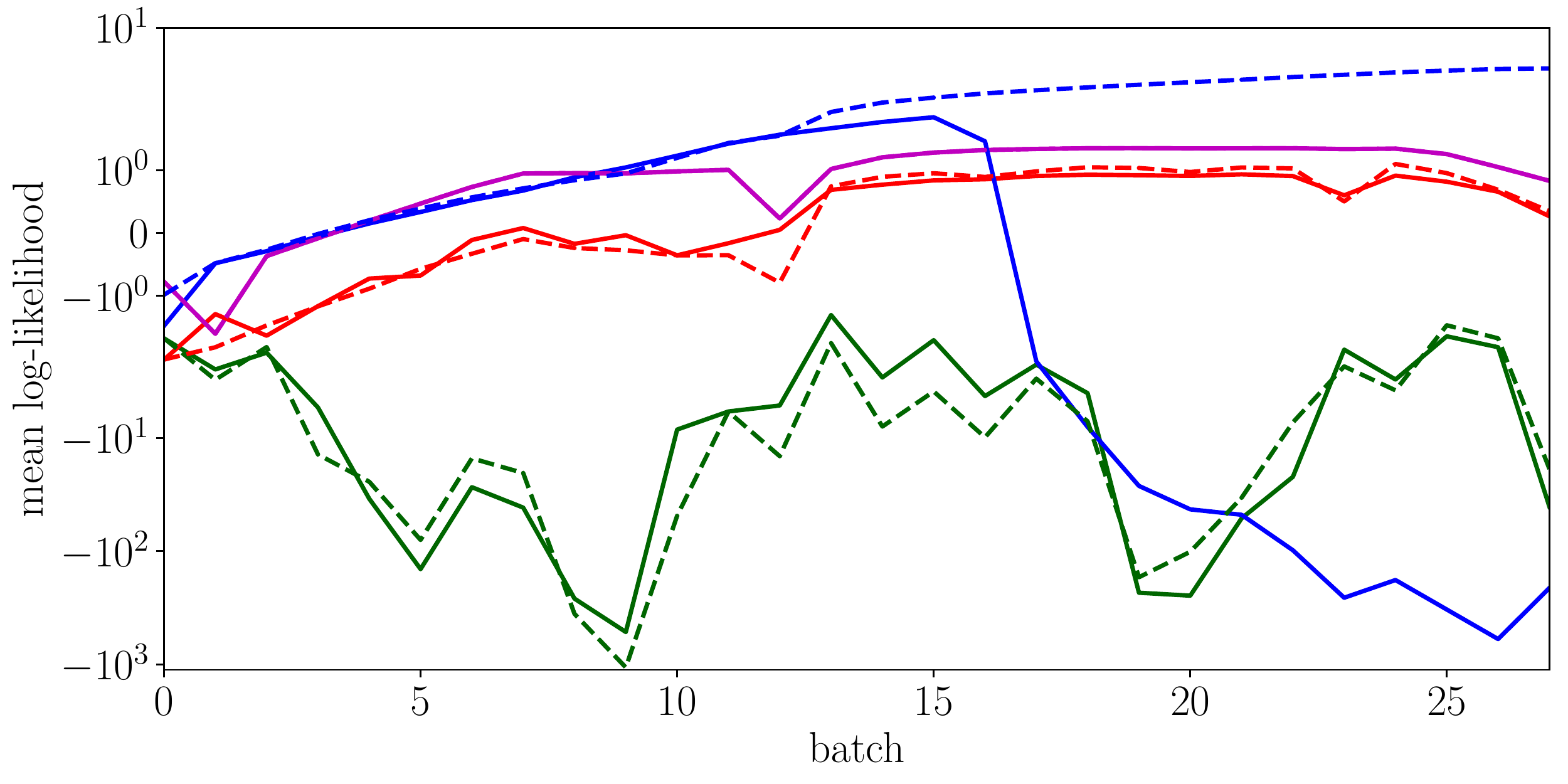}
    \end{subfigure}
    \begin{subfigure}[t]{.48\textwidth}
    \centering
    \includegraphics[width=6cm]{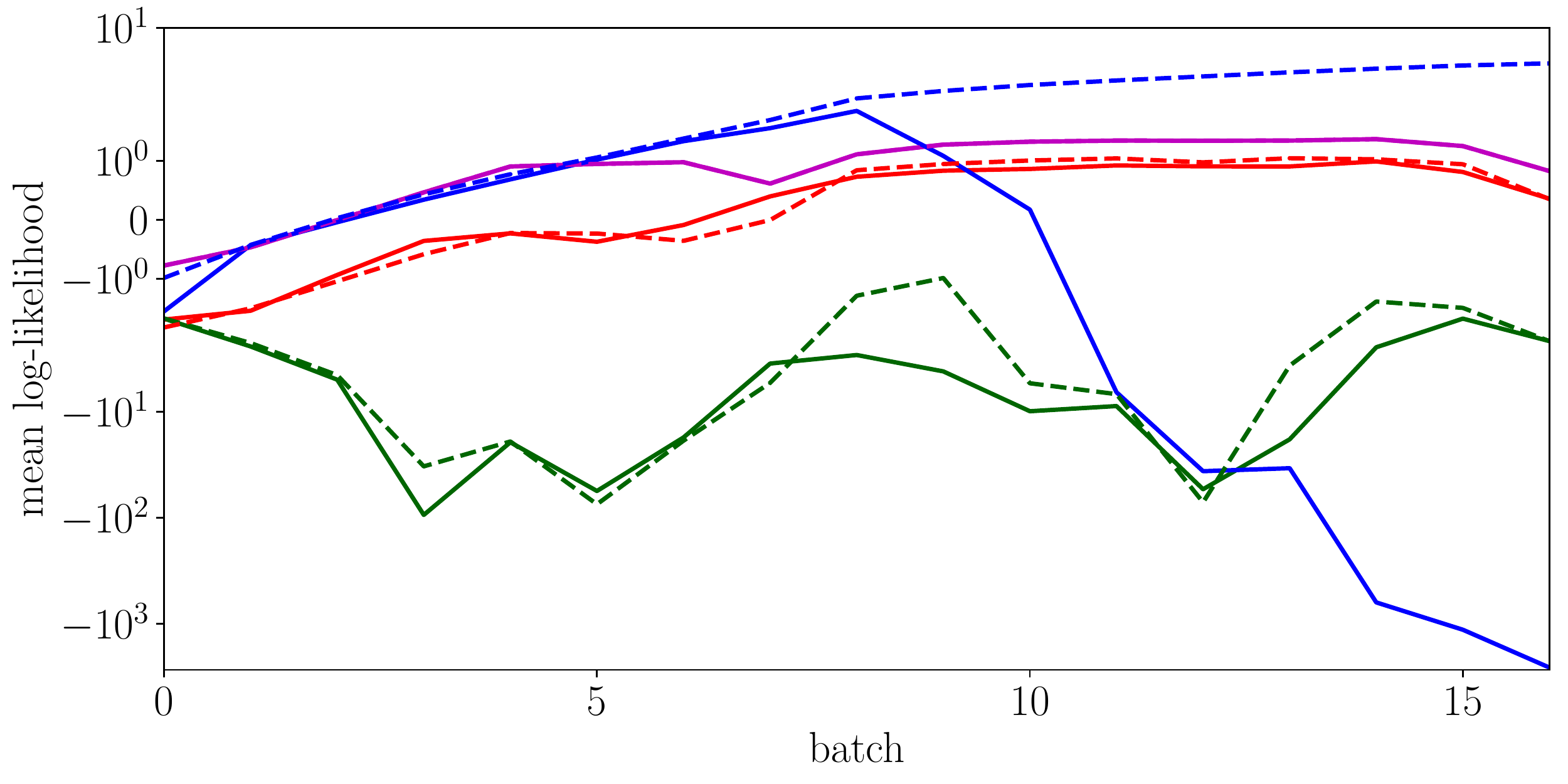}
    \end{subfigure}

    \begin{subfigure}[t]{.48\textwidth}
    \centering
    \includegraphics[width=6.1cm]{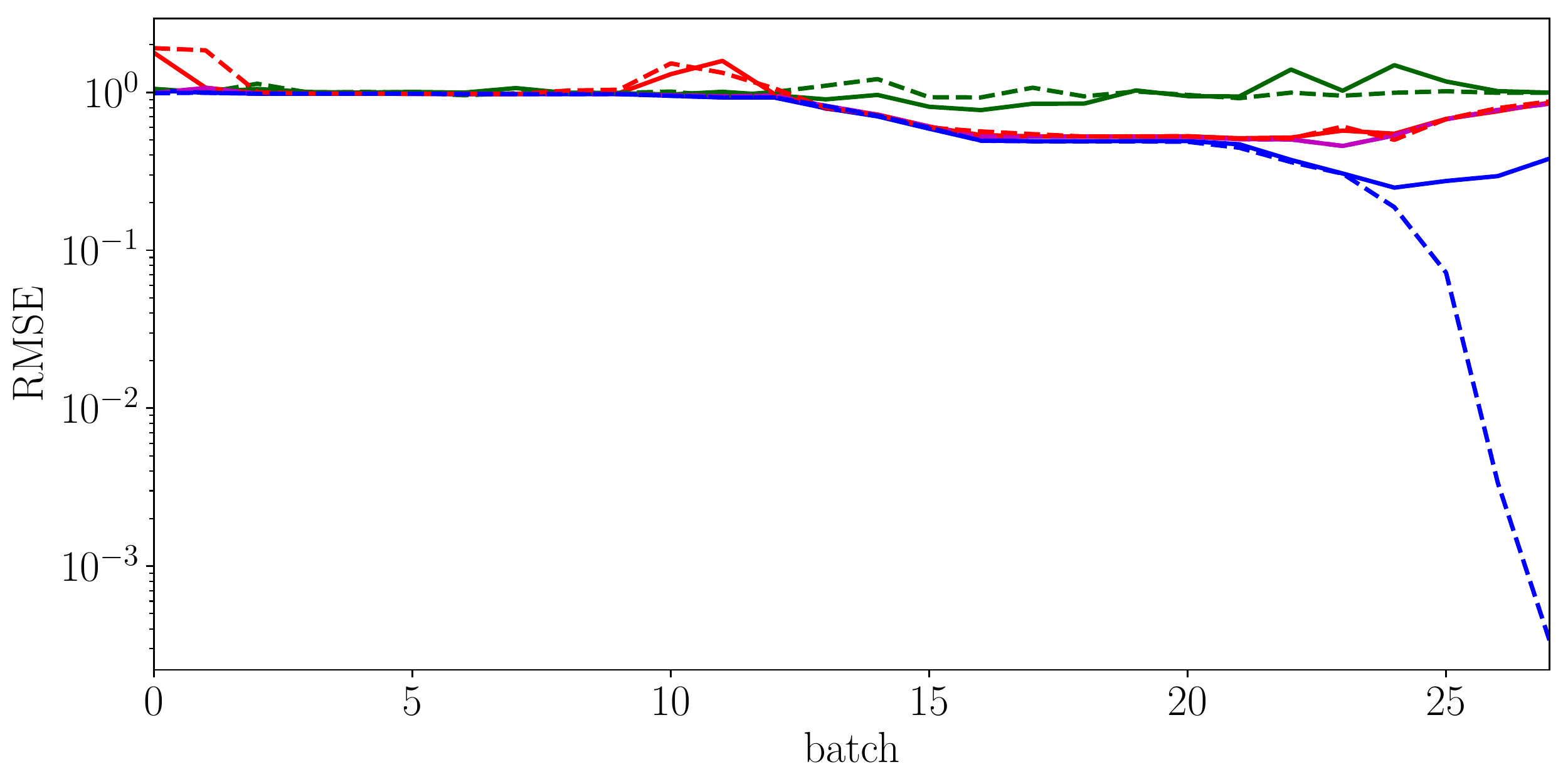}
    \caption*{\qquad audio data, batch size = $300$}
    \end{subfigure}
    \begin{subfigure}[t]{.48\textwidth}
    \centering
    \includegraphics[width=6.1cm]{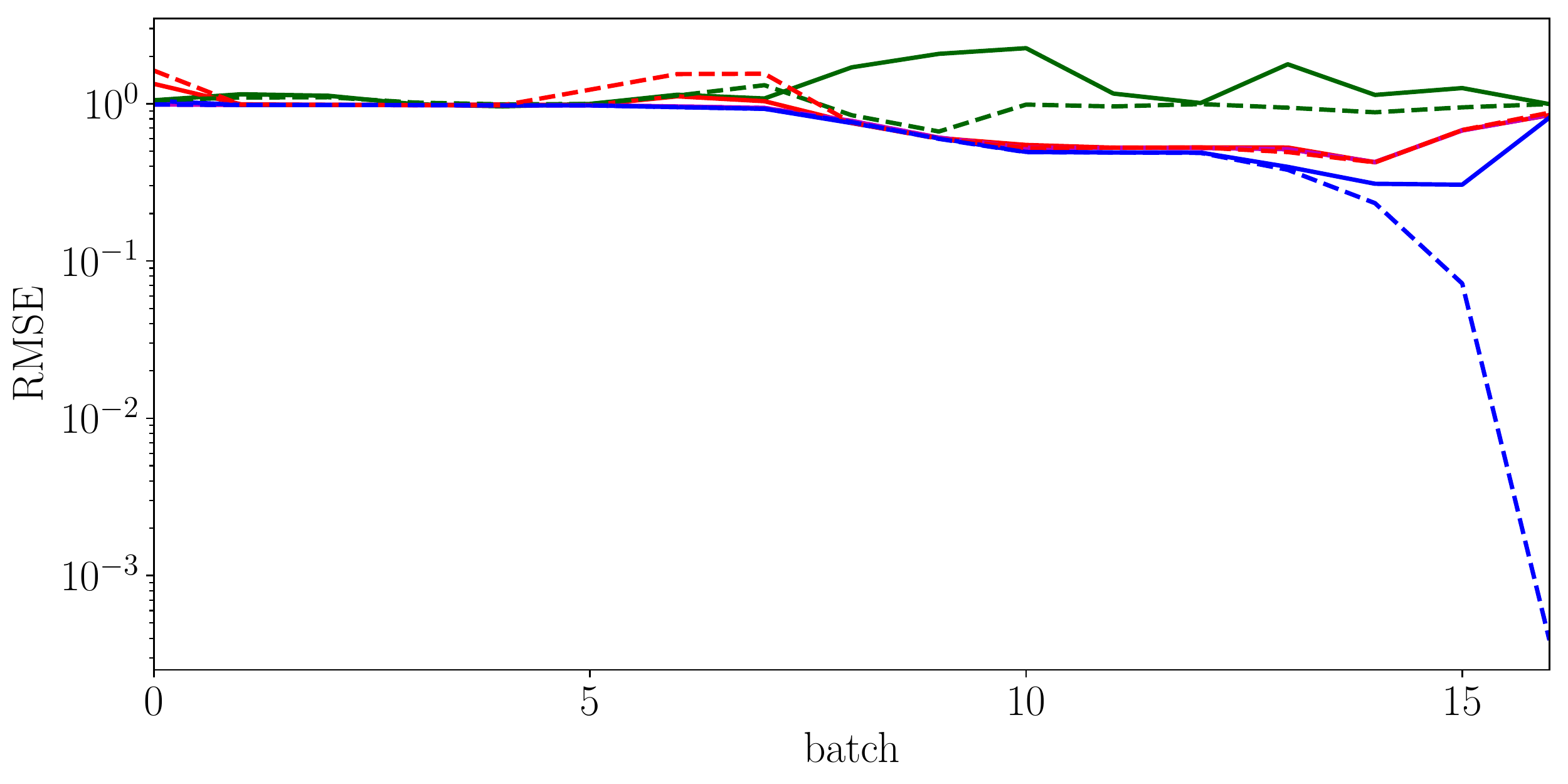}
    \caption*{\qquad audio data, batch size = $500$}
    \end{subfigure}
\caption{Results for time-series datasets with batch sizes $300$ and $500$. The solid and dashed lines are for $M=100, 200$ respectively.}
\label{fig:time-series-aligned}
\end{center}
\end{figure}

\begin{figure}[t]
\begin{center}
	\begin{subfigure}[t]{0.48\textwidth}
    \centering
	\includegraphics[width=6cm]{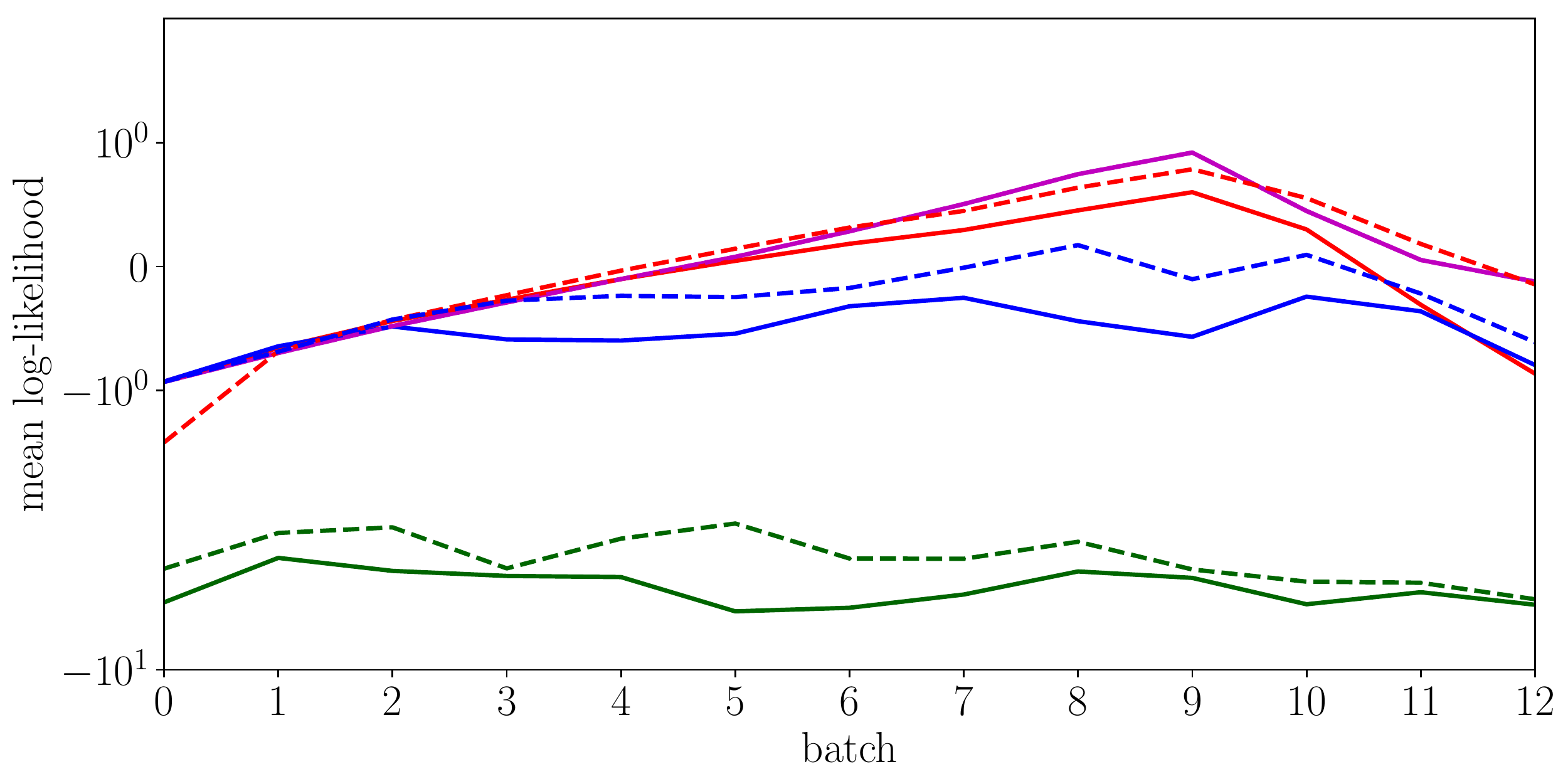}
	\end{subfigure}
	\begin{subfigure}[t]{0.48\textwidth}
    \centering
	\includegraphics[width=6cm]{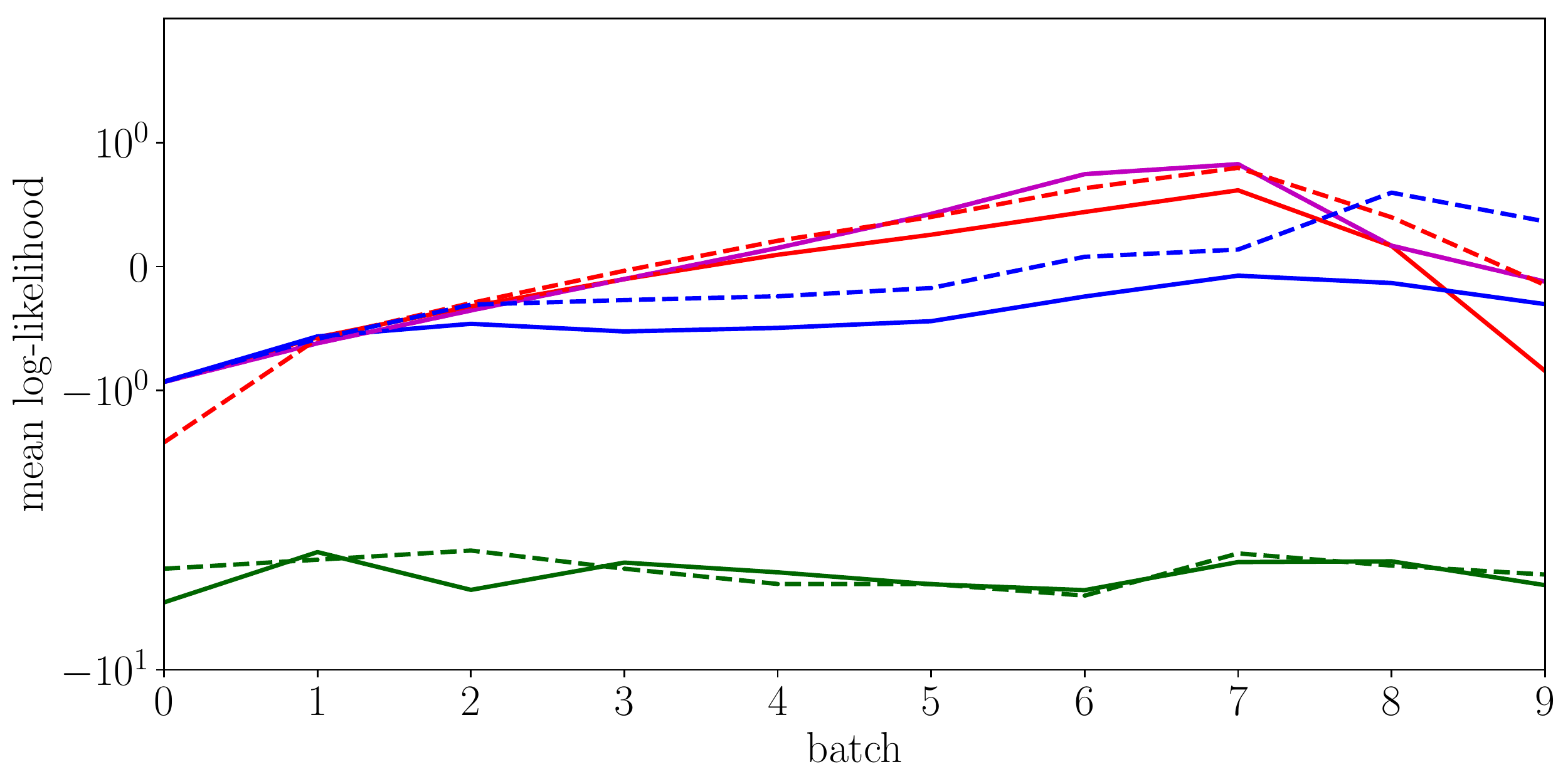}
	\end{subfigure}
    
    \begin{subfigure}[t]{0.48\textwidth}
    \centering
	\includegraphics[width=6cm]{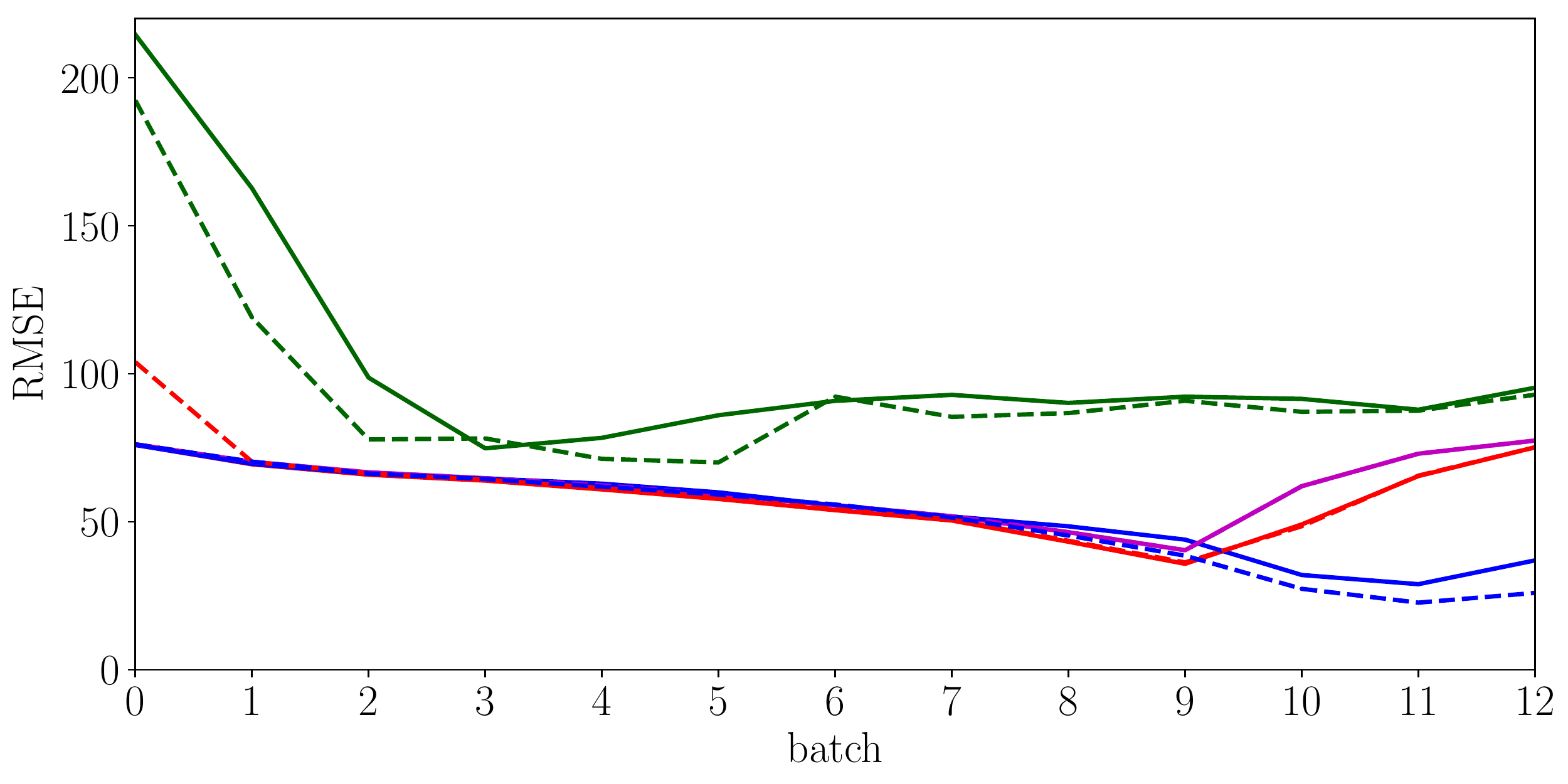}
    \caption*{\qquad terrain data, batch size = $750$}
    \end{subfigure}
    \begin{subfigure}[t]{0.48\textwidth}
    \centering
	\includegraphics[width=6cm]{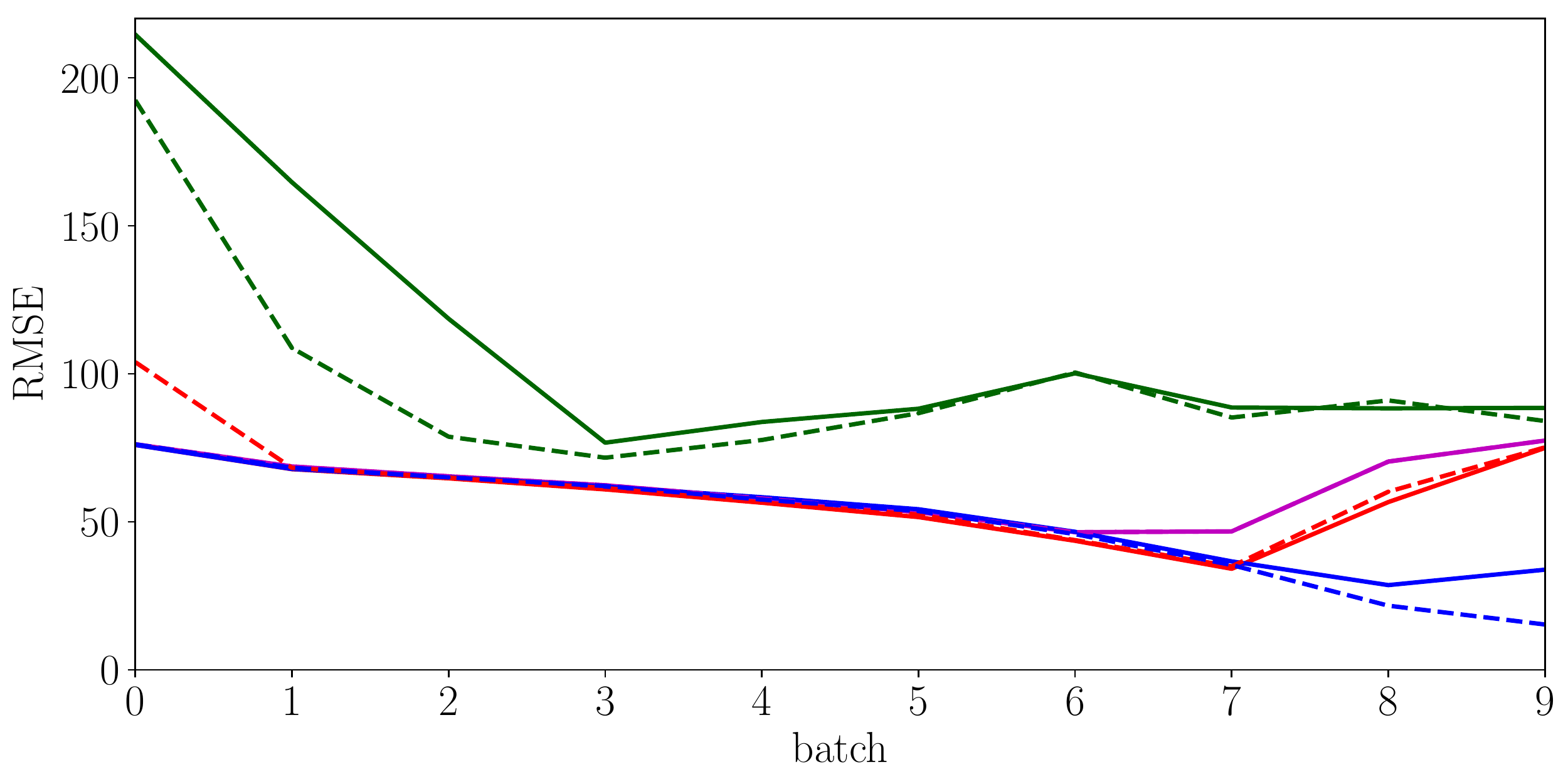}
    \caption*{\qquad terrain data, batch size = $1000$}
    \end{subfigure}
\caption{Results for spatial data (see \cref{fig:time-series-aligned} for the legend). Solid and dashed lines indicate the results for $M = 400, 600$ pseudo-points respectively.}
\label{fig:spatial-aligned}
\end{center}
\end{figure}

In this section, we plot the mean marginal log-likelihood and RMSE against the number of batches for the models in the ``speed versus accuracy'' experiment in the main text.
Fig. \ref{fig:time-series-aligned} shows the results for the time-series datasets while \cref{fig:spatial-aligned} shows the results for the spatial datasets.

\end{appendices}

\end{document}